\documentclass{article}

 \usepackage[preprint]{neurips_2026}

\usepackage[utf8]{inputenc} % allow utf-8 input
\usepackage[T1]{fontenc}    % use 8-bit T1 fonts
\usepackage{hyperref}       % hyperlinks
\usepackage{url}            % simple URL typesetting
\usepackage{booktabs}       % professional-quality tables
\usepackage{amsfonts}       % blackboard math symbols
\usepackage{nicefrac}       % compact symbols for 1/2, etc.
\usepackage{microtype}      % microtypography
\usepackage{xcolor}         % colors
\usepackage{siunitx}
\usepackage{graphicx}
\usepackage{makecell}
\usepackage{caption}

\DeclareSIUnit{\molar}{M}
\usepackage{xcolor}
\usepackage{booktabs}
\usepackage{fancyvrb}
\usepackage{fvextra}        % breaklines/breakanywhere for Verbatim
\usepackage[most,breakable]{tcolorbox}

\definecolor{syscolor}{RGB}{120,120,120}
\definecolor{usercolor}{RGB}{31,90,150}
\definecolor{assistcolor}{RGB}{40,120,60}
\definecolor{toolcolor}{RGB}{180,100,20}
\definecolor{outcolor}{RGB}{100,100,110}
\definecolor{submitcolor}{RGB}{120,40,140}
\definecolor{summarycolor}{RGB}{40,40,40}
\definecolor{artifactcolor}{RGB}{0,80,120}
\definecolor{passcolor}{RGB}{0,120,40}
\definecolor{failcolor}{RGB}{180,30,30}

\fvset{
  fontsize=\scriptsize,
  breaklines=true,
  breakanywhere=true,
  breaksymbol={\textcolor{gray}{\tiny$\hookrightarrow$}},
  breakindent=0pt,
  xleftmargin=0pt,
  xrightmargin=0pt,
}

\tcbset{
  trajbox/.style={
    breakable,
    enhanced jigsaw,
    boxrule=0.4pt,
    arc=1pt,
    left=4pt,right=4pt,top=2pt,bottom=2pt,
    fonttitle=\bfseries\small,
    fontupper=\footnotesize,
    before skip=4pt, after skip=4pt,
  },
}

\newtcolorbox{sysbox}{trajbox, title=System Prompt, colback=syscolor!8, colframe=syscolor!60, coltitle=white, colbacktitle=syscolor!70}
\newtcolorbox{userbox}[1]{trajbox, title=#1, colback=usercolor!8, colframe=usercolor!60, coltitle=white, colbacktitle=usercolor!70}
\newtcolorbox{assistbox}[1]{trajbox, title=#1, colback=assistcolor!8, colframe=assistcolor!60, coltitle=white, colbacktitle=assistcolor!70}
\newtcolorbox{toolbox}[1]{trajbox, title=#1, colback=toolcolor!8, colframe=toolcolor!60, coltitle=white, colbacktitle=toolcolor!70}
\newtcolorbox{outbox}[1]{trajbox, title=#1, colback=outcolor!6, colframe=outcolor!50, coltitle=white, colbacktitle=outcolor!60}
\newtcolorbox{submitbox}[1]{trajbox, title=#1, colback=submitcolor!8, colframe=submitcolor!60, coltitle=white, colbacktitle=submitcolor!70}
\newtcolorbox{summarybox}[1]{trajbox, title=#1, colback=summarycolor!5, colframe=summarycolor!50, coltitle=white, colbacktitle=summarycolor!70}
\newtcolorbox{artifactbox}[1]{trajbox, title=Submitted artifact: \texttt{#1}, colback=artifactcolor!6, colframe=artifactcolor!60, coltitle=white, colbacktitle=artifactcolor!70}
\title{SMDD-Bench: Can LLMs Solve Real-World Small Molecule Drug Design Tasks?}

\author{%
  Kevin Han\thanks{kevinhan@cmu.edu} \\
  Carnegie Mellon University\\
  \And
  Renfei Zhang \\
  Carnegie Mellon University \\
  \texttt{}
  \AND
  Kathy Wei \\
  Stealth \\
  \And
  Hamed Mahdavi \\
  Pennsylvania State University
  \And
  Niloofar Mireshghallah \\
  Carnegie Mellon University
  \And
  Amir Barati Farimani \\
  Carnegie Mellon University
}

\begin{document}

\addtocontents{toc}{\protect\setcounter{tocdepth}{0}}

\maketitle

\begin{abstract}
  LLM agents have incredible potential for scientific discovery applications. However, the performance of LLM agents on real-world, small molecule drug design (SMDD) tasks across diverse chemistries and targets is unclear. Current evaluation methods are either ad hoc, too simple for real-world discovery, limited in scale, or restricted to single-turn question answering. In effort to standardize the evaluation of LLM agents on small molecule design, we introduce SMDD-Bench, a challenging, multi-turn, long-horizon agentic benchmark consisting of 502 guaranteed-solvable task instances spanning 5 task types: 2D Pharmacophore Identification, Interaction Point Discovery, Scaffold Hopping, Lead Optimization, and Fragment Assembly. SMDD-Bench tasks span a wide region of chemical space and involve 102 unique protein targets. Completely solving the benchmark would require having strong chemical and biological reasoning and 3D intuition, understanding specialized tool use, and displaying planning expertise over a limited number of oracle calls. We benchmark 7 frontier open and closed source LLMs and find even the most performant LLM, GPT5.4, solves only 40.2\% of tasks. We hope SMDD-Bench provides a standardized testbed to invigorate the field towards training and evaluating LLM agents for fully autonomous computational drug design. We host a public leaderboard at \url{smddbench.com}.
\end{abstract}

\section{Introduction}

\begin{figure}
  \centering
  \includegraphics[width=0.95\textwidth, trim=50px 150px 50px 100px, clip]{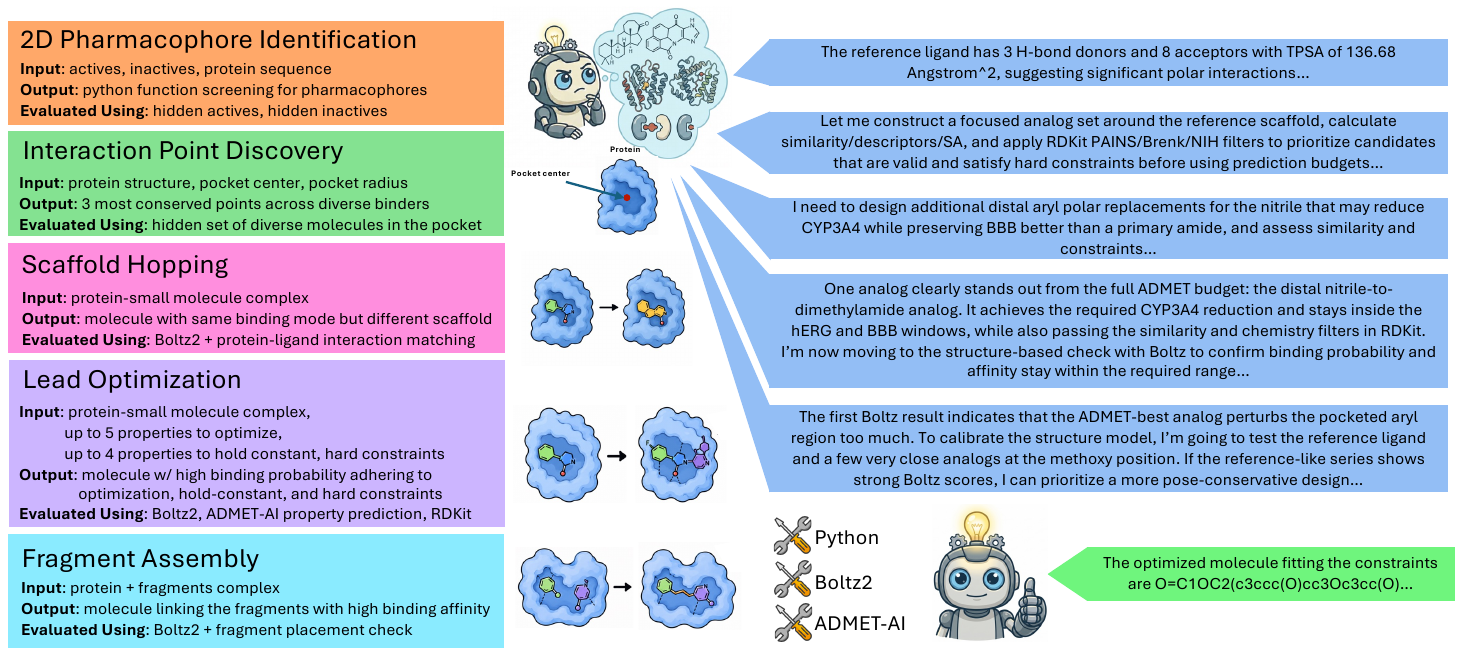}
  \caption{Overview of SMDD-Bench's task types and example reasoning turns.}
  \vspace{-3ex}
  \label{fig:fig1}
\end{figure}

LLMs have seen considerable success in many scientific discovery tasks such as synthesizing scientific literature, orchestrating chemistry experiments, and even performing end to end applied machine learning research \citep{boiko2023autonomous, lu2026towards, asai2026synthesizing}. Within the drug discovery and medicinal chemistry space, LLMs have been benchmarked on tasks including single step retrosynthesis prediction, molecular description, de novo molecular generation, and property prediction \citep{yu2024llasmol, chembench, huang2021therapeutics, bartmann2026moleculariq}. However, these benchmarks are either restricted to single-turn question answering, are rudimentary and unrealistic when compared to the complexity of real-world computational drug discovery, or are limited to only a small non-diverse handful of tasks and protein targets. Although such benchmarks are useful in understanding an LLM's ability to understand basic chemistry principles, they are not as useful for understanding and improving an LLM \textbf{agent's} ability to handle the geometric and chemical complexity and diversity of drug discovery tasks facing expert human medicinal chemists.

In this work, we present SMDD-Bench, an agentic benchmark designed to evaluate LLM agents, multi-agent systems, and agentic workflows on complex and realistic small molecule drug design (SMDD) tasks facing human medicinal chemists. SMDD-Bench is comprised of 5 task types with a total of 502 task instances. The task types are 2D Pharmacophore Identification, Interaction Point Discovery, Scaffold Hopping, Lead Optimization, and Fragment Assembly. Each task is deeply embedded into real-world drug discovery workflows and requires deep chemical intuition, biological reasoning ability, expertise in selecting and running specialized cheminformatics tools, hypothesis generation and validation, and understanding 3D geometries in 3D space. We also introduce the notion of witness-aware task generation, allowing us to generate a large number of chemically valid and guaranteed-solvable task instances procedurally and without a human in the loop. 

Real world small molecule drug discovery requires the LLM to be able to write RDKit code, analyze protein-ligand interactions in 3D space, propose targeted molecular edits, understand how those molecular edits affect predicted properties, binding affinities, and 3D conformations, and run specialized cheminformatics tools. These skills go far beyond the simple single-turn question answering tasks that previous benchmarks study and require the development of a new and challenging agentic benchmark consisting of a diverse spread of chemistries and protein targets. 

However, creating real-world drug discovery tasks can be difficult due to the requirement for there to be at least one known solution to the task. For example, in lead optimization, while it is tempting to create a task instance by sampling a protein-ligand pair before sampling a set of properties to optimize as well as a set of properties to hold constant, there is no guarantee there exists a solution that adheres to the set of provided objectives and constraints. As a result, when creating task instances, in SMDD-Bench, we create a \textit{witness-aware} task generation pipeline, allowing us to generate difficult, chemically valid, and guaranteed-solvable task instances while simultaneously generating an example solution molecule -- all without requiring a human expert in the loop.

We evaluate 7 frontier open and closed source LLMs using a minimalist agent scaffold on SMDD-Bench and find that, although LLMs achieve up to 50\% on simple lead optimization tasks, more challenging lead optimization tasks as well as other task types have near 0\% success rates -- highlighting significant room for growth for LLM applications as autonomous medicinal chemists. We also introduce \textbf{SMDD-Bench Diversity}, a subset of SMDD-Bench along with a principled framework for studying the diversity of SMDD agents' outputs.

We hope that the release of SMDD-Bench standardizes and invigorates research into training and evaluating LLMs for real-world autonomous drug discovery.

\section{SMDD-Bench}

SMDD-Bench consists of 502 instances of 5 task types: 2D Pharmacophore Identification (25 tasks), Interaction Point Discovery (25 tasks), Scaffold Hopping (52 tasks), Lead Optimization (340 tasks), and Fragment Assembly (60 tasks). Each task is generated in a \textbf{witness-aware} fashion, where a witness molecule that solves the task is simultaneously generated along with the starting molecule -- ensuring the task instance is guaranteed to be solvable and has at least one solution. Each task instance is also evaluated purely computationally, without a human in the loop, using tools such as RDKit, PLIP, OpenBabel, Boltz2, and ADMET-AI \citep{passaro2025boltz, admet_package, rdkit, salentin2015plip, oboyle2011openbabel}. Boltz2 is used for protein-ligand co-folded structure prediction, binding affinity\footnote{Note that Boltz2 predicts binding affinity as $log_{10}(\text{IC50 in }\unit{\micro\molar}$). As a result, a decrease in the Boltz2 predicted affinity value leads to stronger binding \citep{passaro2025boltz}}, and binding probability prediction while the ADMET-AI model is used for property prediction. An overview of the benchmark can be found in Figure \ref{fig:fig1}. 

\subsection{Task Types}
Further details on the generation and evaluation of each task type can be found in Appendix \ref{sec:task_details}. Only the \textit{Scaffold Hopping}, \textit{Lead Optimization}, and \textit{Fragment Assembly} tasks require \textbf{witness-aware} generation due to the \textit{2D Pharmacophore Identification} and \textit{Interaction Point Discovery} tasks having ground truth solutions by default. 

\noindent\textbf{1. 2D Pharmacophore Identification}\leavevmode\\
    \emph{Task:} Given a target protein sequence as well as the SMILES strings associated with 10 experimentally active molecules and 10 experimentally inactive molecules from the ChEMBL database \citep{chembl}, create a 2D pharmacophore in the form of a python function, which returns True if an inputted SMILES string matches the pharmacophore and False otherwise. See Appendix \ref{sec:task_type_1} for more details.
    
    \emph{Evaluation:} The python function is graded on its recall and specificity performance on a held-out set of hidden, experimentally validated actives and inactives. 
    
    \emph{Generation:} We filtered ChEMBL activity records for high-confidence, direct-binding assays, split the candidates into actives and inactives as determined by potency, and then pruned with MaxMin diversity picking to ensure the visible set spans chemical space and introduces binding mode diversity.  
    
    This task assesses the agent's ability to extract a generalizable, discriminative structural hypothesis from a small set of actives and inactives. In real world applications, the pharmacophore produced by the agent can be used as a first-filter for the screening of additional candidate compounds. 

\noindent\textbf{2. Interaction point discovery}\leavevmode\\
    \emph{Task:} Given a PDB file of the protein structure as well as the location of the pocket, specify the 3D coordinates of the 3 interaction points within a protein pocket most likely to be conserved across a diverse set of small molecule binders. The agent must output a csv file of 3D coordinates as well as a label for the type of interaction point. The types are chosen as donor, acceptor, aromatic, hydrophobic, cation or anion interaction points.
    
    \emph{Evaluation:} The predicted points are assigned to the ground-truth points via the Hungarian algorithm. Each prediction is filtered to a ground truth point whose type is compatible (exact match, or a permitted cross-type match, see Appendix \ref{sec:task_type_2} for more details), then the Hungarian algorithm computes the optimal one-to-one assignment under a distance cutoff. All three predictions must match a distinct conserved ground truth point in order for the task to be considered a success.
    
    \emph{Generation:} We determine ground truth points by fetching thousands of co-crystal structures per target from the PDB \citep{berman2000pdb}, aligning binding sites, extracting per-ligand protein interactions, and run density-based clustering with a conservation threshold. Steric clashes with the agent-provided prototype PDB are filtered out and ambiguous type labels are merged. As a result, it is possible for there to be more than three ground truth points of well-conserved interaction points. The agent only needs to output points that match to three of these ground truth points.  
    
    This task assesses the agent's ability to identify valuable hotspots within a protein pocket. In real world applications, these hotspots can then be targeted during fragment-based drug discovery campaigns, serve as guardrails during lead optimization, or used as 3D pharmacophores during pharmacophore-based virtual screening.

\noindent\textbf{3. Scaffold hopping}\leavevmode\\
    \emph{Task:} Given a Boltz2-folded protein and reference ligand complex, propose a new SMILES with a distinct scaffold that reproduces the reference ligand's binding interactions with the protein pocket. 
    
    \emph{Evaluation:} The submitted SMILES string must have a low Tanimoto similarity with the reference molecule. The Murcko scaffolds of the submitted and reference molecules must also have a maximum common substructure which is only a portion of the overall heavy atom count of the Murcko scaffold of the reference molecule. Both of these checks ensure that a proper scaffold hop has been performed.
    
    In order to check if the submitted molecule adheres to the same binding mode as the reference molecule, Boltz2 is used to co-fold the submitted SMILES string against the target using the same pocket contacts as the reference molecule. Potential stochasticity within the Boltz2 co-folded structure is mitigated by matching binding interactions via an ensemble of 10 Boltz2 co-folded structures.
    
    \emph{Witness-Aware Generation:} We first mine experimentally known actives for a specific pocket. Afterwards, we consider all potential pairs of actives. Every pair which the two ligands satisfy the Tanimoto distance and scaffold substructure constraint is co-folded via Boltz to the shared target. If both members of the pair share the same interaction fingerprint with the protein pocket and have high predicted binding probability, the pair and target are created into a task instance. One molecule within the pair is labeled as the reference while the other molecule is labeled as the witness. The witness molecule ensures the existence of at least one solution and guarantees the solvability of the task instance. 

    This task assesses the agent's ability to understand bioisosterisms as well as internally reason through binding poses before committing to making limited, computationally heavy oracle calls. In real world applications, scaffold hopping is used for rescuing lead molecules with problematic scaffolds, mitigating off-target binding effects, enabling the extension of SAR exploration beyond a depleted search, and avoiding IP infringement.

\noindent\textbf{4. Lead optimization}\leavevmode\\
    \emph{Task:} Given a reference molecule that is Boltz2 co-folded with a target protein, optimize the reference molecule into a new molecule that improves a set of ADMET and binding affinity objectives while holding other properties constant, satisfying drug-likeness constraints, maintaining high binding probability, and maintaining close Tanimoto similarity to the reference molecule (due to this being a lead optimization task).
    
    \emph{Evaluation:} The submitted molecule must satisfy the set of specified ADMET and binding affinity objectives. RDKit is used to measure drug-likeness, ADMET properties are calculated using the ADMET-AI library, and Boltz2 is used for binding probability and binding affinity prediction. 
    
    \emph{Witness-Aware Generation:} For each selected protein target, we build a pool of pocket-compatible molecules that pass drug-likeness filters and a Boltz2 binding probability gate. We consider all pairs of molecules from this pool that are similar via Tanimoto similarity. From these pairs, if there exists a molecule whose ADMET/binding affinity profile dominates the other molecule on some subset of the 8 properties in SMDD-Bench (found in Table \ref{tab:admet_properties}), the task's optimization objectives and hold-constant constraints are derived from the measured property differences between the two molecules. The dominating molecule becomes the witness while the dominated molecule becomes reference to provide to the agent to start lead optimization. The task instance is only instantiated if the witness's profile satisfies every objective and constraint. As a result, the witness solves the task instance by construction and the task becomes guaranteed solvable. More details on this can be found in Appendix \ref{sec:task_type_4}.

    This task assesses the agent's ability to reason through how certain edits may affect complicated, intertwined relationships between molecular properties. In real world applications, multi-property optimization with optimization objectives and hold-constant constraints define the core loop of any lead optimization campaign. Lead optimization is also where the bulk of time is spent in most drug discovery programs. We envision a setting where hundreds of thousands or millions of high-performing agents are instantiated in parallel with the task of discovering a diverse array of optimized molecules given a single starting reference molecule.

\noindent\textbf{5. Fragment Assembly}\leavevmode\\
    \emph{Task:} Given either one or two 3D molecule fragments placed in a protein pocket, design a single drug-like molecule that contains each fragment as a substructure, covalently links the fragments when more than one fragment is provided, achieves high binding probability and affinity, and preserves the input fragment poses when the submitted molecule is re-docked. 
    
    \emph{Evaluation:} The submitted molecule must incorporate each of the provided fragments as distinct substructures as well as pass drug-likeness checks. The submitted molecule is also co-folded with the protein via Boltz2 and must have a high predicted binding probability and binding affinity with respect to the witness molecule. Finally, the fragment substructures within the submitted molecule must be similar to the provided 3D coordinates of the input fragments after pocket alignment is performed. This is to ensure a molecule of the same binding mode as the input fragments was designed. Otherwise, it would be possible for a submitted molecule to contain the input fragments but have the fragments be in nonessential locations for binding. More details can be found in Appendix \ref{sec:task_type_5}.
    
    \emph{Witness-Aware Generation:} We co-fold known actives via Boltz2 and keep the co-folded structures which pass a binding probability gate. We perform BRICS decomposition \citep{brics} on the molecule to separate the molecule into its synthesizable fragments. Each input fragment must be large enough to be a meaningful pharmacophore, have at least one direct interaction with the protein pocket, and be in close proximity to the protein itself. Fragments are then sampled from the same molecule to create the task instance.  As a result, the co-folded active molecule in which fragments are sampled from becomes the witness molecule by construction -- ensuring the guaranteed solvability of the task instance. Task instances with 2 fragments have the additional constraint for fragments to be sufficiently far and not directly linked together.

    This task assesses the agent's ability to reason through 3D geometry and fragment-pocket complementarity. In real world applications, fragment assembly is the central step of fragment-based drug discovery campaigns, where weakly-binding fragments identified via crystallographic or biophysical screening must be elaborated or linked into a single high-affinity lead molecule.

\subsection{Benchmark Setup and Variations}
\label{sec:min_agent}
\vspace{-2ex}

\noindent\textbf{SMDD-Bench Agent Setup.} To evaluate the agent's innate medicinal chemistry reasoning ability, internet access is strictly prohibited, and agents have no access to academic literature, patent filings, textbooks, or any other source of non-parametric knowledge. We benchmark current frontier LLMs using a minimalist ReAct agent \citep{yao2022react} with a minimal harness, providing access to an isolated filesystem, a python interpreter with relevant biology and chemistry packages installed, and a limited number of Boltz2 and ADMET-AI oracle calls (the same oracles used to evaluate agent submissions). To minimize memorization, all identifying information such as PDB codes, ligand codes, Uniprot IDs, and target names are obfuscated or removed. More details on the minimalist agent can be found in Appendix \ref{sec:min_agent_appendix}.

\begin{figure}
  \centering
  \includegraphics[width=\textwidth, trim=60px 220px 110px 140px, clip]{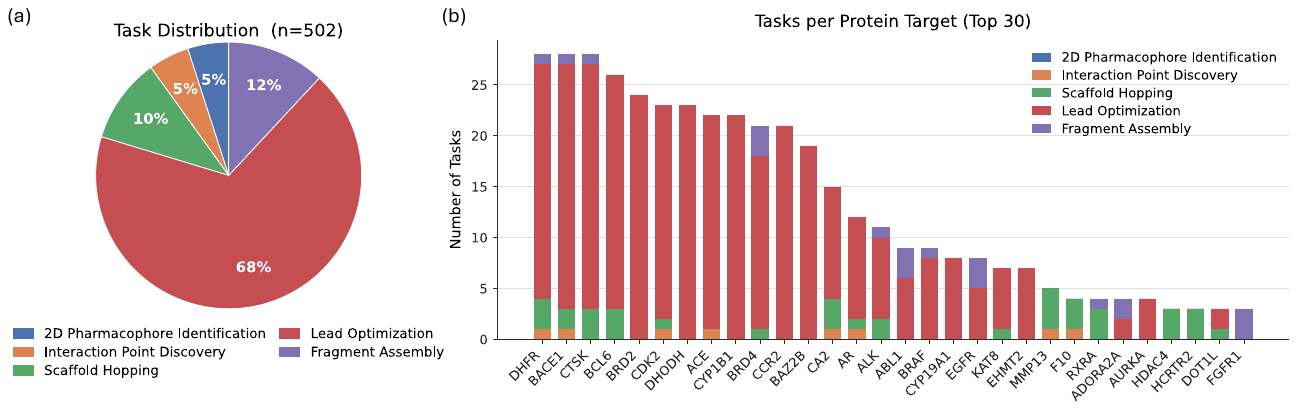}
  \caption{(a) The distribution of SMDD-Bench tasks across the five task types. Each task type is generated to be guaranteed-solvable by construction. (b). The top 30 most frequent protein targets found through all task types, broken down by task type.}
  \label{fig:benchmark_overview}
\end{figure}

\begin{figure}
    \centering
      \includegraphics[width=\textwidth, trim=60px 220px 60px 205px, clip]{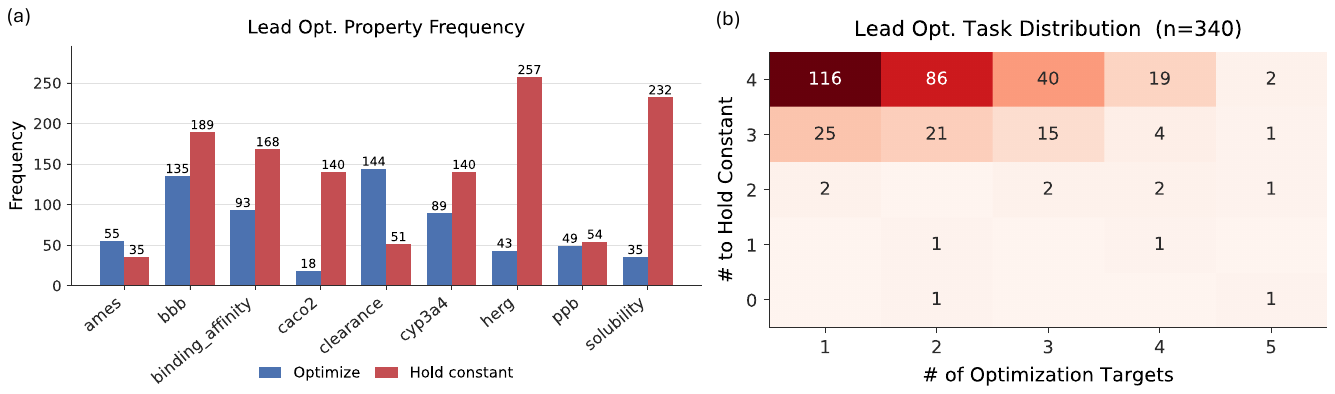}
      \caption{(a) Frequency with which each ADMET and binding affinity property appears as an optimization objective versus a hold-constant constraint across the 340 Lead Optimization instances. (b) Joint distribution of the number of optimization objectives and hold-constant properties across the 340 Lead Optimization task instances. Each cell reports the number of task instances with the given (hold-constant, optimization target) pair.}
      \vspace{-3ex}\label{fig:task4_overview}
\end{figure}

\noindent\textbf{Benchmark Statistics.} Across 502 task instances, SMDD-Bench contains 102 unique protein targets and 855 unique small molecules. Figure \ref{fig:benchmark_overview} shows the distribution of task types and protein targets across the benchmark. Figure \ref{fig:task4_overview}(a) and (b) report, respectively, the frequencies with which each of the 9 properties appears in Lead Optimization tasks and the joint distribution of task complexity. Additional details on benchmark diversity are provided in Appendix \ref{sec:detailed_benchmark_diversity}.

\noindent\textbf{SMDD-Bench Lite.} Due to the heavy computational overhead of running an agent on all 502 tasks, we release SMDD-Bench Lite, a performance-representative subset of 100 task instances designed to encourage broader adoption of the benchmark. SMDD-Bench Lite serves as a testbed for iterating on agent development against a difficulty-diverse and representative sample of the full benchmark. Details on how the SMDD-Bench Lite instances are selected can be found in Appendix~\ref{sec:smdd-bench-lite}.

\section{Results}
\vspace{-1ex}

\noindent\textbf{LLMs do not perform well on real-world small molecule drug discovery tasks.}
We evaluated 7 frontier LLMs on SMDD-Bench using the minimalist agent harness described in Section \ref{sec:min_agent} and present results in Table \ref{tab:overall_results}. As described in Appendix \ref{sec:min_agent_appendix}, each agent was provided with 8 Boltz2 oracle calls and 15 ADMET-AI oracle call -- each ADMET-AI oracle call returns all 8 predicted ADMET properties. Closed source models such as Claude Sonnet 4.6, GPT-5.4 and Gemini 3.1 pro significantly outperform leading open source models such as Kimi K2.5 Thinking, Deepseek V3.2, and Qwen 3.5-397B-A17B with GPT5.4 setting the state of the art at 40.2\% overall success rate.

However, most of the successes across all models are a result of performing well on the Lead Optimization task. Particularly, current LLMs perform poorly on tasks that require a heavy intuition and understanding of 3D geometries and interactions such as the Interaction Point Discovery, Scaffold Hopping and Fragment Assembly tasks. Considering the overwhelming challenges of multi-property, multi-constraint lead optimization, frontier LLMs perform surprisingly well at the Lead Optimization task. More specific results on LLM performance, such as cost breakdown and tool usage can be found in Appendix \ref{sec:additional_agent_results}.

\begin{table}[ht]
\centering
\footnotesize 
\setlength{\tabcolsep}{2.5pt} % Slightly tightened from 2.7pt
\caption{Success rate of 7 frontier closed and open source LLMs on SMDD-Bench, broken down by task type. Each LLM is evaluated using the minimalist agent harness outlined in Section \ref{sec:min_agent} and Appendix \ref{sec:min_agent_appendix} with a limited number of oracle calls.}
\label{tab:overall_results}
\begin{tabular}{l
                S[table-format=2.1]
                S[table-format=2.1]
                S[table-format=2.1]
                S[table-format=2.1]
                S[table-format=2.1]
                S[table-format=2.1]
                S[table-format=1.2]}
\toprule
& \multicolumn{6}{c}{Success Rate (\%)}  \\
\cmidrule(lr){2-7}
Model
& \multicolumn{1}{c}{\makecell{2D Pharma-\\cophore ID}}
& \multicolumn{1}{c}{\makecell{Interaction Point \\ Discovery}}
& \multicolumn{1}{c}{\makecell{Scaffold \\ Hopping}}
& \multicolumn{1}{c}{\makecell{Lead \\ Optimization}}
& \multicolumn{1}{c}{\makecell{Fragment \\ Assembly}}
& \multicolumn{1}{c}{\makecell{Overall}}
& \multicolumn{1}{c}{\makecell{Cost \\ (\$/instance)}} \\
\midrule
Claude Sonnet 4.6     & {\textbf{28.0}} & 0.0 & {\textbf{3.8}} & 53.5 & 0.0 & 38.0 & 1.31 \\
GPT-5.4               & 12.0 & 0.0 & {\textbf{3.8}} & {\textbf{57.6}} & {\textbf{1.7}} & {\textbf{40.2}} & 0.78 \\
Gemini 3.1 Pro        & 20.0 & {\textbf{4.0}} & 0.0 & 55.6 & {\textbf{1.7}} & 39.0 & 0.63 \\
Kimi K2.5 Thinking    & 12.0 & 0.0 & 1.9 & 43.5 & 0.0 & 30.3 & 0.40 \\
DeepSeek V3.2         & 8.0  & 0.0 & {\textbf{3.8}} & 34.7 & 0.0 & 24.3 & 0.43 \\
MiniMax M2.7          & 16.0 & 0.0 & 1.9 & 27.1 & 0.0 & 19.3 & {\textbf{0.36}} \\
Qwen3.5 397B A17B     & 4.0  & 0.0 & 1.9 & 40.0 & 0.0 & 27.5 & 0.75 \\
\bottomrule
\end{tabular}
\vspace{-2ex}
\end{table}

The chemical novelty of the submissions of each agent, broken down into task type, are found in Table \ref{tab:novelty_results}. Chemical novelty is defined as the \% of total agent submissions which are not found in the ChEMBL36, SureChEMBL, PubChem, of BindingDB databases \citep{chembl, papadatos2016surechembl, kim2023pubchem, liu2007bindingdb, liu2025bindingdb}. These databases are chosen due to their potential inclusion of bioactivity data for small molecules with respect to protein targets. The novelty check we prescribe is a conservative measure of the agent's ability to design a molecule in the context of a protein target for which there doesn't exist pre-existing training data on.

\begin{table}
\centering
\footnotesize
\caption{Novelty of agent submissions, defined by \% of submitted molecules not found in ChEMBL, PubChem, SureChEMBL, or BindingDB.}
\label{tab:novelty_results}
\setlength{\tabcolsep}{6pt}
\begin{tabular}{l
                S[table-format=2.1]
                S[table-format=2.1]
                S[table-format=2.1]}
\toprule
Model
& \multicolumn{1}{c}{\makecell{Scaffold\\Hopping}}
& \multicolumn{1}{c}{\makecell{Lead\\Optimization}}
& \multicolumn{1}{c}{\makecell{Fragment\\Assembly}} \\
\midrule
Claude Sonnet 4.6     & 88.5 & 77.9 & 82.8 \\
GPT-5.4               & {\textbf{94.2}} & 75.3 & 68.3 \\
Gemini 3.1 Pro        & 71.2 & 76.5 & {\textbf{83.3}} \\
Kimi K2.5 Thinking    & 80.8 & 77.6 & 65.0 \\
DeepSeek V3.2         & {\textbf{94.2}} & 83.2 & 69.1 \\
MiniMax M2.7          & 82.7 & {\textbf{86.5}} & 67.2 \\
Qwen3.5 397B A17B     & 82.7 & 77.9 & 61.7 \\
\bottomrule
\end{tabular}
\vspace{-2ex}
\end{table}

\noindent\textbf{LLMs are not diverse in their outputs.}
In real world applications, during lead optimization campaigns, researchers may want to instantiate many agents or agentic systems in parallel on the same task instance. The goal would be for the agents to propose a diverse and novel spread of successful solutions rather than converging on a single solution, allowing for a more robust exploration and discovery of lead molecules. In order to test for this behavior, we propose \textbf{SMDD-Bench Diversity}, a strict subset of 20 SMDD-Bench task instances for the Lead Optimization task type. The task instances represent the most challenging lead optimization tasks that are still solvable by current frontier LLMs. See Appendix \ref{sec:selecting-smdd-bench-diversity} for more details on selecting SMDD-Bench Diversity.

For each task on SMDD-Bench Diversity, we run each agent 10 times and report the distribution of novel and diverse molecules which pass the evaluation requirements. In order to measure diversity, we measure the average number of unique successful solutions per task instance, the average pairwise tanimoto similarity of successful solutions, the percentage of successful solutions which are novel (using the same criteria as in Table \ref{tab:novelty_results}), and the average number of successful solutions of each LLM on the subset. Results are found in Table \ref{tab:smdd-bench-diversity}. 

\begin{table}[ht]
\vspace{-2ex}
\centering
\footnotesize
\setlength{\tabcolsep}{5pt}
\caption{Statistics across 10 runs of each of the 20 task instances in the SMDD-Bench Diversity subset. Novelty is defined in Table \ref{tab:novelty_results}. "Avg. Successful" and "Avg. Unique and Successful" categories are out of 10 due to the 10 runs performed on each task instance.}
\label{tab:smdd-bench-diversity}
\begin{tabular}{l
                S[table-format=1.2]
                S[table-format=1.2]
                S[table-format=2.1]
                S[table-format=1.3]}
\toprule
Model
& {\makecell{Avg. \\ Successful}}
& {\makecell{Avg. Unique \\ and Successful}}
& {\makecell{Novel and \\ Successful (\%)}}
& {\makecell{Pairwise \\ Tanimoto Similarity}} \\
\midrule
Claude Sonnet 4.6     & \textbf{8.40} & 3.70 & \textbf{74.0} & 0.823 \\
GPT-5.4               & 7.90 & 2.75 & 64.6 & 0.863 \\
Gemini 3.1 Pro        & 8.00 & 4.00 & 67.6 & 0.809 \\
Kimi K2.5 Thinking    & 6.00 & 3.85 & 65.0 & 0.786 \\
DeepSeek V3.2         & 5.35 & 3.85 & 68.4 & \textbf{0.763} \\
MiniMax M2.7          & 6.00 & \textbf{4.05} & 73.1 & \textbf{0.763} \\
Qwen3.5 397B A17B     & 7.25 & 3.55 & 67.2 & 0.814 \\
\bottomrule
\end{tabular}
\vspace{-1ex}
\end{table}

Although Claude Sonnet 4.6 consistently achieves the highest number of successful submissions, the model converges to the same set of molecules over half the time. Surprisingly, Minimax M2.7 submits the most unique spread of successful molecules on SMDD-Diversity while maintaining strong \% novelty across successful submissions. SMDD agents deployed in the wild must consistently achieve a high number of both unique and novel successful solutions when provided the same task instance.

\noindent\textbf{Enumeration vs. Selection.}
Throughout the solve process, each LLM enumerates many potential candidates, reasons through those candidates, and then runs the most promising of the candidates through an oracle tool call. We investigate whether or not LLMs are capable of enumerating passing molecules but fail at the reasoning and selection step. In order to do so, we use GPT5.4 mini to extract every SMILES string mentioned in each of the agent's traces for the Scaffold Hopping, Lead Optimization, and Fragment Assembly tasks. We then, using the task instance's evaluation pipeline, evaluate all of the molecules that the agent did not perform an oracle call on -- this set of molecules represents the molecules which the agent hypothesized could be strong candidates, but considered to be inferior. We then calculate the overall recovered success rate on SMDD-Bench had the LLM selected the best candidate from its enumeration and report the values in Table \ref{tab:what-if-results}. More details on how we performed this experiment can be found in Appendix \ref{sec:what_if_details}.

\begin{table}[ht]
\vspace{-1ex}
\centering
\footnotesize
% Slightly tightened the padding
\setlength{\tabcolsep}{2.5pt}
\caption{Original vs recovered success rates (\%) and total number of enumerated molecules (\#Mols) for all agents on SMDD-Bench. The recovered success rate is the  rate the agent would have if it had selected to submit the ground truth best molecule that it enumerated during the reasoning process.}
\label{tab:what-if-results}
% Requires \usepackage{graphicx} in your preamble
\resizebox{\textwidth}{!}{%
\begin{tabular}{l
                S[table-format=2.1]
                S[table-format=2.1]
                S[table-format=4.0]
                S[table-format=2.1]
                S[table-format=2.1]
                S[table-format=4.0]
                S[table-format=1.1]
                S[table-format=1.1]
                S[table-format=4.0]
                S[table-format=2.1]
                S[table-format=2.1]
                S[table-format=5.0]}
\toprule
& \multicolumn{3}{c}{\makecell{Scaffold \\ Hopping}}
& \multicolumn{3}{c}{\makecell{Lead \\ Optimization}}
& \multicolumn{3}{c}{\makecell{Fragment \\ Assembly}}
& \multicolumn{3}{c}{Overall} \\
\cmidrule(lr){2-4} \cmidrule(lr){5-7} \cmidrule(lr){8-10} \cmidrule(lr){11-13}
Model
& {Orig. (\%)} & {Recov. (\%)} & {\#Mols}
& {Orig. (\%)} & {Recov. (\%)} & {\#Mols}
& {Orig. (\%)} & {Recov. (\%)} & {\#Mols}
& {Orig. (\%)} & {Recov. (\%)} & {\#Mols} \\
\midrule
Claude Sonnet 4.6  & 3.8 & 32.7 & 2657 & 53.5 & 64.1 & 5872 & 0.0 & 3.3 & 4982 & 40.7 & 52.4 & 13511 \\
DeepSeek V3.2      & 3.8 & 11.5 & 1052 & 34.7 & 39.4 & 3413 & 0.0 & 0.0 & 1358 & 26.5 & 31.0 & 5823  \\
Gemini 3.1 Pro     & 0.0 & 19.2 & 735  & 55.6 & 62.9 & 3328 & 1.7 & 3.3 & 1855 & 42.0 & 50.0 & 5918  \\
GPT-5.4            & 3.8 & 15.4 & 850  & 57.6 & 61.2 & 4486 & 1.7 & 1.7 & 1945 & 44.0 & 48.0 & 7281  \\
Kimi K2.5 Thinking & 1.9 & 3.8  & 986  & 43.5 & 53.8 & 4940 & 0.0 & 1.7 & 2528 & 33.0 & 41.2 & 8454  \\
MiniMax M2.7       & 1.9 & 5.8  & 1203 & 27.1 & 47.1 & 7231 & 0.0 & 0.0 & 2372 & 20.6 & 36.1 & 10806 \\
Qwen3.5 397B       & 1.9 & 3.8  & 1055 & 40.0 & 53.8 & 4367 & 0.0 & 0.0 & 1915 & 30.3 & 40.9 & 7337  \\
\bottomrule
\end{tabular}
}
\end{table}
\vspace{-1ex}

Scaffold hopping sees a significant increase in recovered rate for Claude Sonnet 4.6, Gemini 3.1 Pro, and GPT5.4 -- indicating that the agents are not as capable at reasoning through binding interactions of chemically different analogues without first performing oracle calls. On the other hand, fragment assembly, a task requiring a deep understanding of how the binding pocket and surrounding chemistry affects the 3D geometry of the small molecule binding pose, sees very minimal overall improvement. Interestingly, MiniMax M2.7 nearly doubles its success rate for the Lead Optimization task, indicating its ability to enumerate solutions and inability to select from the enumerations.

\noindent\textbf{SMDD-Bench Lite performance.}
The performance of each agent on the SMDD-Bench Lite subset are reported in Table \ref{tab:lite-success-rates}. We also report the performance of Gemini 3.1 pro on high thinking mode.

\begin{table}[ht]
\vspace{-2ex}
\centering
\footnotesize
\setlength{\tabcolsep}{3.2pt}
\caption{Success rate (\%) by task type and overall cost per instance of 8 frontier LLMs on the SMDD-Bench Lite subset.}
\label{tab:lite-success-rates}
\begin{tabular}{l
                S[table-format=2.1]
                S[table-format=2.1]
                S[table-format=2.1]
                S[table-format=2.1]
                S[table-format=2.1]
                S[table-format=2.1]
                S[table-format=2.2]}
\toprule
& \multicolumn{6}{c}{Success Rate (\%)} & \multicolumn{1}{c}{Cost} \\
\cmidrule(lr){2-7} \cmidrule(lr){8-8}
Model
& \multicolumn{1}{c}{\makecell{2D Pharma-\\cophore ID}}
& \multicolumn{1}{c}{\makecell{Interaction Point \\ Discovery}}
& \multicolumn{1}{c}{\makecell{Scaffold \\ Hopping}}
& \multicolumn{1}{c}{\makecell{Lead \\ Optimization}}
& \multicolumn{1}{c}{\makecell{Fragment \\ Assembly}}
& \multicolumn{1}{c}{\makecell{Overall}}
& \multicolumn{1}{c}{\makecell{Cost \\ (\$/instance)}} \\
\midrule
Claude Sonnet 4.6        & 40.0 & 0.0 & 0.0 & 52.9 & 0.0 & 38.0 & 1.29 \\
GPT-5.4   (medium)       & 0.0  & 0.0 & 0.0 & 54.4 & 0.0 & 37.0 & 0.77 \\
Gemini 3.1 Pro (medium)  & 20.0 & 0.0 & 0.0 & 54.4 & 0.0 & 38.0 & 0.57 \\
Gemini 3.1 Pro (high)    & 0.0  & 0.0 & 20.0 & 52.9 & 8.3 & 39.0 & 0.46 \\
Kimi K2.5 Thinking       & 0.0  & 0.0 & 0.0 & 42.6 & 0.0 & 29.0 & 0.38 \\
DeepSeek V3.2            & 0.0  & 0.0 & 10.0 & 33.8 & 0.0 & 24.0 & 0.40 \\
MiniMax M2.7             & 20.0 & 0.0 & 0.0 & 25.0 & 0.0 & 18.0 & 0.34 \\
Qwen3.5 397B A17B        & 0.0  & 0.0 & 0.0 & 38.2 & 0.0 & 26.0 & 0.79 \\
\bottomrule
\end{tabular}
\vspace{-2ex}
\end{table}

\vspace{-1ex}
\section{Common Failure Modes}
\vspace{-1ex}
Beyond the issues regarding the novelty rate, diversity rate, and enumeration vs. selection performance gap, we investigated individual traces searching for other common failure modes in the frontier LLMs. 

\textbf{Absence of Cross-Turn Structure–Activity Relationship (SAR) Synthesis.} A recurring failure across models is their inability to generalize SAR rules. When a molecule violates predefined constraints, a budget-aware agent should ask whether a specific structural feature caused this failure and apply this exclusion rule to later candidates. Instead, agents repeatedly test variants of already disqualified structures, effectively wasting budget on confirming the same rule multiple times. This reflects a lack of meta-cognitive reasoning: agents fail to convert individual failures into structural generalizations that could narrow down the search space. We provide an annotated reasoning trace in  Appendix~\ref{traj:tt2:failure_mode_no_sar_synthesis}.

\textbf{Incoherent planning across multiple turns.} We notice that a common failure mode involves candidate-generation turns being treated independently rather than as part of an iterative, sequential process. This results in the agent generating new molecules instead of selecting from candidates that passed earlier-stage criteria. Furthermore, the same molecule may be proposed multiple times across turns even though it had already been tested in a previous turn and did not meet proper thresholds. Refer to a representative trace in Appendix~\ref{traj:tt3:failure_mode_repeated_proposal}.

\textbf{Code Execution Errors.} Agents also exhibit tool-specific coding mistakes. These include malformed calls to structure-prediction tools (e.g., incorrect argument types or missing required fields for Boltz binding predictions, cheminformatics rountines (e.g., \texttt{MolToSmiles} conversion), and interaction-analysis pipelines (e.g., PLIP). 

\vspace{-1ex}
\section{Related Work}
\vspace{-1ex}
SMDD-Bench evaluates LLM agents on challenging, real-world small molecule drug design tasks. Existing chemistry and molecular benchmarks -- ChemBench \citep{chembench}, MolecularIQ \citep{bartmann2026moleculariq}, SmolInstruct \citep{yu2024llasmol}, QCBench \citep{xie2025qcbench}, and MuMOInstruct \citep{dey2025gellm3o} -- are restricted to single-turn question answering on relatively simple chemistry tasks and do not evaluate complex agentic systems on real-world problems. Prior work on multi-turn iterative molecular optimization \citep{chennakesavalu2026prescientdesignevals, wang2024efficient} likewise lacks realistic agentic tool calling, instead querying the LLM repeatedly in a loop over a small number of simple optimization objectives. In contrast, harnessing LLM agents as molecular designers lets them use tools to understand the 3D problem setting, analyze previous failed hypotheses, and run simulations and other cheminformatics tools before committing to oracle calls. Iterative prompting approaches are not faithful representations of the true potential for LLM agents in drug discovery.

Beyond single-turn QA, prior LLM agents for drug design and discovery \citep{ock2026agentd, li2026agentic, li2026rl_tdc, zhang2024agentsmddsurvey, wang2024efficient} either report results on a few hand-selected successes or use ad hoc, non-standardized evaluation setups, often retrofitting datasets built for de novo molecular generation such as ZINC, Therapeutics Data Commons, GuacaMol, PMO, and MOSES \citep{tingle2023zinc, huang2021therapeutics, brown2019guacamol, gao2022sample, polykovskiy2020molecular}. Adjacent domains like computational biology have produced agentic benchmarks such as DeepResearch Bench \citep{du2025deepresearch}, BioProBench \citep{liu2025bioprobench}, BioLP-Bench \citep{ivanov2024biolp}, ABC-Bench \citep{liu2025abc}, Lab-Bench \citep{laurent2024lab}, BixBench \citep{mitchener2025bixbench}, and CompBioBench \citep{nair2026compbiobench}, but no prior work evaluates LLMs on agentic small molecule drug design.

The primary contributions of SMDD-Bench compared to previous work are as follows: (1) the first standardized benchmark for evaluating long-horizon, multi-turn, small molecule drug design \textbf{agents} on a chemically diverse array of challenging drug discovery tasks; (2) support for five relevant, real-world task types: 2D Pharmacophore Identification, Interaction Point Discovery, Scaffold Hopping, Lead Optimization, and Fragment Assembly; and (3) guaranteed solvability for all tasks, including the Scaffold Hopping, Lead Optimization, and Fragment Assembly tasks via the existence of a held-out witness molecule which passes all evaluation requirements.

\section{Discussion and Conclusion}
\label{sec:discussion}
\vspace{-1ex}
We present SMDD-Bench, the first agentic benchmark evaluating LLM agents on challenging, long-horizon, real-world small molecule drug design. SMDD-Bench tests scientific and chemical reasoning across diverse chemistries, task types, and protein targets, alongside domain-specific tool-calling capabilities, with every task instance generated to be guaranteed-solvable. Our hope is that SMDD-Bench unifies and standardizes the design, training, and evaluation of LLMs for small molecule drug design.
The dominant approach to small molecule drug design today is conditional generative diffusion and flow models \citep{sun2025gensmddreview, zeng2026propmolflow, zhang2025ecloudgen, morehead2026zatom, klein2023equivariant, hoogeboom2022equivariant}, but LLM-based SMDD agents offer distinct advantages in flexibility and generality. An agent can perform lead optimization on properties unseen at training time, understand the failure modes of specific oracles, selecting the right tool for a given task, and adhere to constraints (such as maintaining protein-protein contacts when designing a PROTAC) that were never specified during training. Communicating objectives in natural language further frees the agent from the rigid train-time settings that bind diffusion and flow models.

\noindent\textbf{Reliance on Boltz2 and ADMET-AI.} Boltz2 and ADMET-AI are imperfect oracles. However, in SMDD-Bench we treat these oracles as the ground truth within the "world" of the benchmark, analogous to an agent requesting wet lab assays. As a result, we deliberately provide a limited number of oracle calls to reflect the sparsity of real-world wet lab experiments. To enforce that the Boltz2 and ADMET-AI predicted distributions are the operative ground truth, all PDB-sourced complexes are re-folded via Boltz2 and re-scored by ADMET-AI during task generation, so the agent only sees structure-activity relationships and property distributions produced by these oracles.
SMDD-Bench is intended as a testbed for developing training frameworks calibrated to the Boltz2 and ADMET-AI distributions. The same frameworks can later be interfaced with an autonomous laboratory, where the oracle becomes a real wet lab that synthesizes, characterizes, and tests candidates, \textbf{allowing agents to be re-trained under the true distribution of real-world structure-activity relationships}.

\begin{ack}
We are especially grateful for support from the Google Cloud Platform for Gemini credits, allowing us to perform in-depth analyses, comprehensive benchmarking and fast iteration. We also thank Thinking Machines, for access to their Tinker platform through the Tinker Research Grant program as well as Modal for access to compute. Furthemore, we appreciate Ben Miller, Aviral Kumar, Max Simchowitz, and Divy Thakkar for insightful conversations and feedback.
\end{ack}

\bibliographystyle{plainnat}
\bibliography{references}

%%%%%%%%%%%%%%%%%%%%%%%%%%%%%%%%%%%%%%%%%%%%%%%%%%%%%%%%%%%%
\newpage
\appendix

\addtocontents{toc}{\protect\setcounter{tocdepth}{2}}

\tableofcontents
\newpage 

\section{Detailed Benchmark Diversity}
\begin{figure}
    \centering
      \includegraphics[height=0.9\textheight, clip]{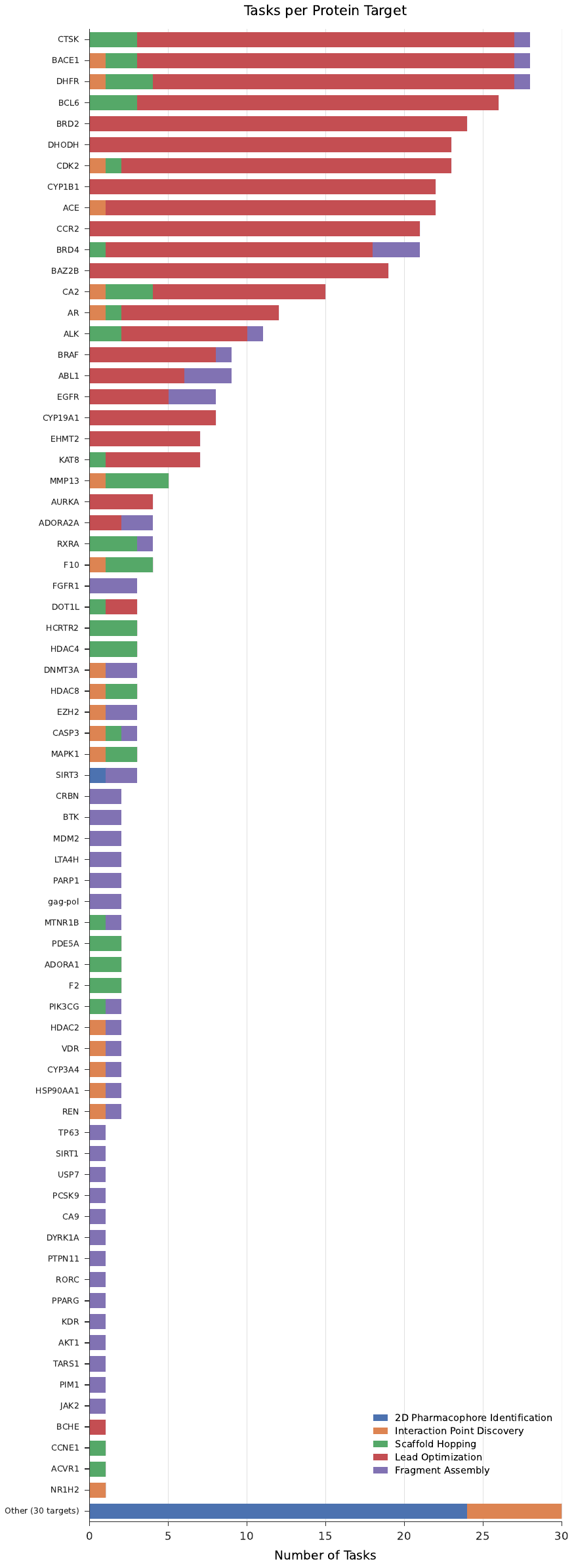}
      \caption{The complete breakdown of task instances into protein targets. There are 102 unique protein targets in SMDD-Bench.}
      \label{fig:full_target_list}
\end{figure}

\begin{figure}
    \centering
      \includegraphics[width=\textwidth, clip]{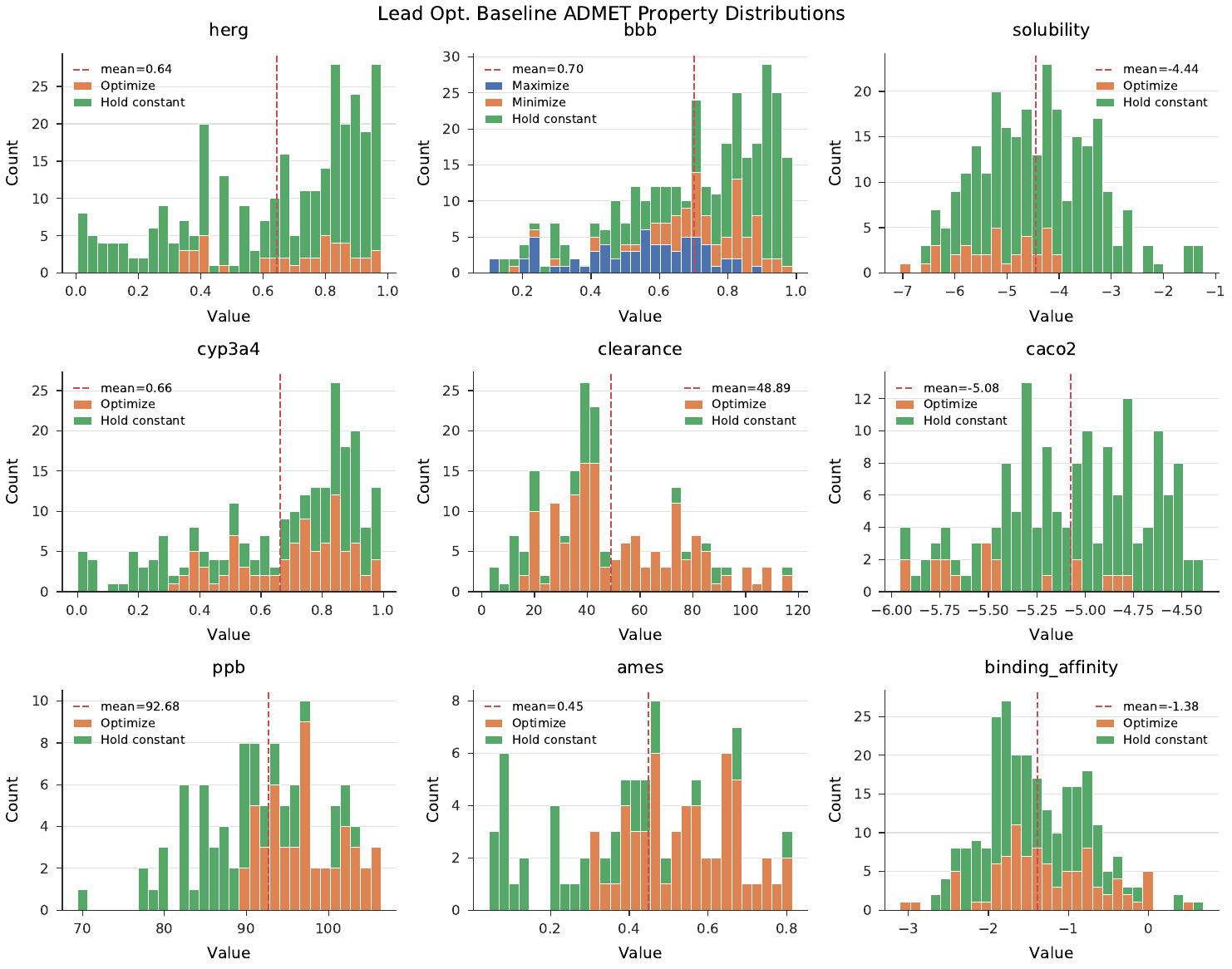}
      \caption{The baseline ADMET property values for all optimization objectives and hold-constant properties for the Lead Optimization task type. These are the values calculated for the reference molecules that the LLM agent must either optimize or hold constant. Note that blood brain barrier penetration (bbb) can either be a property that is minimized or maximized. All other properties are either only minimized or only maximized.}
      \label{fig:task4_baseline_properties}
\end{figure}

The complete list of the 102 unique protein targets broken down into task types can be found in Figure \ref{fig:full_target_list}. In Figure \ref{fig:task4_baseline_properties}, we plot each of the reference molecule baseline property values for optimization properties and hold-constant properties in the Lead Optimization task. Blood brain barrier penetration can either be optimized to enhance penetration or to decrease penetration depending on the real world application. Both directions of optimization are included in SMDD-Bench. 

\begin{figure}
    \centering
      \includegraphics[width=0.7\textwidth, clip]{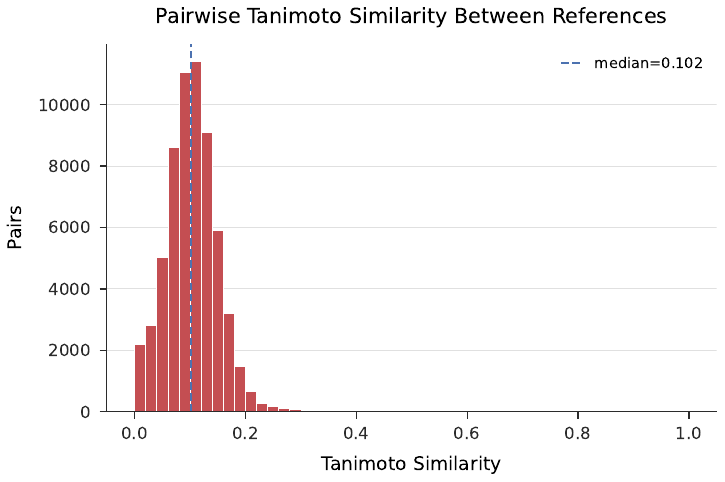}
      \caption{A histogram of the tanimoto similarities between all pairs of reference molecules provided to the LLM agent in the Scaffold Hopping and Lead Optimization tasks. The low median tanimoto similarity between pairs indicates the chemical diversity of SMDD-Bench.}
      \label{fig:reference_tanimoto}
\end{figure}

In Figure \ref{fig:reference_tanimoto}, we plot the pairwise Tanimoto similarity between all pairs of reference molecules in the Scaffold Hopping and Lead Optimization tasks. The low distribution of Tanimoto similarity indicates that the bulk of the reference molecules aren't collapsed to similar points in chemical space -- highlighting the small molecule diversity of SMDD-Bench. 

\begin{figure}
    \centering
      \includegraphics[width=0.7\textwidth, clip]{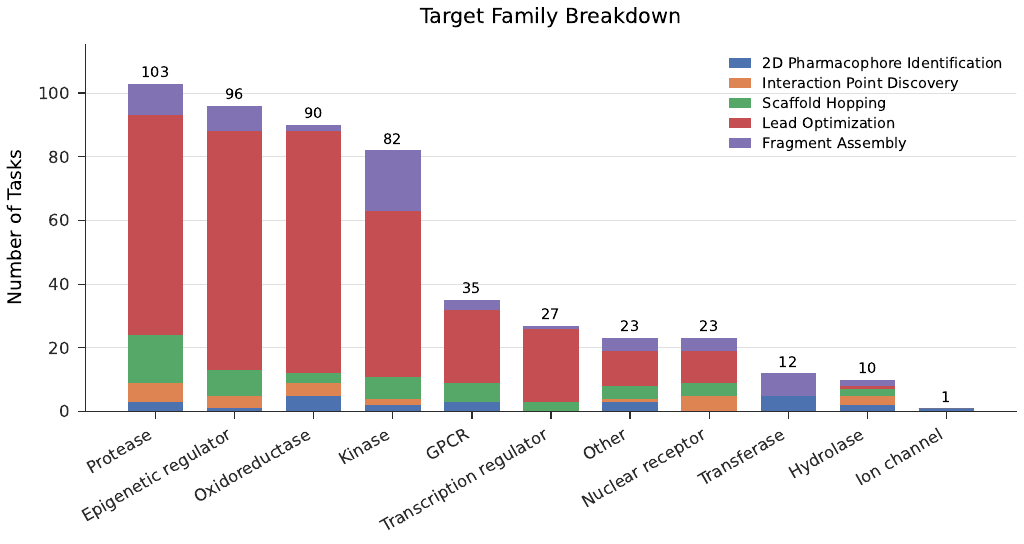}
      \caption{A breakdown of the task types of SMDD-Bench into the families of the protein targets associated with each task instance.}
      \label{fig:target_family_breakdown}
\end{figure}

Figure \ref{fig:target_family_breakdown} shows the breakdown of the protein targets within each task instance into their respective families, highlighting the protein diversity of SMDD-Bench. For an agent to be successful, it must be able to generalize and reason across diverse chemistries, pockets, and biological targets.

\label{sec:detailed_benchmark_diversity}

\section{Details on Task Type Generation}
\label{sec:task_details}
In the following section, we provide the specific details for generating the task instances for each task type. 

\subsection{2D Pharmacophore Identification}
\label{sec:task_type_1}

Each instance of this task type asks the agent to identify a 2D pharmacophore that distinguishes a set of active molecules from a set of inactive molecules for a given protein target. Tasks are generated procedurally from ChEMBL.

\noindent\textbf{Activity data extraction}
We restrict to direct binding assays by requiring \texttt{standard\_type} $\in \{\text{IC}_{50}, K_i\}$, \texttt{standard\_units}~$=$~nM, \texttt{standard\_relation}~$=$~`$=$', and \texttt{confidence\_score}~$\geq 8$. Only targets with \texttt{target\_type}~$=$~`SINGLE PROTEIN' are considered.

\noindent\textbf{Compound normalization and deduplication.}
SMILES are canonicalized via RDKit. We discard multi-component salts/mixtures, molecules with exact molecular weight greater than 900~Da, and entries that fail RDKit parsing. Multiple measurements of the same canonical compound are collapsed by taking the per-compound median activity in nM.

\noindent\textbf{Active/inactive labeling}
Compounds with median activity below 1{,}000~nM (IC$_{50}$/$K_i < 1~\mu$M) are labeled active; compounds with median activity above 10{,}000~nM ($> 10~\mu$M) are labeled inactive. Compounds in the 1--10~$\mu$M ambiguous zone are discarded, providing a $10\times$ potency gap between the two classes that excludes borderline molecules. Targets with fewer than 30 actives or fewer than 30 inactives after filtering are not eligible.

\noindent\textbf{Diverse subset selection}
For each class, ten visible exemplars are selected via the MaxMin algorithm operating on Morgan fingerprints (radius~$=2$, 2048~bits, seed~$=42$) \citep{ashton2002maxmin}. The remaining compounds form a held-out evaluation set that the agent never sees. MaxMin selection forces the visible exemplars to span chemical space, allowing the agent to see the most chemically diverse spread of actives as possible.

\noindent\textbf{Agent inputs and outputs}
Each generated task directory contains: \texttt{actives.smi} and \texttt{inactives.smi} (the 10 visible exemplars per class, given to the agent), \texttt{protein.fasta} (the target sequence). The agent is required to emit \texttt{solution.py} defining \texttt{check\_pharmacophore(smiles:~str)~->~bool}, which is run on every molecule of the held-out evaluation set during scoring.

\noindent\textbf{Design rationale}
Using ChEMBL-measured inactives rather than synthetically constructed decoys (e.g.\ DUD-E) ensures that negatives are experimentally verified non-binders that may share scaffolds with the actives, ruling out solutions that rely on coarse topological dissimilarity. Providing both classes to the agent and requiring generalization to a held-out set forces extraction of a genuine substructural pattern rather than memorization of the visible exemplars.

\subsection{Interaction Point Discovery}
\label{sec:task_type_2}

Each instance of this task type asks the agent to predict, given a single apo-like protein structure and a pocket center, the 3D coordinates and pharmacophoric type of the three interaction points most likely to be conserved across diverse binders of that pocket. Tasks are generated procedurally from the RCSB PDB \citep{berman2000pdb}.

\noindent\textbf{Ensemble construction}
For each specified target from the PDB, we fetch up to 500 unique X-ray crystal structures with resolution better than 2.5~\AA. The first structure containing a drug-like ligand is taken as the \emph{reference}; common crystallization additives (glycerol, PEG, DMSO, sulfate, and similar) are explicitly blacklisted. Within that reference structure, the largest non-blacklisted ligand by heavy-atom count is selected as the reference ligand. The reference receptor is stripped of all heteroatoms, waters, and ions to produce \texttt{receptor.pdb}, and the centroid of the reference ligand defines the pocket center provided to the agent.

\noindent\textbf{Local structural alignment}
All ensemble members are superimposed onto the reference using only the C$_\alpha$ atoms of residues within 12~\AA{} of the pocket center. Restricting the alignment to the local pocket suppresses contributions from global protein motion and induced fit -- minimizing potential global variance in the 3D reference frame.

\noindent\textbf{Pharmacophore feature extraction}
For each aligned structure, we select the ligand whose centroid lies closest to the reference pocket center (within 12~\AA); waters, the same blacklist of crystallization additives and common ions used during reference selection, and small fragments (fewer than six heavy atoms) are excluded from candidacy. RDKit is then used to assign each atom or functional group of the chosen ligand to one of six pharmacophoric families: hydrogen-bond donor, hydrogen-bond acceptor, aromatic, hydrophobic, cation, or anion. Features further than 12~\AA{} from the pocket center are discarded to exclude allosteric binders and crystallographic artifacts.

\noindent\textbf{Consensus clustering}
Clustering proceeds in two stages. First, DBSCAN \citep{ester1996dbscan} is run independently per pharmacophore family at $\varepsilon = 1.0$~\AA{} on the 3D points of that family alone, so that, e.g., donor and acceptor densities never collapse into a single cluster regardless of spatial proximity. A cluster is retained only if it is supported by at least 15\% of the unique structures contributing to the ensemble; the conservation score of a cluster is the fraction of unique PDB entries supporting it. Cross-family relationships between surviving clusters are handled separately during ground-truth point selection.

\noindent\textbf{Ground-truth point selection}
Surviving clusters are post-processed to yield a physically valid, non-redundant ground-truth set: (i) clusters within 2~\AA{} of any heavy atom of the reference receptor are discarded as sterically clashing with the reference protein conformation given to the agent; (ii) clusters of different types (e.g.\ donor and cation arising from a basic amine, or donor and acceptor from a hydroxyl) whose centroids lie within 2~\AA{} are merged into a single point with a set-valued type label, reflecting and allowing for chemical ambiguity in the agent output. The full set of $N \geq 3$ surviving consensus points constitutes the ground truth.

\noindent\textbf{Agent inputs and outputs}
The agent is provided \texttt{receptor.pdb} (the reference protein with all ligands and solvent removed) the pocket center coordinates and the 12~\AA{} pocket radius where all interaction points are within. The agent must emit \texttt{solution.csv} with exactly three rows of the form \texttt{x, y, z, type}, where \texttt{type} is drawn from \{\texttt{Donor}, \texttt{Acceptor}, \texttt{Aromatic}, \texttt{Hydrophobic}, \texttt{Cation}, \texttt{Anion}\}.

\noindent\textbf{Evaluation}
Predicted points are matched against the $N$ ground-truth points by optimal bipartite assignment \citep{kuhn1955hungarianalgorithm} on Euclidean distance, restricted to type-compatible pairs: a prediction is type-compatible with a ground-truth point if its label is in the ground-truth label set, with an additional aromatic/hydrophobic equivalence at scoring time, reflecting that aromatic ring features and hydrophobic features are partially co-tagged by RDKit and that the ground-truth construction may assign only one of the two labels at a position where both are chemically reasonable. This choice was made out of evaluation leniency. A match is counted if the assigned distance is below 2.5~\AA. The task score is (number of matched predictions)~$/~3$, and the task is considered solved only when all three predictions match (score $= 1.0$).

\noindent\textbf{Design rationale}
The local-pocket alignment, conservation threshold, and steric clash filter were designed to ensure that the targets are simultaneously experimentally grounded, chemotype-agnostic, empirically solvable, and medicinally useful in the specific receptor conformation handed to the agent.

\subsection{Scaffold Hopping}
\label{sec:task_type_3}

Each instance of this task type asks the agent to propose a novel small molecule that binds the same protein pocket as a given reference ligand and reproduces its protein--ligand interaction pattern, while presenting a chemically distinct scaffold. Tasks are generated procedurally from the RCSB PDB and ChEMBL, and every task is constructed around an explicit \emph{witness} molecule that is known and evaluated a priori to satisfy the evaluation criteria -- thus guaranteeing the solvability of the task instance.

\noindent\textbf{Target and canonical structure selection}
Targets are sourced from a curated pool of $\sim$300 \texttt{(UniProt, PDB, ligand)} triplets spanning diverse protein families, grouped by UniProt identifier. For each UniProt, the PDB entry with the best X-ray resolution is chosen as the canonical receptor; its FASTA sequence and a single fixed pocket definition are reused throughout downstream steps.

\noindent\textbf{Pocket definition}
The binding pocket is defined as the set of residues with any heavy atom within 6~\AA{} of the canonical ligand. This pocket, expressed as \texttt{(chain, SEQRES position)} pairs, is held fixed for all downstream Boltz2 cofolding runs and is also handed to the agent as part of the task specification.

\noindent\textbf{Candidate ligand collection}
For each target we assemble a candidate ligand pool from two sources: (i) all ligands co-crystallized with the same UniProt across the PDB, with bond orders corrected via Chemical Component Dictionary (CCD) templates and deduplicated by canonical SMILES; and (ii) ChEMBL actives against the target with reported IC$_{50}$/$K_i$/$K_d$ below 10~$\mu$M, pre-filtered to remove unparseable structures, molecules above 1{,}000~Da, and metal-containing species.

\noindent\textbf{Scaffold-diversity pre-filter}
Bemis-Murcko scaffolds are extracted for every candidate, and for each unordered ligand pair we compute the Maximum Common Substructure (MCS) of the two scaffolds \citep{bemis1996bemismurckoscaffold}. A pair is retained only if the MCS covers less than 65\% of the heavy atoms of the larger scaffold; the symmetric denominator ensures the filter does not depend on the eventual reference/witness role assignment.

\noindent\textbf{Crystal-PLIP pre-ranking}
For pairs in which both members appear as PDB co-crystals, each source structure is superimposed onto the canonical receptor using C$_\alpha$ atoms of pocket residues within 12~\AA, and a protein-ligand interaction fingerprint is extracted with PLIP \citep{salentin2015plip}. Pairs are then ordered by descending crystal interaction-fingerprint similarity, with PDB-PDB pairs prioritized over pairs involving ChEMBL-only ligands. This ranking concentrates downstream compute on pairs that are most likely to clear the interaction-similarity gate during the co-folding screen.

\noindent\textbf{Co-folding screen}
For each target, the top-ranked unique candidate ligands are co-folded with the canonical receptor using Boltz2 \citep{passaro2025boltz}, constrained to the fixed pocket definition with 10 diffusion samples per ligand. From the resulting ensemble of 10 conformations, we extract: (i) the Boltz2-predicted binding probability across samples; (ii) per-sample PLIP interaction fingerprints, where each interaction is coarsened to a \texttt{(chain, residue, broad type)} tuple over six interaction families (HBond, Aromatic, Ionic, Halogen, Metal, Hydrophobic); and (iii) a consensus interaction fingerprint defined as the set of tuples present in at least 60\% of the diffusion samples. Any ligand whose predicted binding probability falls below 0.75 is discarded. The consensus interaction fingerprint is required during evaluation to offset variance induced by Boltz2-predicted binding poses.

\noindent\textbf{Witness validation and role assignment}
A pair is promoted to a task only if both members survive the binding gate \emph{and} the set Tanimoto between their consensus interaction fingerprints exceeds 0.75. Within a surviving pair, the ligand with the higher predicted binding probability is assigned as the \emph{reference} (the molecule provided to the agent to hop away from) and the other becomes the witness (the hidden molecule whose existence guarantees the solvability of the task instance). We retain at most three tasks per UniProt target and require distinct reference molecules across tasks of the same target, preventing one well-behaved reference from dominating the benchmark.

\noindent\textbf{Agent inputs and outputs}
The agent is given: \texttt{complex.pdb} (the Boltz2-predicted reference complex, with the ligand on chain L), \texttt{reference.sdf} (the reference ligand conformer extracted from the Boltz2 complex), \texttt{protein.fasta} (the canonical protein sequence), and the target ligand identifier and pocket contact residue list used during the task generation process. The agent is required to emit \texttt{solution.smi} containing a single SMILES string. The witness molecule is never exposed to the agent.

\noindent\textbf{Evaluation}
A submission must clear five gates: (1) the SMILES is parseable by RDKit; (2) the Morgan fingerprint Tanimoto (radius~$=2$, 2048~bits) between the agent's molecule and the reference is below 0.5; (3) the Bemis--Murcko scaffold MCS, normalized by the heavy-atom count of the larger of the two scaffolds, is below 0.65; (4) the agent's molecule, co-folded under the same Boltz protocol used during generation (10 diffusion samples, identical pocket constraint), produces a consensus interaction fingerprint whose set Tanimoto with the reference's consensus fingerprint exceeds 0.75; and (5) the predicted binding probability of the agent's molecule across the same diffusion samples exceeds 0.7. Gates 1, 2, and 3 enforce 2D novelty; gates 4 and 5 enforce 3D mimicry, ensuring the scaffold hopped molecule adheres to the same pharmacophore, and binding competence.

\noindent\textbf{Design rationale}
Every task is anchored to a concrete witness molecule that has already cleared the same evaluation pipeline the agent will be scored against, ensuring the existence of at least one solution for the task instance. The 2D-dissimilarity and 3D-interaction-similarity criteria rule evaluates for a true scaffold hopped molecule and that the submitted molecule matches the binding mode of the original reference molecule. High predicted binding probability of the submitted molecule is also enforced. We score interaction similarity against a consensus fingerprint over the 10-sample diffusion ensemble rather than against the single most confident pose to ensure robustness against variance in the Boltz2 poses. Requiring a tuple to appear in at least 60\% of samples discards interactions that are artifacts of any one diffusion draw and retains only those the model is confident enough to recover under structural perturbation.

\subsection{Lead Optimization}
\label{sec:task_type_4}

Each instance of this task type asks the agent to modify a known active molecule so that it improves a specified set of pharmacological optimization objectives, holds a separate set of properties within tolerance, and clears a fixed panel of hard drug-likeness constraints. As in the Scaffold Hopping task type, every task instance is constructed around an explicit witness molecule that has already been verified to satisfy every gate the agent will be scored against, guaranteeing the solvability of the task instance.

\noindent\textbf{Target and pocket selection}
Targets are drawn from a curated set of UniProt entries with high-resolution co-crystal coverage. For each target, a single representative PDB is chosen; its FASTA sequence and the residues within 6~\AA{} of the crystal ligand define a fixed pocket that is reused for all candidate co-folding within the target.

\noindent\textbf{Candidate pool collection}
For each target we aggregate experimentally reported actives from ChEMBL, BindingDB, and PubChem, deduplicated by canonical SMILES. The pool is then filtered through the same hard-constraint panel that will be applied to the agent's submission at evaluation time. The hard-constraint panel is found in Table \ref{tab:hard_constraints}.

\begin{table}[t]
\centering
\footnotesize
\caption{List of hard-constraints for molecules evaluated in Lead Optimization and Fragment Assembly tasks.}
\label{tab:hard_constraints}
\begin{tabular}{ll}
\toprule
\textbf{Category} & \textbf{Constraint} \\
\midrule
SMILES validity & Valid SMILES only \\
Molecular weight & MW $<$ 600~Da \\
Lipophilicity & $\log P \in [-1, 5]$ \\
Topological polar surface area & TPSA $<$ 140~\AA$^2$ \\
H-bond donors & $\leq 5$ \\
H-bond acceptors & $\leq 10$ \\
Rotatable bonds & $\leq 10$ \\
Formal charge & $\in [-2, +2]$ \\
Synthetic accessibility & SA score $<$ 4.5~\citep{ertl2009syntheticaccessibility} \\
Structural alerts & No PAINS~\citep{baell2016feeling} or Brenk/NIH alerts~\citep{brenk2008lessons,limban2018structuralalert} \\
\bottomrule
\end{tabular}
\end{table}

\noindent\textbf{ADMET annotation}
Every surviving molecule is scored on eight ADMET properties (hERG inhibition, blood-brain-barrier permeability, aqueous solubility, CYP3A4 inhibition, hepatocyte clearance, Caco-2 permeability, plasma protein binding, Ames mutagenicity) via the ADMET-AI ensemble \citep{admet_package} trained on the corresponding TDC datasets \citep{huang2021therapeutics}.

\begin{table}[t]
\centering
\footnotesize
\caption{List of ADMET properties used as optimization objectives and hold-constant constraints.}
\label{tab:admet_properties}
\begin{tabular}{l}
\toprule
\textbf{Property} \\
\midrule
hERG inhibition \\
Blood-brain barrier permeability \\
Aqueous solubility \\
CYP3A4 inhibition \\
Hepatocyte clearance \\
Caco-2 permeability \\
Plasma protein binding \\
Ames mutagenicity \\
\bottomrule
\end{tabular}
\end{table}

\noindent\textbf{Co-folding screen}
Each filtered candidate is co-folded against the canonical receptor with Boltz2 \citep{passaro2025boltz}, constrained to the fixed pocket via fixed pocket contacts with 10 diffusion samples. We retain the predicted binding probability and predicted $\log_{10}(\text{IC}_{50})$ across samples, and discard any molecule whose predicted binding probability falls below 0.75. The pool of molecules surviving every filter constitutes the per-target validated pool from which both reference and witness are drawn.

\noindent\textbf{Pair finding and task generation}
Within each target's validated pool, we compute pairwise Morgan fingerprint Tanimoto (radius~$=2$, 2048~bits) and consider only pairs with similarity $\geq 0.7$. 

For each filtered pair of molecules within the validated pool, we derive the optimization objectives and hold-constant set directly from the measured property differences between the two molecules. A property is eligible to be an optimization objective if the witness improves over the reference by at least a property-specific threshold. A property is eligible to be a hold-constant constraint only if the witness's value already lies within the property's tolerance band relative to the reference. 

Up to five objectives and four hold-constant properties are selected greedily. The direction of BBB optimization is inferred from the witness properties rather than imposed externally. Furthermore, for the task instances without binding affinity as optimization objectives or hold-constant constraints (only a handful in SMDD-Bench), Boltz2 calculated binding probability is enforced as a hard constraint for the agent's submitted molecule to prevent the agent from throwing away binding in the pursuit of optimizing ADMET properties.

\noindent\textbf{Agent inputs and outputs}
The agent is given: \texttt{complex.pdb} (the Boltz2-predicted reference complex, with the ligand on chain L), \texttt{reference.sdf} (the reference ligand conformer extracted from the complex), \texttt{protein.fasta} (the canonical sequence), and the pocket residue list, the target ligand identifier, the list of optimization objectives with directions and thresholds, the list of hold-constant properties with tolerance bands, the full hard-constraint panel, the BBB direction, and the baseline property values measured on the reference. The agent is required to emit \texttt{solution.smi} containing a single SMILES string. The witness molecule is never exposed to the agent.

\noindent\textbf{Evaluation}
A submission must clear five gates: (1) the SMILES is valid and sanitizable; (2) every hard constraint in the panel is satisfied, including a Tanimoto similarity to the reference $\geq 0.7$ that anchors the submission to the same lead series; (3) every hold-constant property lies within its tolerance band relative to the reference baseline; (4) every optimization objective improves over the reference by at least its required threshold; and (5) Boltz2 co-folding of the agent's molecule under the same protocol used during generation produces a predicted binding probability above 0.7 together with the affinity checks implied by the binding-affinity objective or hold-constant constraint. ADMET properties are recomputed from the agent's SMILES with the same ADMET-AI ensemble used in generation.

\noindent\textbf{Design rationale}
Deriving objectives and hold-constant sets from the actual reference--witness property delta, rather than imposing a fixed objective list per target, produces tasks that provably solvable. The high-Tanimoto pairing constraint enforces that the task captures local lead-series chemistry rather than de novo design, which is the regime in which the multi-objective trade-offs are most relevant. Because the entire hard-constraint panel and the ADMET model are applied identically at generation and at evaluation, the agent is scored under the exact filter that defined task feasibility, and any submission that clears the gates is at least as good as a known real molecule on every property the task names.

We investigate the indicators of the difficulty of the Lead Optimization tasks in Section \ref{sec:lead-opt-difficulty}.

\subsection{Fragment Assembly}
\label{sec:task_type_5}

Each instance of this task type asks the agent to design a single drug-like molecule that incorporates one or two given 3D molecular fragments at their specified poses inside a protein binding pocket, covalently links them when more than one is provided, and binds the target well. As in the previous two task types, every task instance is constructed around an explicit witness molecule -- here, the known active from which the fragments were carved -- that has already been verified to satisfy every gate the agent will be scored against, guaranteeing the solvability of the task instance.

\noindent\textbf{Target and active selection}
Targets are sourced from UniProt entries that have at least two PDB entries containing distinct, non-buffer ligands of at least 10 heavy atoms, after filtering crystallization additives. For each (UniProt, PDB, ligand) triplet, the ligand SMILES is reconstructed via Chemical Component Dictionary templates and the binding pocket is defined as the protein residues with any heavy atom within 6~\AA{} of the ligand.

\noindent\textbf{Witness co-folding}
Each candidate active is co-folded against its source protein with Boltz2 \citep{passaro2025boltz} under pocket constraints derived from the crystal pose, with the affinity head enabled and 10 diffusion samples. Only chains that participate in pocket contacts are retained for the co-fold; co-ligands and unrelated chains are excluded. Actives are kept only if their predicted binding probability exceeds 0.75. The resulting predicted $\log_{10}(\text{IC}_{50})$ is recorded as the witness affinity baseline against which the agent's submission will be scored.

\noindent\textbf{Fragment generation}
Surviving actives are decomposed via BRICS \citep{brics} into chemically reasonable fragments at retrosynthetically meaningful bonds. Each fragment is retained only if it (i) contains at least 6 heavy atoms after dummy-atom removal, (ii) makes at least one PLIP \citep{salentin2015plip} protein-ligand interaction with the receptor when scored on the witness's docked pose, and (iii) lies within 4~\AA{} of the protein. The first criterion ensures the fragment is large enough to be a meaningful pharmacophore rather than a methyl-scale stub; the second and third together guarantee that the fragment is genuinely engaged with the pocket rather than solvent-exposed. 

\noindent\textbf{Pair sampling for two-fragment tasks}
When a task instance is to contain two fragments, the pair is drawn from a single source active and is required to have a closest-heavy-atom distance of at least 6~\AA, this prevents the fragments from being spatially close enough to admit a short or simple linker. 

\noindent\textbf{Agent inputs and outputs}
The agent is given: \texttt{pocket.pdb} (the witness's Boltz2-predicted protein structure with the witness ligand stripped), one \texttt{fragment\_i.sdf} per fragment (the 3D pose of each fragment as it sat in the witness's Boltz2 docked complex), the binding affinity baseline target for the submited molecule, and the full hard-constraint panel also used in the Lead Optimization task (see Table \ref{tab:hard_constraints}). The agent is required to emit \texttt{solution.smi} containing a single SMILES string. The witness is never exposed to the agent.

\noindent\textbf{Evaluation}
A submission must clear five gates: (1) the SMILES is valid; (2) every fragment SMARTS pattern is a substructure of the agent's molecule; (3) the fragments admit a disjoint substructure assignment (no shared atoms across fragments) and are pairwise connected by a covalent path in the agent's molecule; (4) the agent's molecule clears the same hard-constraint panel specified in Table \ref{tab:hard_constraints}; and (5) Boltz2 co-folding of the agent's molecule against the source receptor under the same pocket constraint and 10 diffusion samples produces (a) a predicted binding probability above 0.7, (b) a predicted $\log_{10}(\text{IC}_{50})$ no more than 0.1 log units worse than the witness baseline, and (c) a per-fragment 3D pose RMSD below 2.0~\AA. RMSD is computed by SMARTS-matching each fragment against the docked agent ligand, extracting the 3D coordinates of the matched atoms, aligning the docked complex to \texttt{pocket.pdb} by Kabsch superposition on the C$_\alpha$ atoms of the pocket residues, and selecting the best-aligned diffusion sample for the comparison.

\noindent\textbf{Design rationale}
The single-active witness construction guarantees that for every task there exists at least one molecule -- the original active itself -- that contains all fragments at exactly the required poses, links them through known chemistry, and already satisfies the binding and drug-likeness panels under the same Boltz2 protocol used at evaluation time. The 3D pose RMSD threshold enforces the docked geometry to recapitulate the fragment poses given as input, which forces the linker chemistry to be physically compatible with the relative orientation of the fragments rather than just being topologically valid or bind well.

\section{Minimalist Agent Setup}
\label{sec:min_agent_appendix}

To isolate model capability from harness engineering, all baseline runs use a deliberately minimalist ReAct loop. The agent sees only
the obfuscated task description and the tool schemas. All tools run in an isolate docker container with an isolated filesystem. The standard agent configuration can be found in Table \ref{tab:agent-config}.

\noindent\textbf{Tools.} The agent is provided four tools: (i)~\texttt{python(code)}: executes Python in the sandbox with RDKit, OpenBabel, PLIP, BioPython, NumPy, SciPy, and scikit-learn preinstalled; (ii)~\texttt{predict\_boltz(smiles, protein\_sequences, \ldots)}: runs Boltz-2 co-folding with affinity prediction; (iii)~\texttt{predict\_admet(smiles)}: runs ADMET-AI property predictors and returns all the values of all properties of the molecule in a single call; The Boltz/ADMET oracles live in an isolated venv that the agent's \texttt{python} tool cannot import, forcing all oracle calls through the budgeted interface.

\noindent\textbf{Sandbox.} Each task runs in a fresh \texttt{smdd-agent} container on an internal Docker network with no external routing; the only reachable endpoint is \texttt{api.colabfold.com:443} via an nginx TCP-passthrough sidecar (required by Boltz's MSA server). Task files (e.g.\ \texttt{complex.pdb}, \texttt{reference.sdf}) are obfuscated to remove target names, PDB codes, and ligand three letter codes to suppress data-leakage shortcuts via memorized identifiers.

\begin{table}[h]
\centering
\small
\caption{Baseline agent harness configuration. All harness values are constants across tasks and models. Boltz diffusion samples was set to 10 to mitigate stochasticity in Boltz outputs. We find empirically find settings above 3 diffusion samples to sufficiently reduce stochasticity across multiple runs.}
\label{tab:agent-config}
\begin{tabular}{lll}
\toprule
\textbf{Parameter} & \textbf{Value} & \textbf{Notes} \\
\midrule
\multicolumn{3}{l}{\emph{Sampling}} \\
Temperature                & $1.0$              & --- \\
Sampling seed              & $42$               & ---  \\
Max ReAct turns            & $100$              & one LLM call per turn \\
Max cumulative tokens      & $10{,}000{,}000$   & input $+$ output, summed across turns \\
Max consecutive refusals   & $15$               & terminates run if exceeded \\
\midrule
\multicolumn{3}{l}{\emph{Oracle budgets (per task)}} \\
Boltz-2 calls              & $8$                & shared across rollout \\
ADMET-AI calls             & $15$               & shared across rollout \\
Boltz diffusion samples    & $10$               & matches evaluator setting \\
\bottomrule
\end{tabular}
\end{table}

\subsection{Selecting Oracle Calls}
\label{sec:choosing_oracle_calls}

By default, across all evaluations, we set the total number of available Boltz2 calls to 8 and the number of ADMET-AI calls to 15. Note that a single ADMET-AI call returns all 8 ADMET properties for the inputted SMILES string. The complete reasoning behind only providing the LLM with a limited number of oracle calls is outlined in Section \ref{sec:discussion}. In short, our goal is to treat the oracle calls as though they were ground truth, real-world wet lab experiments, hence the limited number of oracle calls. Ultimately, SMDD-Bench can be considered a testbed for developing the right training strategies and agent harnesses for SMDD agents -- even though agents trained on SMDD-Bench adhere to the Boltz2/ADMET-AI distribution of binding affinities, structures, and properties. Afterwards, the agents can be trained using those same strategies with a wet-lab-in-the-loop -- where the oracle calls are directly made to an automated wet lab, performing the synthesis and then the pertinent assays completely autonomously. 

The amount of Boltz2 and ADMET-AI calls must be tight enough to reward oracle-efficient reasoning instead of brute-force enumeration but generous enough such that a competent agent is not bottlenecked by oracle access.

In order to determine a standard  number of oracle calls, we performed a sweep across oracle budget choices using a small calibration subset. We constructed the calibration split of $25$ tasks drawn proportionally from each task type, with a minimum of one task per type (each task yielding $\max(1, \lfloor 25 \cdot n_t / N \rfloor)$, where $n_t$ is the number of tasks of type $t$ and $N$ the total benchmark size). The Lead Optimization task instances were sampled stratified by problem complexity, defined as the sum of the number of optimization objectives and the number of hold-constant constraint, such that the calibration set spans the full difficulty range of the task type.

We swept each oracle independently while pinning the other to a generous ceiling:
\begin{itemize}
  \item \textbf{Boltz sweep:} ADMET budget fixed at $35$; Boltz budget
        varied over $\{5, 8, 12\}$.
  \item \textbf{ADMET sweep:} Boltz budget fixed at $12$; ADMET budget
        varied over $\{15, 25, 35\}$.
\end{itemize}
The shared end-point $(\text{boltz}=12, \text{admet}=35)$ was reused, giving five distinct $(B, A)$ configurations in total. All other agent-harness settings were held at their defaults; the agent model used for calibration was \texttt{GPT5.4}. 

For each sweep, we computed the per-type pass rate at every budget level and selected the "knee" of the plot, where the increase in performance began to saturate as the number of oracle calls increased.

\section{Additional Agent Results}
\label{sec:additional_agent_results}
\begin{figure}
    \centering
    \includegraphics[width=0.4\linewidth, trim=28px 10px 28px 22px, clip]{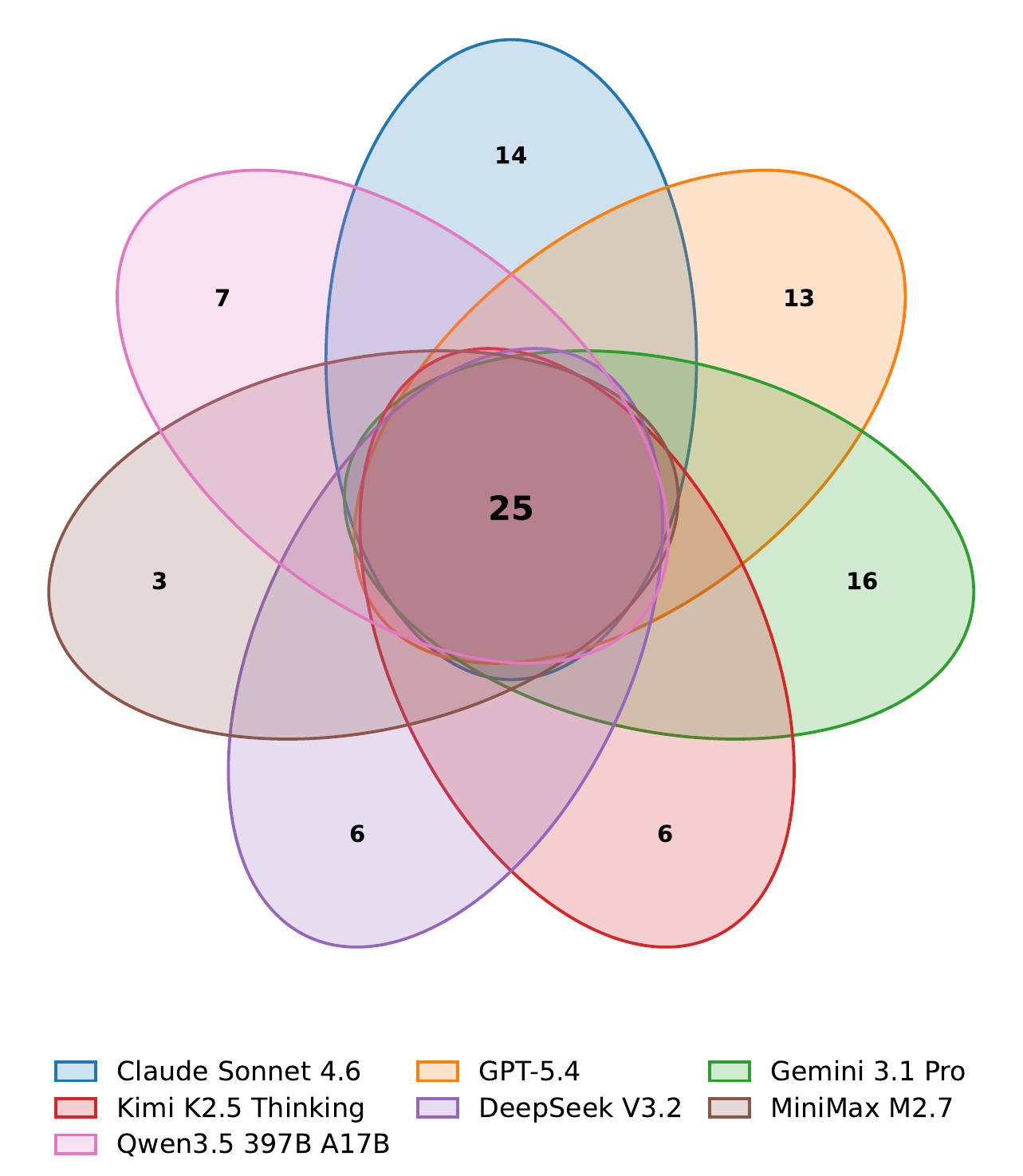}
    \caption{The number of task instances that each LLM agent gets correct as well as the number of task instances that each LLM \textbf{uniquely} gets correct on the full SMDD-Bench dataset.}
    \label{fig:venn_diagram_full}
\end{figure}

\begin{figure}
  \centering
  \includegraphics[width=0.4\textwidth, trim=28px 10px 28px 22px, clip]{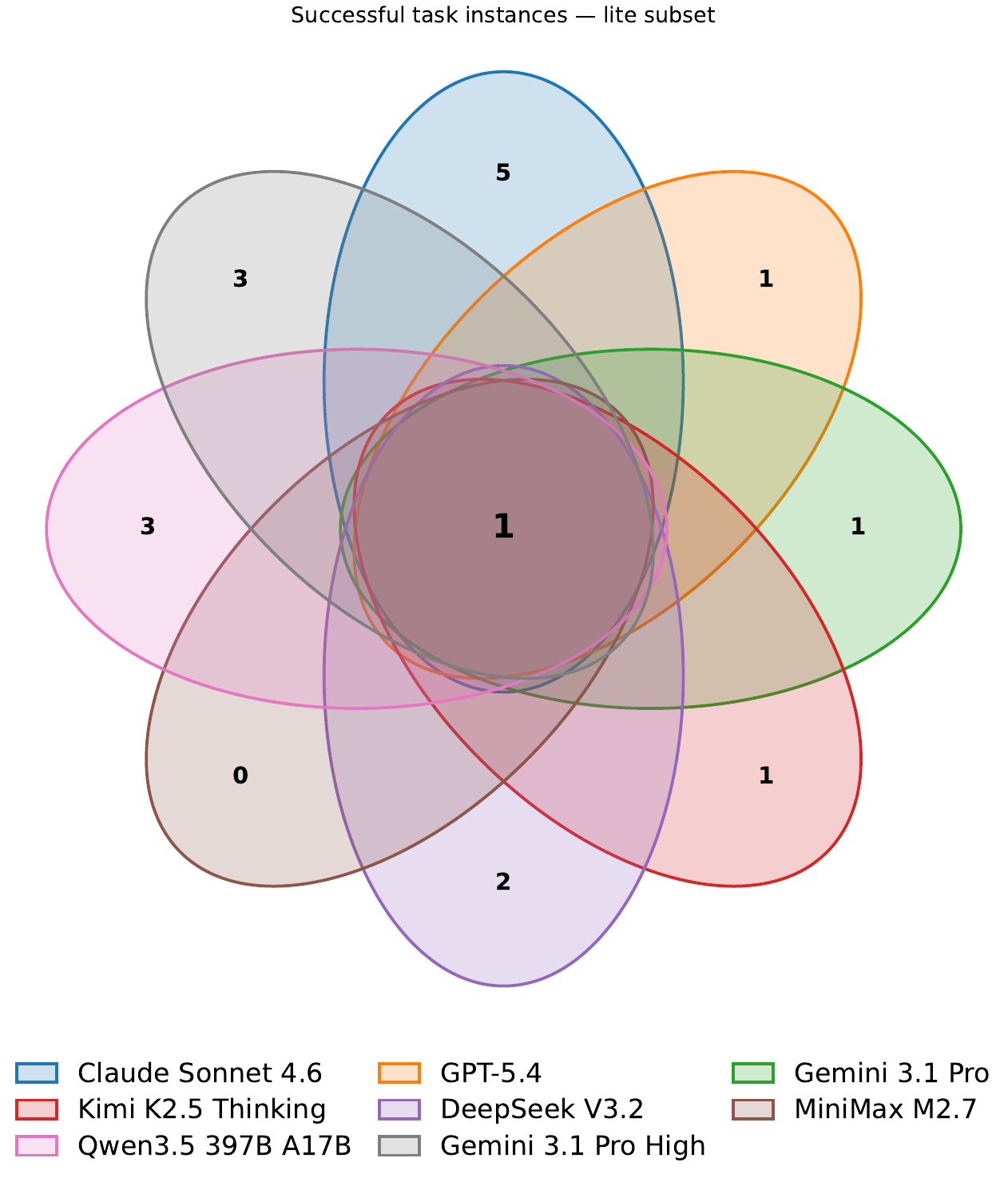}
  \caption{The number of task instances that each LLM agent gets correct as well as the number of task instances that each LLM \textbf{uniquely} gets correct on the full SMDD-Bench Lite dataset.}
  \label{fig:venn_diagram_lite}
\end{figure}

Table \ref{tab:cost_breakdown} shows the cost breakdown of each of the LLMs evaluated on the full SMDD-Bench benchmark while Table \ref{tab:wallclock_time} shows the average amount of wall-clock time required for task and model. Each task instance was run on either an A800 or A6000 GPU. Table \ref{tab:witness_match} investigates the rate at which each LLM agent was able to recover exactly the witness that was found during task generatio out of all successful instances. Figure \ref{fig:venn_diagram_full} and Figure \ref{fig:venn_diagram_lite} depict the number of task instances each LLM can uniquely get correct as well as the number of task instances all LLMs get correct for both SMDD-Bench and SMDD-Bench Lite respectively. Table \ref{tab:type2_breakdown} shows the number of Interaction Point Discovery task instances correctly discovering 0, 1, 2, or 3 of the most conserved interaction points. Table \ref{tab:turn_and_tool_counts} depicts

\begin{table}[ht]
\centering
\footnotesize 
\setlength{\tabcolsep}{3.5pt} 
\caption{Cost breakdown per instance for Task Types 1 through 5 and Overall averages across 7 frontier LLMs.}
\label{tab:cost_breakdown}
\begin{tabular}{l
                S[table-format=1.2]
                S[table-format=1.2]
                S[table-format=1.2]
                S[table-format=1.2]
                S[table-format=1.2]
                S[table-format=1.2]}
\toprule
& \multicolumn{6}{c}{Average Cost (\$/task instance)}  \\
\cmidrule(lr){2-6}
Model
& \multicolumn{1}{c}{\makecell{2D Pharma-\\cophore ID}}
& \multicolumn{1}{c}{\makecell{Interaction Point \\ Discovery}}
& \multicolumn{1}{c}{\makecell{Scaffold \\ Hopping}}
& \multicolumn{1}{c}{\makecell{Lead \\ Optimization}}
& \multicolumn{1}{c}{\makecell{Fragment \\ Assembly}}
& \multicolumn{1}{c}{\makecell{Overall}} \\
\midrule
Claude Sonnet 4.6     & 1.09          & 0.30          & 0.92          & 1.01          & 3.85          & 1.31          \\
GPT-5.4               & 0.61          & 0.72          & 0.62          & 0.72          & 1.31          & 0.78          \\
Gemini 3.1 Pro        & 0.39          & 0.16          & 0.31          & 0.62          & 1.24          & 0.63          \\
Kimi K2.5 Thinking    & 0.41          & 0.09          & {\textbf{0.18}} & 0.43          & 0.53          & 0.40          \\
DeepSeek V3.2         & {\textbf{0.21}} & 0.09          & 0.29          & 0.47          & 0.57          & 0.43          \\
MiniMax M2.7          & 0.41          & {\textbf{0.07}} & 0.31          & {\textbf{0.36}} & {\textbf{0.46}} & {\textbf{0.36}} \\
Qwen3.5 397B A17B     & 1.09          & 0.15          & 0.63          & 0.78          & 0.81          & 0.75          \\
\bottomrule
\end{tabular}
\end{table}

\begin{table}[ht]
\centering
\footnotesize
\setlength{\tabcolsep}{3.5pt}
\caption{Average wall-clock time (minutes) per task, split by success status.}
\label{tab:wallclock_time}
\begin{tabular}{l l c c c c c c}
\toprule
& & \multicolumn{6}{c}{Average Time (minutes)} \\
\cmidrule(lr){3-8}
Model & Status
& \makecell{2D Pharma-\\cophore ID}
& \makecell{Interaction Point \\ Discovery}
& \makecell{Scaffold \\ Hopping}
& \makecell{Lead \\ Optimization}
& \makecell{Fragment \\ Assembly}
& \makecell{Overall} \\
\midrule

Claude Sonnet 4.6 
& Passed & \num{8.1} & -- & \num{9.3} & \num{14.9} & -- & \num{14.6} \\
& Failed & \num{12.1} & \num{2.9} & \num{18.9} & \num{27.9} & \num{58.7} & \num{29.5} \\
\midrule

GPT-5.4 
& Passed & \num{2.6} & -- & \num{14.4} & \num{18.4} & \num{41.3} & \num{18.2} \\
& Failed & \num{5.2} & \num{5.7} & \num{27.4} & \num{28.4} & \num{39.9} & \num{26.9} \\
\midrule

Gemini 3.1 Pro 
& Passed & \num{3.1} & \num{2.0} & -- & \num{12.4} & \num{77.4} & \num{12.4} \\
& Failed & \num{3.7} & \num{1.9} & \num{13.5} & \num{25.9} & \num{41.0} & \num{23.4} \\
\midrule

Kimi K2.5 Thinking 
& Passed & \num{8.0} & -- & \num{4.4} & \num{21.9} & -- & \num{21.5} \\
& Failed & \num{6.1} & \num{2.2} & \num{28.7} & \num{39.5} & \num{44.2} & \num{34.0} \\
\midrule

DeepSeek V3.2 
& Passed & \num{10.0} & -- & \num{25.9} & \num{39.0} & -- & \num{38.3} \\
& Failed & \num{14.0} & \num{7.3} & \num{39.4} & \num{51.2} & \num{50.8} & \num{44.4} \\
\midrule

MiniMax M2.7 
& Passed & \num{20.7} & -- & \num{13.2} & \num{36.9} & -- & \num{36.0} \\
& Failed & \num{27.6} & \num{8.0} & \num{54.2} & \num{54.8} & \num{49.2} & \num{49.6} \\
\midrule

Qwen3.5 397B A17B 
& Passed & \num{9.0} & -- & \num{4.1} & \num{16.4} & -- & \num{16.2} \\
& Failed & \num{9.8} & \num{2.0} & \num{15.2} & \num{22.3} & \num{24.9} & \num{19.5} \\
\bottomrule
\end{tabular}
\end{table}

\begin{table}[ht]
\centering
\footnotesize
\setlength{\tabcolsep}{3.5pt}
\caption{Rate (\%) at which the LLM agents were able to recover the witness molecule exactly. The denominator is the number of \textbf{successful} instances for the LLM on the specific task type. }
\label{tab:witness_match}
\newcommand{\rate}[1]{\num{#1}\%}
\begin{tabular}{l
                c
                c
                c}
\toprule
Model
& \makecell{Scaffold \\ Hopping}
& \makecell{Lead \\ Optimization}
& \makecell{Fragment \\ Assembly} \\
\midrule

Claude Sonnet 4.6 & \rate{0.0} & \rate{16.5} & -- \\
GPT-5.4           & \rate{0.0} & \rate{21.9} & \rate{0.0} \\
Gemini 3.1 Pro    & --         & \rate{21.7} & \rate{0.0} \\
Kimi K2.5 Thinking& \rate{0.0} & \rate{16.9} & -- \\
DeepSeek V3.2     & \rate{0.0} & \rate{16.1} & -- \\
MiniMax M2.7      & \rate{0.0} & \rate{14.1} & -- \\
Qwen3.5 397B A17B & \rate{0.0} & \rate{19.1} & -- \\

\bottomrule
\end{tabular}
\end{table}

\begin{table}[t]
    \centering
    \caption{The number of task instances in the Interaction Point Discovery task type that each LLM was able to successfully pinpoint 0, 1, 2, or 3 interaction points for. Only instances where the LLM was able to find 3 points were considered a success. Note that there may be more than 3 ground truth points the LLM-discovered points are able to match to.}
    \label{tab:type2_breakdown}
    \footnotesize
    \renewcommand{\arraystretch}{1.15}
    \begin{tabular}{lcccc}
    \toprule
    Model & \textbf{0/3} & \textbf{1/3} & \textbf{2/3} & \textbf{3/3} \\
    \midrule
    Claude Sonnet 4.6      & 21 & 4  & 0 & 0 \\
    GPT-5.4                & 11 & 11 & 3 & 0 \\
    Gemini 3.1 Pro         & 14 & 9  & 1 & 1 \\
    Kimi K2.5 Thinking     & 23 & 2  & 0 & 0 \\
    DeepSeek V3.2          & 25 & 0  & 0 & 0 \\
    MiniMax M2.7           & 24 & 1  & 0 & 0 \\
    Qwen3.5 397B A17B      & 20 & 5  & 0 & 0 \\
    \bottomrule
    \end{tabular}
\end{table}

\begin{table}[ht]
\centering
\footnotesize
\setlength{\tabcolsep}{3.5pt}
\caption{The total number of turns and tool calls, broken down by task type for the SMDD-Bench full runs. The Boltz2 oracle budget was 8 calls while the ADMET-AI oracle budget was 15 calls. Most of the turns were spent on visualization, analysis, and simulation using the python tool.}
\label{tab:turn_and_tool_counts}
\begin{tabular}{ll
                S[table-format=2.1]
                S[table-format=2.1]
                S[table-format=2.1]
                S[table-format=2.1]
                S[table-format=2.1]
                S[table-format=2.1]}
\toprule
Model
& Metric
& \multicolumn{1}{c}{\makecell{2D Pharma-\\cophore ID}}
& \multicolumn{1}{c}{\makecell{Interaction Point \\ Discovery}}
& \multicolumn{1}{c}{\makecell{Scaffold \\ Hopping}}
& \multicolumn{1}{c}{\makecell{Lead \\ Optimization}}
& \multicolumn{1}{c}{\makecell{Fragment \\ Assembly}}
& \multicolumn{1}{c}{\makecell{Overall}} \\
\midrule

Claude Sonnet 4.6
& Total Turns           & 29.4 & 11.6 & 27.1 & 28.0 & 57.6 & 30.7 \\
& python          & 28.4 & 10.6 & 23.4 & 20.0 & 50.6 & 23.9 \\
& predict\_boltz  & 0.0  & 0.0  & 3.8  & 3.0  & 7.8  & 3.3 \\
& predict\_admet  & 0.0  & 0.0  & 0.8  & 10.8 & 0.8  & 7.5 \\
& other           & 0.0  & 0.0  & 0.0  & 0.0  & 0.0  & 0.0 \\
\midrule

GPT-5.4
& Total Turns           & 13.8 & 13.8 & 15.0 & 14.6 & 21.4 & 15.4 \\
& python          & 12.6 & 12.8 & 12.2 & 9.7  & 18.3 & 11.3 \\
& predict\_boltz  & 0.0  & 0.0  & 5.5  & 4.3  & 7.9  & 4.4 \\
& predict\_admet  & 0.0  & 0.0  & 0.8  & 12.7 & 0.5  & 8.7 \\
& other           & 0.0  & 0.0  & 0.0  & 0.0  & 0.0  & 0.0 \\
\midrule

Gemini 3.1 Pro
& Total Turns           & 14.2 & 7.4  & 11.1 & 19.3 & 31.4 & 19.1 \\
& python          & 13.2 & 6.4  & 7.8  & 9.3  & 23.5 & 10.9 \\
& predict\_boltz  & 0.0  & 0.0  & 2.7  & 2.9  & 7.3  & 3.1 \\
& predict\_admet  & 0.0  & 0.0  & 0.1  & 10.7 & 0.8  & 7.4 \\
& other           & 0.0  & 0.0  & 0.0  & 0.0  & 0.0  & 0.0 \\
\midrule

Kimi K2.5 Thinking
& Total Turns           & 29.2 & 11.6 & 20.8 & 35.8 & 35.9 & 32.7 \\
& python          & 28.2 & 10.6 & 13.9 & 17.8 & 24.8 & 18.4 \\
& predict\_boltz  & 0.0  & 0.0  & 5.2  & 5.2  & 7.7  & 5.0 \\
& predict\_admet  & 0.0  & 0.0  & 0.8  & 12.0 & 2.6  & 8.5 \\
& other           & 0.0  & 0.0  & 0.0  & 0.0  & 0.0  & 0.0 \\
\midrule

DeepSeek V3.2
& Total Turns           & 18.0 & 10.8 & 24.0 & 36.0 & 31.7 & 32.1 \\
& python          & 17.0 & 9.8  & 13.6 & 15.7 & 19.5 & 15.7 \\
& predict\_boltz  & 0.0  & 0.0  & 6.8  & 5.3  & 7.3  & 5.2 \\
& predict\_admet  & 0.0  & 0.0  & 2.3  & 12.6 & 1.9  & 9.0 \\
& other           & 0.0  & 0.0  & 0.0  & 0.0  & 0.0  & 0.0 \\
\midrule

MiniMax M2.7
& Total Turns           & 40.0 & 14.0 & 39.2 & 43.4 & 48.8 & 42.0 \\
& python          & 38.9 & 12.9 & 30.9 & 27.4 & 39.1 & 29.0 \\
& predict\_boltz  & 0.0  & 0.0  & 6.7  & 5.9  & 7.7  & 5.6 \\
& predict\_admet  & 0.0  & 0.0  & 0.3  & 12.8 & 0.5  & 8.8 \\
& other           & 0.0  & 0.0  & 0.0  & 0.0  & 0.0  & 0.0 \\
\midrule

Qwen3.5 397B A17B
& Total Turns           & 42.0 & 13.4 & 40.3 & 48.5 & 46.2 & 45.3 \\
& python          & 37.2 & 11.0 & 31.7 & 26.8 & 36.0 & 28.2 \\
& predict\_boltz  & 0.0  & 0.0  & 6.8  & 5.7  & 7.7  & 5.5 \\
& predict\_admet  & 0.0  & 0.0  & 0.4  & 11.7 & 1.0  & 8.0 \\
& other           & 0.0  & 0.0  & 0.0  & 0.0  & 0.0  & 0.0 \\

\bottomrule
\end{tabular}
\end{table}

\subsection{Indicators of the Difficulty for Lead Optimization Tasks}
\label{sec:lead-opt-difficulty}

\begin{figure}
  \centering
  \includegraphics[width=\textwidth, trim=0px 0px 0px 0px, clip]{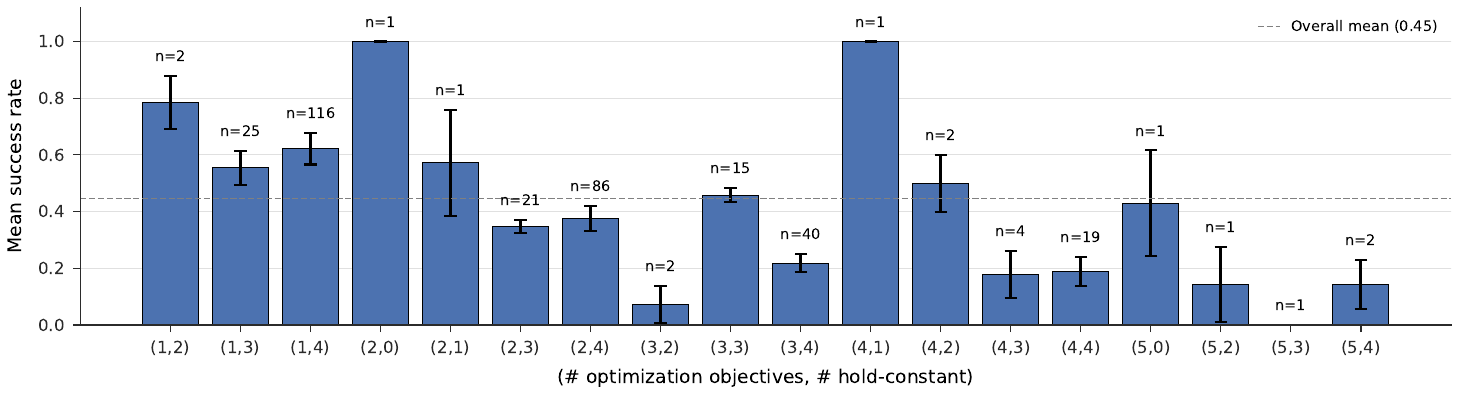}
  \caption{The average success rate across all 7 evaluated frontier LLM agents with respect to the number of optimization objectives and hold-constant constraints. There is no noticeable trend indicating the difficulty of the task instance increases as the number of objectives or constraints increases.}
  \label{fig:task4_success_by_tuple}
\end{figure}

We find that the number of optimization objectives and hold constant constraints are not indicative of the difficulty of a Lead Optimization task instance, as seen in Figure \ref{fig:task4_success_by_tuple}, where the difficulty of a task instance is defined by the average success rate of the 7 evaluated frontier LLM agents. This may be a byproduct of the greedy task generation algorithm used to generate Lead Optimization task instances -- for every pair of molecules, all conserved and optimized properties of a molecule within the pair are used to create the task instance. Therefore, properties with synergistic effects can be simultaneously co-optimized via the same local molecular edit. 

Rather than the number of optimization objectives and hold-constant constraints, we find that tasks with a large number of properties with clashing effects to be the greatest indicator for task difficulty. The incompatible, clashing property pairs can be found in Table \ref{tab:incompatible_pairs} \citep{waring2010lipophilicity, gleeson2008generation, leeson2007influence, varma2009physicochemical}. These properties are chemically anti-correlated -- standard local modifications intended to optimize for one property result in the degradation of the other. As a result, molecular modifications are typically multi-step and more complex in nature compared to task instances with fewer incompatible property pairs. 

\begin{table}[h]
  \footnotesize
  \centering
  \caption{The incompatible property pairs defined in SMDD-Bench. These are pairs of properties which are not sampled for co-optimization targets but can be found in hold-constant constraints or have one property as an optimization objective and the other as a hold-constant constraint.}
  \label{tab:incompatible_pairs}
  \begin{tabular}{llp{0.45\linewidth}}
  \toprule
  Property A (direction) & Property B (direction) & Rationale \\
  \midrule
  Aqueous solubility ($\uparrow$)        & BBB penetration ($\uparrow$)        & Solubility favors polar/hydrophilic compounds; BBB
  penetration favors lipophilic compounds. \\
  Plasma protein binding ($\downarrow$)  & BBB penetration ($\uparrow$)        & Low PPB requires reduced lipophilicity; BBB penetration
  is driven by lipophilicity. \\
  Caco-2 permeability ($\uparrow$)       & Plasma protein binding ($\downarrow$) & Caco-2 permeability tracks lipophilicity, which also
  drives plasma protein binding. \\
  Aqueous solubility ($\uparrow$)        & Hepatic clearance ($\downarrow$)    & Increasing polarity to improve solubility tends to expose
   substrates to renal/metabolic clearance. \\
  Aqueous solubility ($\uparrow$)        & Caco-2 permeability ($\uparrow$)    & Solubility favors polarity; Caco-2 permeability favors
  lipophilicity. \\
  Caco-2 permeability ($\uparrow$)       & BBB penetration ($\downarrow$)      & Both are passive permeability measures; maximizing one
  while minimizing the other is contradictory. \\
  \bottomrule
  \end{tabular}
\end{table}

\begin{figure}
  \centering
  \includegraphics[width=\textwidth, trim=85px 220px 100px 140px, clip]{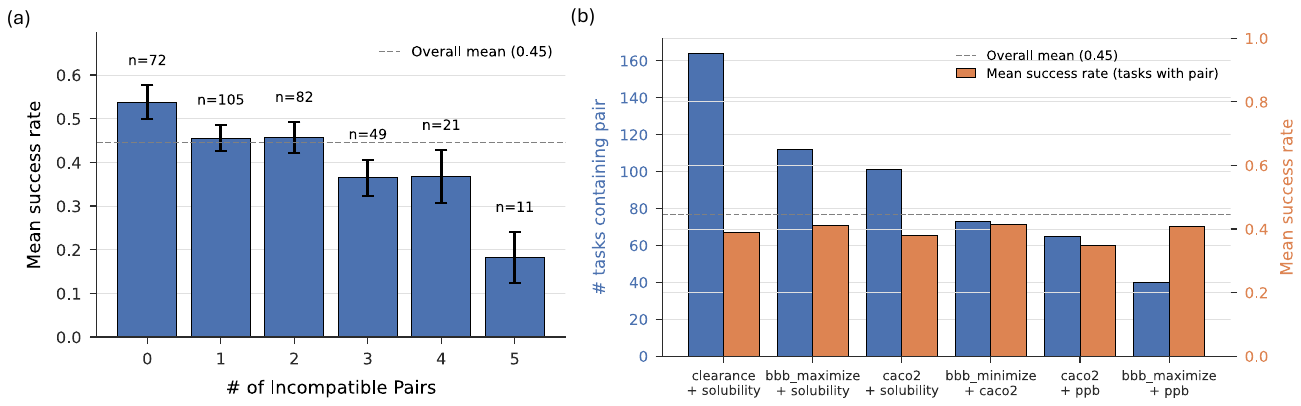}
  \caption{(a) The mean success rate across all 7 agents evalauted on SMDD-Bench plotted against the number of incompatible pairs in the optimization objectives or hold constant constraints. Pairs are not mutually exclusive, meaning one property may be incompatible with multiple other properties. (b) The mean success rate across all agents plotted against the specific incompatible pair. There is no relationship between the difficulty of the task and the specific pairing.}
  \label{fig:incompatible_pairs}
\end{figure}

\subsection{Analysis on inter-agent submission diversity}
\label{sec:turn_count_trends}
\begin{figure}
  \centering
  \includegraphics[width=\textwidth, trim=80px 220px 90px 210px, clip]{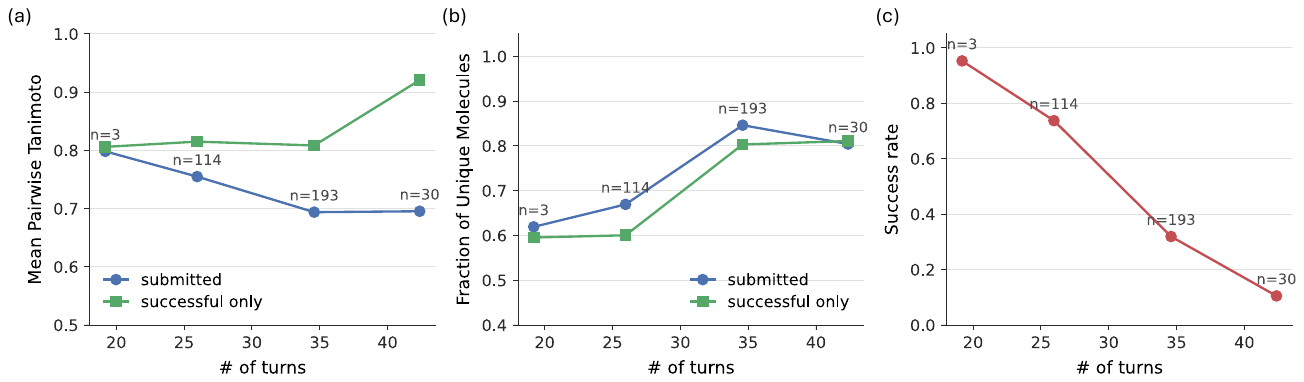}
  \caption{We investigate the relationship between the submissions of multiple different LLMs on the same task instance for the Lead Optimization task type. (a) The average pairwise Tanimoto similarity between the submissions of the 7 LLMs evaluated on the Lead Optimization task in SMDD-Bench plotting against the number of turns to solve the task instance. (b) The fraction of unique molecules submitted by the 7 LLMs with respect to number of turns. (c) The success rate of the 7 LLMs with respect to the number of turns. }
  \label{fig:turn_count_trends}
\end{figure}

As seen in Figure \ref{fig:turn_count_trends}, we find, perhaps intuitively, that an increase in the number of turns an LLM takes to solve a task is correlated with the increasing difficulty of the task (as indicated by success rate). We also observe that, as the LLM agents require more turns for more challenging tasks, the pairwise Tanimoto similarity between successful submissions of different LLMs for the same task increases while the fraction of unique molecules also increases. This is a surprising result, indicating that, for challenging tasks, the successful agents are converging onto the same narrow region in chemical space but via different approaches -- leading to successful submissions with high Tanimoto similarity but also high unique fraction.

Conversely, simpler tasks lead to a low fraction of unique molecules; most likely resulting from the agents identifying and attempting the same set of straightforward strategies which passed the oracle calls on an early turn count. 

\section{Selecting SMDD-Bench Lite}
\label{sec:smdd-bench-lite}

Running the full $502$-task benchmark is expensive in wall clock time and GPU compute requirement (see Table \ref{tab:wallclock_time}). As a result, we introduce the SMDD-Bench Lite subset, a 100 task instance representative subset of SMDD-Bench, making faster model and harness iteration faster and more unified. 

Lite preserves the per-type mix of the full benchmark as seen in Table \ref{tab:lite_breakdown}.

\begin{table}[h]
    \centering
    \caption{The task breakdown for SMDD-Bench Lite compared to SMDD-Bench.}
    \label{tab:lite_breakdown}
    \footnotesize
    \begin{tabular}{lrr}
    \toprule
    Task type & Full & Lite \\
    \midrule
    2D Pharmacophore Identification          &  $25$ &  $5$ \\
    Interaction Point Discovery              &  $25$ &  $5$ \\
    Scaffold Hopping                         &  $52$ & $10$ \\
    Lead Optimisation                        & $340$ & $68$ \\
    Fragment Assembly                         &  $60$ & $12$ \\
    \midrule
    Total                                           & $502$ & $100$ \\
    \bottomrule
    \end{tabular}
\end{table}
    
Selection proceeds independently within each task type, and combines stratified sampling with a greedy representativeness/diversity objective. For each task we define a stratum used as a difficulty/discrimination
proxy:
\begin{itemize}
  \item \textbf{Lead optimization}: the tuple $(n_{\text{opt}}, n_{\text{hold}})$ of the
        number of optimisation objectives and hold-constant properties.
  \item \textbf{2D Pharmacophore ID, Interaction Point Discovery, Scaffold Hopping, Fragment Assembly}: the integer success rate of the task across the seven evaluated models,
\end{itemize}

\noindent\textbf{Stratum allocation.}
Given a per-type budget $k$ and stratum sizes $\{n_s\}$, we allocate $k$ slots across strata in proportion to $n_s$ using the largest-remainder (Hamilton) rule: each stratum gets $\lfloor k \cdot n_s / N \rfloor$ slots, and the remaining slots are awarded to the strata with the largest fractional parts, breaking ties by stratum key.

\noindent\textbf{Within-stratum greedy pick.}
Within each stratum we pick tasks one at a time. Let $\mathbf{r}_{\text{full}} \in [0,1]^{6}$ be the per-model pass-rate vector of the full type, and let $\mathbf{r}_S$ be the mean pass-rate vector of the currently-selected Lite tasks. At each step we pick the candidate $t$ minimizing
\[
\mathcal{L}(t) \;=\; \bigl\| \mathbf{r}_{S \cup \{t\}} - \mathbf{r}_{\text{full}} \bigr\|_1
\;+\; \lambda \cdot c(\text{target}(t)),
\]
where $c(\cdot)$ counts how many already-selected Lite tasks share the same protein/target as $t$, and $\lambda = 0.15$. The target field is parsed from the task id. The protein-reuse term acts as a tie-breaker between near-equivalently representative candidates rather than overriding the representativeness objective; ties are broken deterministically by sorted \texttt{task\_id}. Strata are processed in order of decreasing size so that the largest contributors to $\mathbf{r}_{\text{full}}$ shape the selection first.

\section{Selecting SMDD-Bench Diversity}
\label{sec:selecting-smdd-bench-diversity}

During a lead optimization campaign, one potential use case involves running many SMDD agents on the same task in parallel in order to achieve a diverse spread of successful molecules. In order to test this, we develop the SMDD-Bench Diversity subset, a standardized 20 instance subset of SMDD-Bench meant for each task instance to be run multiple times. 

In order to select the SMDD-Bench Diversity subset, we seek to find the most challenging consistently solvable instances. Task instances that are too simple are trivially solvable with minimal molecular edits. Task instances which are too challenging may be out of reach for the current frontier LLMs. 

We first rank all Lead Optimization task instances solvable by 5/7 LLMs and rank them by turn count. We use turn count as a proxy for ranking the difficulty and chemical solution space of a task instance, following insight from Appendix \ref{sec:turn_count_trends}. We then hand-select the 20 instances with the highest average turn counts across the 7 LLMs while ensuring target diversity (no single target dominates the subset). This selection strategy ensures the tasks are at the edge of the realm of solvable tasks for current frontier LLM agents. 

Note that, across the SMDD-Bench Diversity runs outlined in Table \ref{tab:smdd-bench-diversity}, none of the LLMs failed a task instance all 10 times the task instance was run.

\section{Enumeration vs. Selection Details}
\label{sec:what_if_details}

A failure on SMDD-Bench can come from two qualitatively different shortcomings of the agent: (i) the agent never enumerated a passing molecule at all, or (ii) the agent enumerated a passing molecule somewhere in its trajectory but never selected it to be tested by an oracle call. The enumeration vs. selection experiments are designed to disentangle these shortcomings: for every failed task, we extract and evaluate every SMILES string enumerated by the agent but not selected for an oracle call. We define a ``recovered'' task to be one whose original submission fails evaluation but where at least one trajectory-resident SMILES would have passed the full evaluator. 

We run the experiment for all seven evaluated models and the three generation-style task types: Scaffold Hopping, Lead Optimization, and Fragment Assembly. 2D Pharmacophore Identification and Interaction Point Discovery are excluded because their outputs are not molecules. In order to extract the SMILES string, we compact each trajectory to only the reasoning trace, LLM output, and python interpreter input before feeding the compacted trajectory to GPT5.4-mini-high. 

\noindent\textbf{Extraction Prompt}
The system prompt sent to GPT-5.4-mini is reproduced verbatim below:

\begin{quote}
{\small
\begin{verbatim}
You are a careful chemical-structure extractor.

You will receive a JSON list of trajectory events from an LLM agent that
was working on a small-molecule drug-discovery task. Each event has a
`turn` number, a `source` in {assistant_text, reasoning_content,
python_code}, and the raw text for that event.

Your job: extract EVERY SMILES string that appears in those events. For
each occurrence, return:

- "turn":   the turn number (integer)
- "source": one of "assistant_text", "reasoning_content", "python_code"
- "smiles": the raw SMILES string as it appears in the text
- "note":   a short (<=120 char) free-text note describing the context
            (e.g. "proposed as analog A", "printed in a Python list of
            candidates", "mentioned while analyzing reference ligand").

Rules:
- Do NOT invent or canonicalize SMILES. Copy them verbatim from the text.
- Report duplicates across turns / sources as separate occurrences.
- Include SMILES that appear inside Python source code as string literals.
- If a code snippet builds a SMILES via string concatenation and you can
  statically determine the final value, include it with a note that says
  so. Otherwise, skip it.
- Skip protein sequences, fragment placeholders like `[*]`, and reaction
  SMARTS.
- If you are unsure whether a token is a real SMILES, include it ---
  downstream RDKit validation will filter invalid ones.

Return a JSON object of the form:

{"occurrences": [{"turn": int, "source": str, "smiles": str, "note":
str}, ...]}

Return an empty list if no SMILES are present. Do not include any other
keys.
\end{verbatim}
}
\end{quote}

\section{Example Task Instances, Trajectories, and Analysis}
In this section, we show one successful LLM agent trajectory for each task type. Due to space considerations, each trajectory is limited to only the first 5 and last 5 turns. For all trajectories across all LLM runs, please refer to the official GitHub repository. 

\subsection{2D Pharmacophore Identification}
\subsubsection{Passing run: GPT-5.4 (medium reasoning)}
\label{traj:tt1:smdd-001-Q99873-0:passing}
\noindent\textbf{Task ID:} \texttt{smdd\_001\_Q99873\_0}\quad\textbf{Model:} GPT-5.4 (medium reasoning)\quad\textbf{Run:} \texttt{gpt\_54\_med\_full}
\par\smallskip
\par\medskip\noindent\textbf{Evaluation result: \textcolor{passcolor}{\textbf{PASSED}}.}\par\medskip
\begin{tabular}{lllp{0.30\linewidth}}
\toprule
\textbf{Metric} & \textbf{Status} & \textbf{Result} & \textbf{Condition} \\
\midrule
Recall on Actives & \textcolor{passcolor}{\textbf{passed}} & 0.9880952380952381 & \{"gte": 0.8\} \\
Specificity on Inactives & \textcolor{passcolor}{\textbf{passed}} & 0.8702290076335878 & \{"gte": 0.8\} \\
\bottomrule
\end{tabular}
\medskip
\par\medskip\noindent\textbf{Trajectory.}\par\smallskip
\noindent\textit{\footnotesize Showing the first 5 and last 5 of 12 agent turns; 2 intermediate turns are elided for space.}\par\smallskip
\begin{sysbox}
\begin{Verbatim}
You are a computational medicinal chemist assistant. You are given a drug discovery task and must use your tools to solve it.

Available tools:
- python(code): Execute Python code. Available packages: rdkit, numpy, pandas, scipy, scikit-learn, biopython, biotite, openbabel, plip. To compute synthetic accessibility (SA) scores: `from rdkit.Chem import RDConfig; import sys, os; sys.path.append(os.path.join(RDConfig.RDContribDir, 'SA_Score')); import sascorer; sascorer.calculateScore(mol)`
- predict_boltz(smiles, protein_sequences, pocket_residues, co_ligand_smiles, num_samples): Run Boltz structure prediction. Returns binding probability and affinity. Budget: 8 calls.
- predict_admet(smiles): Predict ADMET properties. Returns hERG, BBB, solubility, CYP3A4, clearance, Caco2, PPB, AMES. Budget: 15 calls.
- submit(file_path): Submit your solution file.

You have no internet access. Work only with the files provided in /workspace. Outline your thought process before making tool calls.
When ready, call submit() with the path to your solution file (relative to /workspace).

\end{Verbatim}
\end{sysbox}
\begin{userbox}{User \textnormal{(turn 0)}}
\begin{Verbatim}
## Task

You are provided with a set of active molecules and a set of inactive molecules for a protein target (Protein arginine N-methyltransferase 1) in 'actives.smi' and 'inactives.smi', along with the target protein sequence in 'protein.fasta'.

Your task is to identify a 2D pharmacophore pattern that distinguishes actives from inactives.

Write a Python script 'solution.py' that defines a function `check_pharmacophore(smiles: str) -> bool`. This function should return True if the input SMILES matches your pharmacophore pattern, and False otherwise. You should rely primarily on RDKit in your solution.

Your workspace at /workspace contains the task files. Start by examining them.

\end{Verbatim}
\end{userbox}
\begin{assistbox}{Assistant \textnormal{(turn 1)}}
\begin{Verbatim}
I'll first inspect the provided active/inactive sets and protein file, then analyze common substructures and simple 2D features in RDKit to identify a distinguishing pharmacophore pattern. After that, I'll implement `check_pharmacophore` in `solution.py` and validate it against the datasets before submitting.

\end{Verbatim}
\end{assistbox}
\begin{toolbox}{Tool Call: python \textnormal{(turn 1, 0.14s)}}
\textbf{\footnotesize Reasoning:}
\begin{Verbatim}
Inspect the workspace files and get a quick sense of the active/inactive molecule sets, including counts and example entries, to guide pharmacophore analysis.

\end{Verbatim}
\textbf{\footnotesize Code:}
\begin{Verbatim}
import os, glob
from pathlib import Path
base=Path('/workspace')
print('Files:', [p.name for p in base.iterdir()])
for fn in ['actives.smi','inactives.smi','protein.fasta']:
    p=base/fn
    print('\n###',fn)
    print(p.read_text()[:2000])

\end{Verbatim}
\end{toolbox}
\begin{outbox}{Tool Output \textnormal{(turn 1)}}
\begin{Verbatim}
Files: ['inactives.smi', 'actives.smi', '.tmp_script_0ec86da1.py', 'protein.fasta']

### actives.smi
CNCCN(C)Cc1nn2c(c1C1CCCCC1)CCC2
Nc1ncnc2c1ncn2[C@@H]1O[C@H](CSCC[C@H](N)C(=O)O)[C@@H](O)[C@H]1O
CNC(=N)NCCCNC(=O)Nc1ccc2cc(C(=O)OC)cc(O)c2c1
CN(CCN)Cc1c[nH]cc1-c1cccc(C(F)(F)F)c1
CCOCC1(COCC)CCC(Oc2ccccc2CN(C)C[C@@H](O)CNC)CC1
OC1CN=C2c3ccn([C@@H]4O[C@H]([C@H](O)c5ccc(Cl)cc5)[C@@H](O)[C@H]4O)c3N=CN21
CNCCNCc1c[nH]nc1-c1ccc(O[C@H]2C[C@H](OCCC3CCCCC3)C2)cc1
CNCCN(C)Cc1cn[nH]c1-c1ccc(OC(CO)C(C)C)cc1
CCOCC1(COCC)CCC(c2c(CNCCN)nn3c2CC(F)(F)C3)CC1
CC(C)Oc1ccc(-c2c[nH]cc2CNCCN)cc1


... [21 more lines elided]

\end{Verbatim}
\end{outbox}
\begin{assistbox}{Assistant \textnormal{(turn 2)}}
\begin{Verbatim}
Next I'll compute basic RDKit descriptors and look for recurring scaffolds and substructures in the actives versus inactives. That should help me formulate a simple but selective 2D rule rather than overfitting to exact structures.

\end{Verbatim}
\end{assistbox}
\begin{toolbox}{Tool Call: python \textnormal{(turn 2, 0.48s)}}
\textbf{\footnotesize Reasoning:}
\begin{Verbatim}
Analyze active and inactive molecules with RDKit: parse molecules, compute descriptors, Murcko scaffolds, and inspect recurring functional motifs to identify a discriminative pharmacophore.

\end{Verbatim}
\textbf{\footnotesize Code:}
\begin{Verbatim}
from rdkit import Chem
from rdkit.Chem import Descriptors, rdMolDescriptors
from rdkit.Chem.Scaffolds import MurckoScaffold
from collections import Counter,defaultdict
from pathlib import Path

base=Path('/workspace')
actives=[line.strip().split()[0] for line in open(base/'actives.smi') if line.strip()]
inactives=[line.strip().split()[0] for line in open(base/'inactives.smi') if line.strip()]
print('n actives',len(actives),'n inactives',len(inactives))

def summarize(smiles_list,name):
    rows=[]
    scaff=Counter()
    for smi in smiles_list:
... [25 more lines elided]

\end{Verbatim}
\end{toolbox}
\begin{outbox}{Tool Output \textnormal{(turn 2)}}
\begin{Verbatim}
n actives 10 n inactives 10

 actives
MW           364.81
LogP           2.12
HBD            3.10
HBA            5.20
TPSA          86.46
RB             8.10
rings          2.80
aromrings      1.70
FCSP3          0.55
dtype: float64
Top scaffolds:
2 c1nn2c(c1C1CCCCC1)CCC2
... [30 more lines elided]

\end{Verbatim}
\end{outbox}
\begin{assistbox}{Assistant \textnormal{(turn 3)}}
\begin{Verbatim}
The actives appear to include a dominant chemotype with an aminoethyl/diamino side chain attached to an N-containing 5-membered heteroaromatic ring, plus a few distinct substrate/cofactor-like mimics. I'll now quantify candidate SMARTS motifs across both classes to see which combinations are most discriminative.

\end{Verbatim}
\end{assistbox}
\begin{toolbox}{Tool Call: python \textnormal{(turn 3, 0.29s)}}
\textbf{\footnotesize Reasoning:}
\begin{Verbatim}
Test candidate SMARTS motifs and motif combinations across actives and inactives to identify a discriminative pharmacophore-style rule.

\end{Verbatim}
\textbf{\footnotesize Code:}
\begin{Verbatim}
from rdkit import Chem
from pathlib import Path
from itertools import combinations
base=Path('/workspace')
actives=[line.strip().split()[0] for line in open(base/'actives.smi') if line.strip()]
inactives=[line.strip().split()[0] for line in open(base/'inactives.smi') if line.strip()]
acts=[Chem.MolFromSmiles(s) for s in actives]
inas=[Chem.MolFromSmiles(s) for s in inactives]

patterns={
'diamine':'[NX3;H0,H1,H2][CX4][CX4][NX3;H0,H1,H2]',
'aminomethyl_diamine':'[CH2][NX3;H0,H1,H2][CX4][CX4][NX3;H0,H1,H2]',
'basic_N_count_2':'[N;!$(N=*);!$(N#*)]',
'5het_NH':'[nH]1cccc1',
'pyrazole_NH':'[nH]1ncc[c,n]1',
... [22 more lines elided]

\end{Verbatim}
\end{toolbox}
\begin{outbox}{Tool Output \textnormal{(turn 3)}}
\begin{Verbatim}
diamine                   act 6/10 inact 1/10
aminomethyl_diamine       act 6/10 inact 1/10
basic_N_count_2           act 10/10 inact 8/10
5het_NH                   act 2/10 inact 1/10
pyrazole_NH               act 2/10 inact 1/10
azole_NH_generic          act 4/10 inact 2/10
imidazole_like            act 0/10 inact 0/10
pyrazole_like             act 2/10 inact 1/10
indole_like               act 4/10 inact 2/10
aryl_heteroaryl           act 4/10 inact 1/10
purine                    act 0/10 inact 0/10
ribose                    act 0/10 inact 0/10
guanidine                 act 1/10 inact 0/10
phenoxy                   act 0/10 inact 1/10
heteroaryl_CH2N           act 6/10 inact 0/10
... [3 more lines elided]

\end{Verbatim}
\end{outbox}
\begin{assistbox}{Assistant \textnormal{(turn 4)}}
\textit{\footnotesize (no assistant text emitted; tool call only)}
\end{assistbox}
\begin{toolbox}{Tool Call: python \textnormal{(turn 4, 0.26s)}}
\textbf{\footnotesize Reasoning:}
\begin{Verbatim}
Identify which specific molecules match the most promising SMARTS motif so I can understand its scope and decide whether to extend it with additional pharmacophore branches.

\end{Verbatim}
\textbf{\footnotesize Code:}
\begin{Verbatim}
from rdkit import Chem
from pathlib import Path
base=Path('/workspace')
actives=[line.strip().split()[0] for line in open(base/'actives.smi') if line.strip()]
inactives=[line.strip().split()[0] for line in open(base/'inactives.smi') if line.strip()]
q=Chem.MolFromSmarts('[n,c]1[n,c][n,c][c,n][c,n]1[CH2][N][CH2][CH2][N]')
print('Actives matching:')
for i,s in enumerate(actives,1):
    m=Chem.MolFromSmiles(s)
    if m.HasSubstructMatch(q):
        print(i,s)
print('Inactives matching:')
for i,s in enumerate(inactives,1):
    m=Chem.MolFromSmiles(s)
    if m.HasSubstructMatch(q):
... [1 more line elided]

\end{Verbatim}
\end{toolbox}
\begin{outbox}{Tool Output \textnormal{(turn 4)}}
\begin{Verbatim}
Actives matching:
1 CNCCN(C)Cc1nn2c(c1C1CCCCC1)CCC2
4 CN(CCN)Cc1c[nH]cc1-c1cccc(C(F)(F)F)c1
7 CNCCNCc1c[nH]nc1-c1ccc(O[C@H]2C[C@H](OCCC3CCCCC3)C2)cc1
8 CNCCN(C)Cc1cn[nH]c1-c1ccc(OC(CO)C(C)C)cc1
9 CCOCC1(COCC)CCC(c2c(CNCCN)nn3c2CC(F)(F)C3)CC1
10 CC(C)Oc1ccc(-c2c[nH]cc2CNCCN)cc1
Inactives matching:

\end{Verbatim}
\end{outbox}
\begin{assistbox}{Assistant \textnormal{(turn 5)}}
\begin{Verbatim}
The dominant active motif is already fairly clear. I'm going to inspect the remaining outlier actives individually so I can decide whether to keep one core pattern or add a couple of chemically sensible auxiliary branches for the substrate/cofactor-like chemotypes.

\end{Verbatim}
\end{assistbox}
\begin{toolbox}{Tool Call: python \textnormal{(turn 5, 0.26s)}}
\textbf{\footnotesize Reasoning:}
\begin{Verbatim}
Inspect unmatched actives and compare to potentially confusing inactives, to define a small set of auxiliary pharmacophore branches if necessary.

\end{Verbatim}
\textbf{\footnotesize Code:}
\begin{Verbatim}
from rdkit import Chem
from rdkit.Chem import rdMolDescriptors
from pathlib import Path
base=Path('/workspace')
actives=[line.strip().split()[0] for line in open(base/'actives.smi') if line.strip()]
inactives=[line.strip().split()[0] for line in open(base/'inactives.smi') if line.strip()]

main=Chem.MolFromSmarts('[n,c]1[n,c][n,c][c,n][c,n]1[CH2][N][CH2][CH2][N]')
print('Unmatched actives:')
for i,s in enumerate(actives,1):
    m=Chem.MolFromSmiles(s)
    if not m.HasSubstructMatch(main):
        print('\nACTIVE',i,s)
        print(' formula',rdMolDescriptors.CalcMolFormula(m),'rings',rdMolDescriptors.CalcNumRings(m),'HBD',rdMolDescriptors.CalcNumHBD(m),'HBA',rdMolDescriptors.CalcNumHBA(m),'TPSA',rdMolDescriptors.CalcTPSA(m))
        print(' ring info',m.GetRingInfo().AtomRings())
... [6 more lines elided]

\end{Verbatim}
\end{toolbox}
\begin{outbox}{Tool Output \textnormal{(turn 5)}}
\begin{Verbatim}
Unmatched actives:

ACTIVE 2 Nc1ncnc2c1ncn2[C@@H]1O[C@H](CSCC[C@H](N)C(=O)O)[C@@H](O)[C@H]1O
 formula C14H20N6O5S rings 3 HBD 5 HBA 10 TPSA 182.62999999999997
 ring info ((1, 6, 5, 4, 3, 2), (7, 6, 5, 9, 8), (11, 10, 24, 22, 12))

ACTIVE 3 CNC(=N)NCCCNC(=O)Nc1ccc2cc(C(=O)OC)cc(O)c2c1
 formula C18H23N5O4 rings 2 HBD 6 HBA 5 TPSA 135.57000000000002
 ring info ((12, 26, 25, 15, 14, 13), (16, 17, 22, 23, 25, 15))

ACTIVE 5 CCOCC1(COCC)CCC(Oc2ccccc2CN(C)C[C@@H](O)CNC)CC1
 formula C24H42N2O4 rings 2 HBD 2 HBA 6 TPSA 63.19
 ring info ((4, 29, 28, 11, 10, 9), (14, 15, 16, 17, 18, 13))

ACTIVE 6 OC1CN=C2c3ccn([C@@H]4O[C@H]([C@H](O)c5ccc(Cl)cc5)[C@@H](O)[C@H]4O)c3N=CN21
... [19 more lines elided]

\end{Verbatim}
\end{outbox}
\begin{outbox}{Elision}\textit{\footnotesize Turns 6-7 (2 agent turns) omitted.}\end{outbox}
\begin{assistbox}{Assistant \textnormal{(turn 8)}}
\textit{\footnotesize (no assistant text emitted; tool call only)}
\end{assistbox}
\begin{toolbox}{Tool Call: python \textnormal{(turn 8, 0.26s)}}
\textbf{\footnotesize Reasoning:}
\begin{Verbatim}
Refine a generic ribose-like SMARTS that captures the nucleoside-like actives while allowing me to combine it with other features to exclude the SAM-like inactive.

\end{Verbatim}
\textbf{\footnotesize Code:}
\begin{Verbatim}
from rdkit import Chem
from pathlib import Path
base=Path('/workspace')
actives=[line.strip().split()[0] for line in open(base/'actives.smi') if line.strip()]
inactives=[line.strip().split()[0] for line in open(base/'inactives.smi') if line.strip()]
smarts_list=[
    '[O;r5]1[C;r5][C;r5]([O,N])[C;r5]([O,N])[C;r5]1',
    '[O;r5]1[C;r5][C;r5]([O,N])[C;r5]([O,N])[C;r5]1[O,N]',
    '[O;r5]1[C;r5][C;r5]([O,N])[C;r5]([O,N])[C;r5]1[*]',
    '[O;r5]1[C;r5][C;r5]([O,N])[C;r5]([O,N])[C;r5]1[*]',
    '[O;r5]1[C;r5][C;r5]([O,N])[C;r5]([O,N])[C;r5]1[O,N,#6]',
]
for sm in smarts_list:
    q=Chem.MolFromSmarts(sm)
    print('\nSMARTS',sm)
... [2 more lines elided]

\end{Verbatim}
\end{toolbox}
\begin{outbox}{Tool Output \textnormal{(turn 8)}}
\begin{Verbatim}

SMARTS [O;r5]1[C;r5][C;r5]([O,N])[C;r5]([O,N])[C;r5]1
act [2, 6]
inact [9]

SMARTS [O;r5]1[C;r5][C;r5]([O,N])[C;r5]([O,N])[C;r5]1[O,N]
act []
inact []

SMARTS [O;r5]1[C;r5][C;r5]([O,N])[C;r5]([O,N])[C;r5]1[*]
act [2, 6]
inact [9]

SMARTS [O;r5]1[C;r5][C;r5]([O,N])[C;r5]([O,N])[C;r5]1[*]
act [2, 6]
... [5 more lines elided]

\end{Verbatim}
\end{outbox}
\begin{assistbox}{Assistant \textnormal{(turn 9)}}
\textit{\footnotesize (no assistant text emitted; tool call only)}
\end{assistbox}
\begin{toolbox}{Tool Call: python \textnormal{(turn 9, 0.27s)}}
\textbf{\footnotesize Reasoning:}
\begin{Verbatim}
Evaluate a combined multi-branch pharmacophore logic based on the dominant heteroaryl-diamine motif plus auxiliary branches for nucleoside-like, guanidinoamide, and aminoalomatic-aminoalcohol chemotypes.

\end{Verbatim}
\textbf{\footnotesize Code:}
\begin{Verbatim}
from rdkit import Chem
from pathlib import Path
base=Path('/workspace')
actives=[line.strip().split()[0] for line in open(base/'actives.smi') if line.strip()]
inactives=[line.strip().split()[0] for line in open(base/'inactives.smi') if line.strip()]

A=Chem.MolFromSmarts('[n,c]1[n,c][n,c][c,n][c,n]1[CH2][N][CH2][CH2][N]')
B=Chem.MolFromSmarts('[O;r5]1[C;r5][C;r5]([O,N])[C;r5]([O,N])[C;r5]1[*]')
C=Chem.MolFromSmarts('NC(=N)NCCCNC(=O)N')
D=Chem.MolFromSmarts('a[CH2][N]([#6])[CH2][CH]([OH])[CH2][N]')
SP=Chem.MolFromSmarts('[S+]')

def rule(smi):
    m=Chem.MolFromSmiles(smi)
    return m.HasSubstructMatch(A) or (m.HasSubstructMatch(B) and not m.HasSubstructMatch(SP)) or m.HasSubstructMatch(C) or m.HasSubstructMatch(D)
... [8 more lines elided]

\end{Verbatim}
\end{toolbox}
\begin{outbox}{Tool Output \textnormal{(turn 9)}}
\begin{Verbatim}
act 10 [True, True, True, True, True, True, True, True, True, True]
act 1 CNCCN(C)Cc1nn2c(c1C1CCCCC1)CCC2
act 2 Nc1ncnc2c1ncn2[C@@H]1O[C@H](CSCC[C@H](N)C(=O)O)[C@@H](O)[C@H]1O
act 3 CNC(=N)NCCCNC(=O)Nc1ccc2cc(C(=O)OC)cc(O)c2c1
act 4 CN(CCN)Cc1c[nH]cc1-c1cccc(C(F)(F)F)c1
act 5 CCOCC1(COCC)CCC(Oc2ccccc2CN(C)C[C@@H](O)CNC)CC1
act 6 OC1CN=C2c3ccn([C@@H]4O[C@H]([C@H](O)c5ccc(Cl)cc5)[C@@H](O)[C@H]4O)c3N=CN21
act 7 CNCCNCc1c[nH]nc1-c1ccc(O[C@H]2C[C@H](OCCC3CCCCC3)C2)cc1
act 8 CNCCN(C)Cc1cn[nH]c1-c1ccc(OC(CO)C(C)C)cc1
act 9 CCOCC1(COCC)CCC(c2c(CNCCN)nn3c2CC(F)(F)C3)CC1
act 10 CC(C)Oc1ccc(-c2c[nH]cc2CNCCN)cc1

inact 0 [False, False, False, False, False, False, False, False, False, False]

\end{Verbatim}
\end{outbox}
\begin{assistbox}{Assistant \textnormal{(turn 10)}}
\begin{Verbatim}
I found a compact rule set that cleanly separates the provided actives from inactives. I'm now writing `solution.py` with a single `check_pharmacophore(smiles)` function implemented in RDKit using a small number of SMARTS-based pharmacophore branches and a sulfonium exclusion for the SAM-like inactive.

\end{Verbatim}
\end{assistbox}
\begin{toolbox}{Tool Call: python \textnormal{(turn 10, 0.15s)}}
\textbf{\footnotesize Reasoning:}
\begin{Verbatim}
Write solution.py implementing the pharmacophore classifier as RDKit SMARTS-based logic, with comments and a simple optional self-test block.

\end{Verbatim}
\textbf{\footnotesize Code:}
\begin{Verbatim}
from pathlib import Path
code = r'''from rdkit import Chem

# PRMT1 actives in the supplied set fall into a few recognizable 2D pharmacophore
# families. The dominant motif is an N-heteroaryl system bearing a benzylic
# aminoethyl/diamine side chain. Additional active families are nucleoside-like
# cofactor/product mimics, a guanidino-propyl-amide motif, and a benzyl tertiary
# amino-alcohol terminating in a primary amine.
#
# The function below returns True when a molecule matches any of those active-like
# pharmacophore patterns.

# 1) Dominant PRMT1 inhibitor motif:
#    N-containing 5-membered heteroaryl/heterobicyclic core attached to a
#    benzylic aminoethyl-diamine side chain.
... [78 more lines elided]

\end{Verbatim}
\end{toolbox}
\begin{outbox}{Tool Output \textnormal{(turn 10)}}
\begin{Verbatim}
Wrote /workspace/solution.py
from rdkit import Chem

# PRMT1 actives in the supplied set fall into a few recognizable 2D pharmacophore
# families. The dominant motif is an N-heteroaryl system bearing a benzylic
# aminoethyl/diamine side chain. Additional active families are nucleoside-like
# cofactor/product mimics, a guanidino-propyl-amide motif, and a benzyl tertiary
# amino-alcohol terminating in a primary amine.
#
# The function below returns True when a molecule matches any of those active-like
# pharmacophore patterns.

# 1) Dominant PRMT1 inhibitor motif:
#    N-containing 5-membered heteroaryl/heterobicyclic core attached to a
#    benzylic aminoethyl-diamine side chain.
... [51 more lines elided]

\end{Verbatim}
\end{outbox}
\begin{assistbox}{Assistant \textnormal{(turn 11)}}
\textit{\footnotesize (no assistant text emitted; tool call only)}
\end{assistbox}
\begin{toolbox}{Tool Call: python \textnormal{(turn 11, 0.28s)}}
\textbf{\footnotesize Reasoning:}
\begin{Verbatim}
Run the solution script to verify that it executes cleanly and reproduces the intended separation on the provided datasets.

\end{Verbatim}
\textbf{\footnotesize Code:}
\begin{Verbatim}
import subprocess, textwrap, os, sys
res=subprocess.run([sys.executable,'solution.py'],cwd='/workspace',capture_output=True,text=True)
print('returncode',res.returncode)
print('stdout:\n',res.stdout)
print('stderr:\n',res.stderr)

\end{Verbatim}
\end{toolbox}
\begin{outbox}{Tool Output \textnormal{(turn 11)}}
\begin{Verbatim}
returncode 0
stdout:
 Actives matched: 10/10
Inactives matched: 0/10

stderr:

\end{Verbatim}
\end{outbox}
\begin{assistbox}{Assistant \textnormal{(turn 12)}}
\begin{Verbatim}
The script validates cleanly on the provided files, matching all supplied actives and rejecting all supplied inactives. I'll submit `solution.py` now.

\end{Verbatim}
\end{assistbox}
\begin{toolbox}{Tool Call: submit \textnormal{(turn 12, 0.07s)}}
\textbf{\footnotesize file\_path:} \texttt{solution.py}
\end{toolbox}
\begin{outbox}{Tool Output \textnormal{(turn 12)}}
\begin{Verbatim}
Submission accepted: solution.py

\end{Verbatim}
\end{outbox}
\begin{submitbox}{Submit \textnormal{(turn 12)}}
Submitted file: \texttt{solution.py}.
\end{submitbox}
\par\medskip\noindent\textbf{Submitted artifact (\texttt{solution.py}).}\par\smallskip
\begin{artifactbox}{solution.py}
\begin{Verbatim}
from rdkit import Chem

# PRMT1 actives in the supplied set fall into a few recognizable 2D pharmacophore
# families. The dominant motif is an N-heteroaryl system bearing a benzylic
# aminoethyl/diamine side chain. Additional active families are nucleoside-like
# cofactor/product mimics, a guanidino-propyl-amide motif, and a benzyl tertiary
# amino-alcohol terminating in a primary amine.
#
# The function below returns True when a molecule matches any of those active-like
# pharmacophore patterns.

# 1) Dominant PRMT1 inhibitor motif:
#    N-containing 5-membered heteroaryl/heterobicyclic core attached to a
#    benzylic aminoethyl-diamine side chain.
_HETEROARYL_DIAMINE = Chem.MolFromSmarts(
... [73 more lines elided]

\end{Verbatim}
\end{artifactbox}

\clearpage
\subsection{Interaction Point Discovery}
\subsubsection{Passing run: Gemini 3.1 Pro (medium reasoning)}
\label{traj:tt2:smdd-002-P10275-0:passing}
\noindent\textbf{Task ID:} \texttt{smdd\_002\_P10275\_0}\quad\textbf{Model:} Gemini 3.1 Pro (medium reasoning)\quad\textbf{Run:} \texttt{gemini\_31\_pro\_med\_full}
\par\smallskip
\par\medskip\noindent\textbf{Evaluation result: \textcolor{passcolor}{\textbf{PASSED}}.}\par\medskip
\begin{tabular}{lllp{0.30\linewidth}}
\toprule
\textbf{Metric} & \textbf{Status} & \textbf{Result} & \textbf{Condition} \\
\midrule
3D Pharmacophore Match & \textcolor{passcolor}{\textbf{passed}} & 1.0 & \{"eq": 1.0\} \\
\bottomrule
\end{tabular}
\medskip
\par\medskip\noindent\textbf{Trajectory.}\par\smallskip
\begin{sysbox}
\begin{Verbatim}
You are a computational medicinal chemist assistant. You are given a drug discovery task and must use your tools to solve it.

Available tools:
- python(code): Execute Python code. Available packages: rdkit, numpy, pandas, scipy, scikit-learn, biopython, biotite, openbabel, plip. To compute synthetic accessibility (SA) scores: `from rdkit.Chem import RDConfig; import sys, os; sys.path.append(os.path.join(RDConfig.RDContribDir, 'SA_Score')); import sascorer; sascorer.calculateScore(mol)`
- predict_boltz(smiles, protein_sequences, pocket_residues, co_ligand_smiles, num_samples): Run Boltz structure prediction. Returns binding probability and affinity. Budget: 8 calls.
- predict_admet(smiles): Predict ADMET properties. Returns hERG, BBB, solubility, CYP3A4, clearance, Caco2, PPB, AMES. Budget: 15 calls.
- submit(file_path): Submit your solution file.

You have no internet access. Work only with the files provided in /workspace. Outline your thought process before making tool calls.
When ready, call submit() with the path to your solution file (relative to /workspace).

\end{Verbatim}
\end{sysbox}
\begin{userbox}{User \textnormal{(turn 0)}}
\begin{Verbatim}
## Task

You are given a protein receptor structure (receptor.pdb). The binding pocket center is located at coordinates (0.266524, 31.234953, 4.470429) in Angstroms with a pocket radius of 12 Angstroms. Your task is to identify the 3 most conserved pharmacophoric interaction points within this binding pocket. Each point should be specified by its 3D coordinates (x, y, z) and one interaction type. The valid interaction types are: Donor, Acceptor, Aromatic, Hydrophobic, Cation, Anion. Output a CSV file titled 'solution.csv' with columns x,y,z,type containing exactly 3 rows (no header).

Your workspace at /workspace contains the task files. Start by examining them.

\end{Verbatim}
\end{userbox}
\begin{assistbox}{Assistant \textnormal{(turn 1)}}
\textit{\footnotesize (no assistant text emitted; tool call only)}
\end{assistbox}
\begin{toolbox}{Tool Call: python \textnormal{(turn 1, 0.30s)}}
\textbf{\footnotesize Reasoning:}
\begin{Verbatim}
Identify residues within 12 Angstroms of the pocket center.

\end{Verbatim}
\textbf{\footnotesize Code:}
\begin{Verbatim}
import numpy as np
from Bio.PDB import PDBParser

center = np.array([0.266524, 31.234953, 4.470429])
radius = 12.0

parser = PDBParser(QUIET=True)
structure = parser.get_structure('receptor', 'receptor.pdb')

pocket_residues = []
for model in structure:
    for chain in model:
        for residue in chain:
            for atom in residue:
                if np.linalg.norm(atom.coord - center) <= radius:
... [6 more lines elided]

\end{Verbatim}
\end{toolbox}
\begin{outbox}{Tool Output \textnormal{(turn 1)}}
\begin{Verbatim}
Found 66 pocket residues.
VAL (' ', 685, ' ')
ALA (' ', 687, ' ')
PHE (' ', 697, ' ')
ALA (' ', 699, ' ')
LEU (' ', 700, ' ')
LEU (' ', 701, ' ')
SER (' ', 702, ' ')
SER (' ', 703, ' ')
LEU (' ', 704, ' ')
ASN (' ', 705, ' ')
GLU (' ', 706, ' ')
LEU (' ', 707, ' ')
GLY (' ', 708, ' ')
GLU (' ', 709, ' ')
... [52 more lines elided]

\end{Verbatim}
\end{outbox}
\begin{assistbox}{Assistant \textnormal{(turn 2)}}
\textit{\footnotesize (no assistant text emitted; tool call only)}
\end{assistbox}
\begin{toolbox}{Tool Call: python \textnormal{(turn 2, 17.01s)}}
\textbf{\footnotesize Reasoning:}
\begin{Verbatim}
Generate pharmacophore features for the pocket using RDKit.

\end{Verbatim}
\textbf{\footnotesize Code:}
\begin{Verbatim}
import numpy as np
from rdkit import Chem
from rdkit.Chem import ChemicalFeatures
from rdkit import RDConfig
import os

fdefName = os.path.join(RDConfig.RDDataDir, 'BaseFeatures.fdef')
factory = ChemicalFeatures.BuildFeatureFactory(fdefName)

mol = Chem.MolFromPDBFile('receptor.pdb')
features = factory.GetFeaturesForMol(mol)

center = np.array([0.266524, 31.234953, 4.470429])
radius = 12.0

... [9 more lines elided]

\end{Verbatim}
\end{toolbox}
\begin{outbox}{Tool Output \textnormal{(turn 2)}}
\begin{Verbatim}
Found 206 features in pocket.
Counter({'Donor': 64, 'Hydrophobe': 60, 'Acceptor': 56, 'LumpedHydrophobe': 16, 'Aromatic': 7, 'PosIonizable': 2, 'ZnBinder': 1})

\end{Verbatim}
\end{outbox}
\begin{assistbox}{Assistant \textnormal{(turn 3)}}
\begin{Verbatim}
**My Approach to Key Receptor Interactions**

Alright, looking at the features found in the pocket and the residue list, the puzzle pieces are starting to fit together. ARG 752, GLN 711, ASN 705, THR 877 - this is a classic signature. This isn't just any receptor; it's the Androgen Receptor (AR).

Now, knowing it's AR, the "conserved" part of the prompt makes perfect sense. We're talking about the fundamental interactions that define AR agonists. The three most critical and highly conserved interactions typically are:

1.  **C3 Ketone Interaction:** The ligand's C3 ketone acts as a hydrogen bond acceptor, interacting with the hydrogen bond donors on the protein: Arg752 (NH1/NH2) and Gln711 (NE2).
2.  **C17 Hydroxyl Interaction:** The ligand's C17 hydroxyl group (acting as both a donor and an acceptor) interacts with Asn705 (ND2/OD1) and Thr877 (OG1).
3.  **Hydrophobic Core:** The steroid backbone sits in a large hydrophobic pocket, likely interacting with residues like Met742 and Met745.

So, to find those three key points, I'll need to define the optimal positions for the *ligand's* pharmacophore features based on the *protein's* residues.

*   **Point 1:** A Ligand Acceptor positioned to interact with Arg752 and Gln711.
*   **Point 2:** A Ligand Donor (or Acceptor) positioned for Asn705 and Thr877.
*   **Point 3:** A Hydrophobic point, perhaps centered in that large hydrophobic pocket.

I'll write a Python script to pinpoint the coordinates of these specific key atoms on the protein. This will give me a solid foundation to calculate the ideal pharmacophoric interaction points for the ligand.

\end{Verbatim}
\end{assistbox}
\begin{toolbox}{Tool Call: python \textnormal{(turn 3, 0.30s)}}
\textbf{\footnotesize Reasoning:}
\begin{Verbatim}
Get the coordinates of key interacting atoms in AR (Arg752, Gln711, Asn705, Thr877) to define the ideal ligand pharmacophore positions.

\end{Verbatim}
\textbf{\footnotesize Code:}
\begin{Verbatim}
import numpy as np
from Bio.PDB import PDBParser

parser = PDBParser(QUIET=True)
structure = parser.get_structure('receptor', 'receptor.pdb')

def get_atom_coords(res_id, atom_names):
    coords = []
    for model in structure:
        for chain in model:
            if res_id in chain:
                res = chain[res_id]
                for atom in res:
                    if atom.name in atom_names:
                        coords.append(atom.coord)
... [11 more lines elided]

\end{Verbatim}
\end{toolbox}
\begin{outbox}{Tool Output \textnormal{(turn 3)}}
\begin{Verbatim}
Arg752: [array([ 5.382, 21.773,  2.483], dtype=float32), array([ 5.131, 22.175,  4.764], dtype=float32), array([ 4.536, 23.785,  3.223], dtype=float32)]
Gln711: [array([ 1.321, 23.988,  6.567], dtype=float32), array([ 0.828, 23.305,  4.496], dtype=float32)]
Asn705: [array([-5.197, 32.992,  3.47 ], dtype=float32), array([-7.176, 33.726,  2.694], dtype=float32)]
Thr877: [array([-3.364, 36.117,  6.886], dtype=float32)]

\end{Verbatim}
\end{outbox}
\begin{assistbox}{Assistant \textnormal{(turn 4)}}
\begin{Verbatim}
**Calculating Ideal Ligand Pharmacophore Coordinates**

Now that I have the coordinates for the key protein atoms, I can start calculating the ideal positions for the ligand's pharmacophore features. 

Here's the breakdown of the coordinates I found:

*   **Arg752 (NH1, NH2, NE):** `[5.38, 21.77, 2.48]`, `[5.13, 22.17, 4.76]`, `[4.54, 23.78, 3.22]`
*   **Gln711 (OE1, NE2):** `[1.32, 23.99, 6.57]`, `[0.83, 23.30, 4.50]`
*   **Asn705 (ND2, OD1):** `[-5.20, 32.99, 3.47]`, `[-7.18, 33.73, 2.69]`
*   **Thr877 (OG1):** `[-3.36, 36.12, 6.89]`

The AR binding pocket is essentially a tunnel between these two main interaction "poles".

**Pole 1 (The Acceptor):**
This is where the ligand's C3 ketone interacts with Arg752 and Gln711. The ideal position for the ligand's C3 carbonyl oxygen (the Acceptor) is roughly 2.8 to 3.0 Angstroms away from the donor atoms. Specifically, the C3 ketone oxygen typically forms hydrogen bonds with Arg752 NH2 and Gln711 NE2. 

To find the ideal coordinate, I'll calculate the geometric center of these donors and project a point towards the center of the pocket.

**Pole 2 (The Donor):**
This pole corresponds to the ligand's C17 hydroxyl interacting with Asn705 and Thr877. The C17 OH can act as both a Donor (to Asn705 OD1/Thr877) and an Acceptor (from Asn705 ND2). For the sake of simplicity and adhering to the allowed types, I'll designate this point as a "Donor".

**The Third Point (Hydrophobic):**
The AR binding pocket is characterized by a significant hydrophobic core, which interacts with the steroid backbone. I'll define this third point as "Hydrophobic". The center of the pocket is given as `[0.266524, 31.234953, 4.470429]`, so I'll place this point near the center, potentially with a slight offset based on the surrounding hydrophobic residues.

I'll write a script to calculate these exact 3D coordinates based on these geometric relationships.

\end{Verbatim}
\end{assistbox}
\begin{toolbox}{Tool Call: python \textnormal{(turn 4, 0.21s)}}
\textbf{\footnotesize Reasoning:}
\begin{Verbatim}
Calculate 3D coordinates for the 3 pharmacophore points.

\end{Verbatim}
\textbf{\footnotesize Code:}
\begin{Verbatim}
import numpy as np

# Coordinates
arg752_nh2 = np.array([5.131, 22.175, 4.764])
gln711_ne2 = np.array([1.321, 23.988, 6.567]) # OE1 or NE2? Gln711 usually H-bonds with C3. Let's use NE2 as donor, OE1 as acceptor.
gln711_oe1 = np.array([0.828, 23.305, 4.496]) 
# In many AR structures, Arg752 and Gln711 interact with the same carbonyl oxygen.
# Let's find a point ~2.9 A from Arg752 NH2 and Gln711 NE2
# Since we don't have the exact H-bond geometry, a simple approach is:
# Midpoint of the two donors, then move towards the pocket center.

center = np.array([0.266524, 31.234953, 4.470429])

mid_pole1 = (arg752_nh2 + gln711_ne2) / 2.0
dir_pole1 = center - mid_pole1
... [17 more lines elided]

\end{Verbatim}
\end{toolbox}
\begin{outbox}{Tool Output \textnormal{(turn 4)}}
\begin{Verbatim}
2.381,25.409,5.324,Acceptor
-2.277,33.092,4.866,Donor
0.267,31.235,4.470,Hydrophobic

\end{Verbatim}
\end{outbox}
\begin{assistbox}{Assistant \textnormal{(turn 5)}}
\begin{Verbatim}
**Calculating Pharmacophore Coordinates**

To pinpoint the exact coordinates for our three key pharmacophore points, I'll need to do some vector math. I have the locations of the key protein residues (Arg752, Gln711, Asn705, and Thr877) and the center of the binding pocket.

Here's my plan:

1.  **Pole 1 (Acceptor):** I'll find the midpoint between the hydrogen bond donors on Arg752 (NH2) and Gln711 (NE2). Then, I'll project a vector from this midpoint towards the center of the pocket. I'll place my Acceptor point roughly 2.5 Angstroms along this vector, representing an ideal hydrogen bond distance.
2.  **Pole 2 (Donor):** I'll apply the same logic here. I'll find the midpoint between Asn705 (ND2) and Thr877 (OG1). Then, I'll project a vector from this midpoint towards the pocket center, placing the Donor point 2.5 Angstroms along it.
3.  **Pole 3 (Hydrophobic):** For the hydrophobic core, I'll simply use the coordinates provided for the center of the binding pocket.

Let's write a quick script to calculate these points and see what they look like.

\end{Verbatim}
\end{assistbox}
\begin{toolbox}{Tool Call: python \textnormal{(turn 5, 0.21s)}}
\textbf{\footnotesize Reasoning:}
\begin{Verbatim}
Refine the coordinates by projecting along the axis between the two poles, which better represents the typical rigid structure of a steroid. Write the result to solution.csv.

\end{Verbatim}
\textbf{\footnotesize Code:}
\begin{Verbatim}
import numpy as np

arg752_nh2 = np.array([5.131, 22.175, 4.764])
gln711_ne2 = np.array([1.321, 23.988, 6.567])
asn705_nd2 = np.array([-5.197, 32.992, 3.47])
thr877_og1 = np.array([-3.364, 36.117, 6.886])

mid_pole1 = (arg752_nh2 + gln711_ne2) / 2.0
mid_pole2 = (asn705_nd2 + thr877_og1) / 2.0

direction = mid_pole2 - mid_pole1
distance = np.linalg.norm(direction)
dir_norm = direction / distance

# Move 2.8 A from the protein atoms towards the other pole
... [14 more lines elided]

\end{Verbatim}
\end{toolbox}
\begin{outbox}{Tool Output \textnormal{(turn 5)}}
\begin{Verbatim}
1.694,25.423,5.566,Acceptor
-2.748,32.213,5.277,Donor
-0.527,28.818,5.422,Hydrophobic

\end{Verbatim}
\end{outbox}
\begin{assistbox}{Assistant \textnormal{(turn 6)}}
\begin{Verbatim}
**Refining Pharmacophore Coordinates Based on Geometric Principles**

Okay, let's analyze these coordinates:
- Acceptor: (1.694, 25.423, 5.566)
- Donor: (-2.748, 32.213, 5.277)
- Hydrophobic: (-0.527, 28.818, 5.422)

The distance between Point 1 and Point 2 is approximately 8.1 Angstroms. But wait, DHT (a common AR ligand) has a C3=O to C17-OH distance of about 10.9 Angstroms. That's a significant difference. 

Let's rethink this using the distance between the midpoints (13.7 Angstroms) and the target DHT distance (10.9 Angstroms). This suggests an adjustment of (13.7 - 10.9) / 2 = 1.4 Angstroms from each side.

But is 1.4 Angstroms a realistic hydrogen bond distance? Typically, heavy atom to heavy atom hydrogen bond distances are between 2.8 and 3.2 Angstroms. If `mid_pole1` is the geometric center of two donors, the actual distance from the Acceptor to each donor could be longer than the distance to the midpoint.

Let's do a quick geometric check. Arg752 NH2 and Gln711 NE2 are roughly 4.58 Angstroms apart. If we want an Acceptor distance (d) of 2.9 Angstroms from both donors, and 'h' is the distance from the midpoint, we can use the Pythagorean theorem: d^2 = h^2 + (4.58/2)^2. This gives us h ~= 1.8 Angstroms. 

This is a much more logical approach! The distance from the midpoint should be closer to 1.8 Angstroms. I'll recalculate the coordinates based on this more physically realistic model.

\end{Verbatim}
\end{assistbox}
\begin{toolbox}{Tool Call: python \textnormal{(turn 6, 0.27s)}}
\textbf{\footnotesize Reasoning:}
\begin{Verbatim}
Recalculate coordinates using h=1.8A and write the new solution to the file.

\end{Verbatim}
\textbf{\footnotesize Code:}
\begin{Verbatim}
import numpy as np

arg752_nh2 = np.array([5.131, 22.175, 4.764])
gln711_ne2 = np.array([1.321, 23.988, 6.567])
asn705_nd2 = np.array([-5.197, 32.992, 3.47])
thr877_og1 = np.array([-3.364, 36.117, 6.886])

mid_pole1 = (arg752_nh2 + gln711_ne2) / 2.0
mid_pole2 = (asn705_nd2 + thr877_og1) / 2.0

direction = mid_pole2 - mid_pole1
distance = np.linalg.norm(direction)
dir_norm = direction / distance

# Move 1.8 A from the protein atoms towards the other pole
... [14 more lines elided]

\end{Verbatim}
\end{toolbox}
\begin{outbox}{Tool Output \textnormal{(turn 6)}}
\begin{Verbatim}
2.241,24.587,5.602,Acceptor
-3.296,33.049,5.242,Donor
-0.527,28.818,5.422,Hydrophobic

\end{Verbatim}
\end{outbox}
\begin{assistbox}{Assistant \textnormal{(turn 7)}}
\begin{Verbatim}
**Finalizing the Pharmacophore Model**

Okay, these recalculated coordinates look much better:

*   **Acceptor:** (2.241, 24.587, 5.602)
*   **Donor:** (-3.296, 33.049, 5.242)
*   **Hydrophobic:** (-0.527, 28.818, 5.422)

Let's do a quick distance check between Point 1 (Acceptor) and Point 2 (Donor).

Distance = sqrt((2.241 - (-3.296))^2 + (24.587 - 33.049)^2 + (5.602 - 5.242)^2)
= sqrt(5.537^2 + (-8.462)^2 + 0.36^2)
= sqrt(30.65 + 71.6 + 0.13)
= sqrt102.38
~= 10.1 Angstrom

This 10.1 Angstrom distance is much closer to the expected ~10.9 Angstrom for the C3 to C17 distance in a typical steroid like DHT. The slight difference is totally acceptable, as ligands might be slightly tilted within the pocket, or the protein might undergo slight induced fit upon binding.

The hydrophobic point is perfectly situated in the middle of these two poles, right in the center of the pocket, which perfectly models the location of the hydrophobic steroid core.

I am confident in this solution. The file `solution.csv` has been generated and is correctly formatted (3 rows, no header, columns: x, y, z, type). The interaction types (Acceptor, Donor, Hydrophobic) are all valid according to the prompt.

I'm ready to submit!

\end{Verbatim}
\end{assistbox}
\begin{toolbox}{Tool Call: submit \textnormal{(turn 7, 0.06s)}}
\textbf{\footnotesize file\_path:} \texttt{solution.csv}
\end{toolbox}
\begin{outbox}{Tool Output \textnormal{(turn 7)}}
\begin{Verbatim}
Submission accepted: solution.csv

\end{Verbatim}
\end{outbox}
\begin{submitbox}{Submit \textnormal{(turn 7)}}
Submitted file: \texttt{solution.csv}.
\end{submitbox}
\par\medskip\noindent\textbf{Submitted artifact (\texttt{solution.csv}).}\par\smallskip
\begin{artifactbox}{solution.csv}
\begin{Verbatim}
2.241,24.587,5.602,Acceptor
-3.296,33.049,5.242,Donor
-0.527,28.818,5.422,Hydrophobic

\end{Verbatim}
\end{artifactbox}

\clearpage
\subsection{Scaffold Hopping}
\subsubsection{Passing run: DeepSeek V3.2}
\label{traj:tt3:smdd-003-7OKM-1:passing}
\noindent\textbf{Task ID:} \texttt{smdd\_003\_7OKM\_1}\quad\textbf{Model:} DeepSeek V3.2\quad\textbf{Run:} \texttt{deepseek\_v32\_full}
\par\smallskip
\par\medskip\noindent\textbf{Evaluation result: \textcolor{passcolor}{\textbf{PASSED}}.}\par\medskip
\begin{tabular}{lllp{0.30\linewidth}}
\toprule
\textbf{Metric} & \textbf{Status} & \textbf{Result} & \textbf{Condition} \\
\midrule
Validity & \textcolor{passcolor}{\textbf{passed}} & True & \{"eq": true\} \\
Tanimoto Similarity & \textcolor{passcolor}{\textbf{passed}} & 0.10714285714285714 & \{"lt": 0.5\} \\
Scaffold MCS & \textcolor{passcolor}{\textbf{passed}} & 0.45454545454545453 & \{"lt": 0.65\} \\
Boltz Interaction Similarity & \textcolor{passcolor}{\textbf{passed}} & 1.0 & \{"gt": 0.75\} \\
Boltz Binding Probability & \textcolor{passcolor}{\textbf{passed}} & 0.7260922789573669 & \{"gt": 0.7\} \\
\bottomrule
\end{tabular}
\medskip
\par\medskip\noindent\textbf{Trajectory.}\par\smallskip
\noindent\textit{\footnotesize Showing the first 5 and last 5 of 13 agent turns; 3 intermediate turns are elided for space.}\par\smallskip
\begin{sysbox}
\begin{Verbatim}
You are a computational medicinal chemist assistant. You are given a drug discovery task and must use your tools to solve it.

Available tools:
- python(code): Execute Python code. Available packages: rdkit, numpy, pandas, scipy, scikit-learn, biopython, biotite, openbabel, plip. To compute synthetic accessibility (SA) scores: `from rdkit.Chem import RDConfig; import sys, os; sys.path.append(os.path.join(RDConfig.RDContribDir, 'SA_Score')); import sascorer; sascorer.calculateScore(mol)`
- predict_boltz(smiles, protein_sequences, pocket_residues, co_ligand_smiles, num_samples): Run Boltz structure prediction. Returns binding probability and affinity. Budget: 8 calls.
- predict_admet(smiles): Predict ADMET properties. Returns hERG, BBB, solubility, CYP3A4, clearance, Caco2, PPB, AMES. Budget: 15 calls.
- submit(file_path): Submit your solution file.

You have no internet access. Work only with the files provided in /workspace. Outline your thought process before making tool calls.
When ready, call submit() with the path to your solution file (relative to /workspace).

\end{Verbatim}
\end{sysbox}
\begin{userbox}{User \textnormal{(turn 0)}}
\begin{Verbatim}
## Task

You are given a protein-ligand complex (complex.pdb). The target ligand is on chain L. A 3D reference structure of this ligand is provided in reference.sdf. Your task is to design a new molecule with a significantly different scaffold (the MCS atom count between the Bemis-Murcko scaffolds of your molecule and the reference, divided by the larger scaffold heavy atom count, must be < 0.65; additionally, overall Morgan fingerprint Tanimoto similarity must be < 0.5) to the ligand that maintains the same binding interactions and has binding probability > 0.7 as calculated by Boltz. The pocket is defined by residues (chain, 1-indexed SEQRES position): A:47, A:50, A:51, A:52, A:53, A:54, A:55, A:57, A:88, A:112, A:113, A:114, A:115. Output a single SMILES string to a file titled `solution.smi`.

Your workspace at /workspace contains the task files. Start by examining them.

\end{Verbatim}
\end{userbox}
\begin{assistbox}{Assistant \textnormal{(turn 1)}}
\begin{Verbatim}
I'll start by examining the provided files to understand the protein-ligand complex and the reference ligand structure.

\end{Verbatim}
\end{assistbox}
\begin{toolbox}{Tool Call: python \textnormal{(turn 1, 0.44s)}}
\textbf{\footnotesize Reasoning:}
\begin{Verbatim}
First, I need to examine the provided files: complex.pdb and reference.sdf. I'll load them to understand the protein structure, the ligand, and analyze the reference ligand scaffold.

\end{Verbatim}
\textbf{\footnotesize Code:}
\begin{Verbatim}
import os
from rdkit import Chem
from rdkit.Chem import AllChem, Draw, rdFMCS, rdMolDescriptors
from rdkit.Chem.Draw import IPythonConsole
import pandas as pd
import numpy as np

# List files in workspace
print("Files in workspace:")
print(os.listdir('/workspace'))

# Load the reference ligand SDF
ref_path = '/workspace/reference.sdf'
print(f"\nLoading reference SDF from {ref_path}")
suppl = Chem.SDMolSupplier(ref_path)
... [27 more lines elided]

\end{Verbatim}
\end{toolbox}
\begin{outbox}{Tool Output \textnormal{(turn 1)}}
\begin{Verbatim}
Traceback (most recent call last):
  File "/workspace/.tmp_script_7278e82f.py", line 4, in <module>
    from rdkit.Chem.Draw import IPythonConsole
  File "/opt/conda/lib/python3.10/site-packages/rdkit/Chem/Draw/IPythonConsole.py", line 16, in <module>
    import IPython
ModuleNotFoundError: No module named 'IPython'

\end{Verbatim}
\end{outbox}
\begin{assistbox}{Assistant \textnormal{(turn 2)}}
\begin{Verbatim}
Let me fix the import error and continue:

\end{Verbatim}
\end{assistbox}
\begin{toolbox}{Tool Call: python \textnormal{(turn 2, 0.66s)}}
\textbf{\footnotesize Reasoning:}
\begin{Verbatim}
Remove the IPython import and continue analyzing the files. I need to understand the protein structure and extract the ligand from the complex.

\end{Verbatim}
\textbf{\footnotesize Code:}
\begin{Verbatim}
import os
from rdkit import Chem
from rdkit.Chem import AllChem, rdFMCS, rdMolDescriptors, PandasTools
import pandas as pd
import numpy as np

# List files in workspace
print("Files in workspace:")
print(os.listdir('/workspace'))

# Load the reference ligand SDF
ref_path = '/workspace/reference.sdf'
print(f"\nLoading reference SDF from {ref_path}")
suppl = Chem.SDMolSupplier(ref_path)
ref_mols = [mol for mol in suppl if mol is not None]
... [80 more lines elided]

\end{Verbatim}
\end{toolbox}
\begin{outbox}{Tool Output \textnormal{(turn 2)}}
\begin{Verbatim}
Files in workspace:
['reference.sdf', '.tmp_script_637aee5d.py', 'complex.pdb']

Loading reference SDF from /workspace/reference.sdf
Number of molecules in reference.sdf: 1
Reference molecule has 35 atoms
Reference SMILES: N#Cc1cnn2c3cc(nc12)N1CCC[C@@H]1CC(=O)NCCOc1cc(cc2c1NC(=O)CC2)N3
Reference scaffold SMILES: O=C1C[C@H]2CCCN2c2cc(n3nccc3n2)Nc2cc3c(c(c2)OCCN1)NC(=O)CC3
Reference scaffold heavy atom count: 33
Reference Morgan fingerprint generated
[08:41:51] DEPRECATION WARNING: please use MorganGenerator
Traceback (most recent call last):
  File "/workspace/.tmp_script_637aee5d.py", line 33, in <module>
    print(f"Molecular weight: {Chem.Descriptors.ExactMolWt(ref_mol):.2f}")
AttributeError: module 'rdkit.Chem' has no attribute 'Descriptors'. Did you mean: 'rdMolDescriptors'?

\end{Verbatim}
\end{outbox}
\begin{assistbox}{Assistant \textnormal{(turn 3)}}
\begin{Verbatim}
Let me fix the descriptor imports and continue:

\end{Verbatim}
\end{assistbox}
\begin{toolbox}{Tool Call: python \textnormal{(turn 3, 0.60s)}}
\textbf{\footnotesize Reasoning:}
\begin{Verbatim}
Fix the imports and continue analyzing. I need to understand the protein sequence and pocket residues to use Boltz prediction.

\end{Verbatim}
\textbf{\footnotesize Code:}
\begin{Verbatim}
import os
from rdkit import Chem
from rdkit.Chem import AllChem, rdFMCS, rdMolDescriptors, PandasTools
from rdkit.Chem import Descriptors
import pandas as pd
import numpy as np

# Load the reference ligand SDF
ref_path = '/workspace/reference.sdf'
print(f"Loading reference SDF from {ref_path}")
suppl = Chem.SDMolSupplier(ref_path)
ref_mols = [mol for mol in suppl if mol is not None]

if ref_mols:
    ref_mol = ref_mols[0]
... [99 more lines elided]

\end{Verbatim}
\end{toolbox}
\begin{outbox}{Tool Output \textnormal{(turn 3)}}
\begin{Verbatim}
Loading reference SDF from /workspace/reference.sdf
Reference SMILES: N#Cc1cnn2c3cc(nc12)N1CCC[C@@H]1CC(=O)NCCOc1cc(cc2c1NC(=O)CC2)N3
Reference scaffold SMILES: O=C1C[C@H]2CCCN2c2cc(n3nccc3n2)Nc2cc3c(c(c2)OCCN1)NC(=O)CC3
Reference scaffold heavy atom count: 33
Molecular weight: 472.20
LogP: 2.10
Num H-bond donors: 3
Num H-bond acceptors: 8
Num rotatable bonds: 0
TPSA: 136.68


Analyzing protein complex from /workspace/complex.pdb
Chain A sequence length: 128
Chain A sequence (first 100 residues): GPGADSCIQFTRHASDVLLNLNRLRSRDILTDVVIVVSREQFRAHKTVLMACSGLFYSIFTDQLKCNLSVINLDPEINPEGFCILLDFMYTSRLNLREGN
... [22 more lines elided]

\end{Verbatim}
\end{outbox}
\begin{assistbox}{Assistant \textnormal{(turn 4)}}
\begin{Verbatim}
Great! Now I have the reference ligand and protein information. Let me analyze the reference ligand more carefully and understand its scaffold. I need to design a molecule with significantly different scaffold (MCS atom count ratio < 0.65) and Morgan fingerprint Tanimoto similarity < 0.5, but still binds with probability > 0.7.

First, let me check the reference ligand structure more carefully and think about scaffold modifications:

\end{Verbatim}
\end{assistbox}
\begin{toolbox}{Tool Call: python \textnormal{(turn 4, 0.78s)}}
\textbf{\footnotesize Reasoning:}
\begin{Verbatim}
Analyze the reference ligand scaffold and think about potential modifications. I need to understand the key interactions this ligand makes with the pocket residues.

\end{Verbatim}
\textbf{\footnotesize Code:}
\begin{Verbatim}
import os
from rdkit import Chem
from rdkit.Chem import AllChem, rdFMCS, rdMolDescriptors, Draw
from rdkit.Chem import Descriptors
from rdkit.Chem.Scaffolds import MurckoScaffold
import matplotlib.pyplot as plt

# Load reference molecule
ref_path = '/workspace/reference.sdf'
suppl = Chem.SDMolSupplier(ref_path)
ref_mol = [mol for mol in suppl if mol is not None][0]

print("=== REFERENCE LIGAND ANALYSIS ===")
print(f"SMILES: {Chem.MolToSmiles(ref_mol)}")
print(f"Heavy atoms: {ref_mol.GetNumHeavyAtoms()}")
... [61 more lines elided]

\end{Verbatim}
\end{toolbox}
\begin{outbox}{Tool Output \textnormal{(turn 4)}}
\begin{Verbatim}
=== REFERENCE LIGAND ANALYSIS ===
SMILES: N#Cc1cnn2c3cc(nc12)N1CCC[C@@H]1CC(=O)NCCOc1cc(cc2c1NC(=O)CC2)N3
Heavy atoms: 35

Reference Bemis-Murcko scaffold:
  SMILES: O=C1C[C@H]2CCCN2c2cc(n3nccc3n2)Nc2cc3c(c(c2)OCCN1)NC(=O)CC3
  Heavy atoms: 33

Reference ligand saved to /workspace/reference.png
Reference scaffold saved to /workspace/reference_scaffold.png

=== FUNCTIONAL GROUP ANALYSIS ===
Contains nitrile (C#N): False
Contains amide (C(=O)N): True
Contains aromatic N: 3
... [21 more lines elided]

\end{Verbatim}
\end{outbox}
\begin{assistbox}{Assistant \textnormal{(turn 5)}}
\begin{Verbatim}
Good! The simple test molecule already meets the scaffold difference criteria (ratio 0.182 < 0.65, similarity 0.056 < 0.5). But I need to design something that maintains binding interactions. Let me think about the pocket residues and what interactions the reference ligand likely makes.

Looking at the pocket residues: A:47-THR, A:50-MET, A:51-ALA, A:52-CYS, A:53-SER, A:54-GLY, A:55-LEU, A:57-TYR, A:88-PHE, A:112-GLN, A:113-MET, A:114-GLU, A:115-HIS.

These suggest potential interactions:
- Aromatic residues (TYR57, PHE88, HIS115) could have pi-pi stacking or hydrophobic interactions
- Polar residues (THR47, SER53, GLN112, GLU114, HIS115) could form hydrogen bonds
- CYS52 could form disulfide or hydrophobic interactions

The reference ligand has 3 H-bond donors and 8 acceptors with TPSA of 136.68 Angstrom^2, suggesting significant polar interactions.

Let me design a new scaffold that maintains similar pharmacophores but with different core structure. I'll start by designing a molecule, then test it with Boltz.

\end{Verbatim}
\end{assistbox}
\begin{toolbox}{Tool Call: python \textnormal{(turn 5, 0.49s)}}
\textbf{\footnotesize Reasoning:}
\begin{Verbatim}
Design a new molecule with different scaffold but similar pharmacophore properties. I'll create a molecule with aromatic rings, amide groups, and similar H-bond donors/acceptors, but with a completely different ring system.

\end{Verbatim}
\textbf{\footnotesize Code:}
\begin{Verbatim}
from rdkit import Chem
from rdkit.Chem import AllChem, rdFMCS, rdMolDescriptors
from rdkit.Chem import Descriptors
from rdkit.Chem.Scaffolds import MurckoScaffold
import numpy as np

# Load reference
ref_path = '/workspace/reference.sdf'
suppl = Chem.SDMolSupplier(ref_path)
ref_mol = [mol for mol in suppl if mol is not None][0]
ref_scaffold = MurckoScaffold.GetScaffoldForMol(ref_mol)
ref_fp = AllChem.GetMorganFingerprint(ref_mol, 2)

# Design 1: Biphenyl-based scaffold with amide linkages
design1_smiles = "CC(=O)Nc1ccc(cc1)C1=NC(=O)NC(=O)C1c1ccc(NC(=O)C)cc1"
... [68 more lines elided]

\end{Verbatim}
\end{toolbox}
\begin{outbox}{Tool Output \textnormal{(turn 5)}}
\begin{Verbatim}
=== MOLECULE DESIGNS ===

Design1:
  SMILES: CC(=O)Nc1ccc(cc1)C1=NC(=O)NC(=O)C1c1ccc(NC(=O)C)cc1
  MW: 378.1, LogP: 2.43, TPSA: 116.7
  HBD: 3, HBA: 4
  Scaffold atoms: 20
  MCS ratio: 0.333 (GOOD)
  FP similarity: 0.118 (GOOD)
  [ok] Meets scaffold difference criteria!

Design2:
  SMILES: CC(=O)Nc1ccc(cc1)C1=NC2=CC=CC=C2C(=O)N1c1ccc(NC(=O)C)cc1
  MW: 412.2, LogP: 3.97, TPSA: 93.1
  HBD: 2, HBA: 4
... [31 more lines elided]

\end{Verbatim}
\end{outbox}
\begin{outbox}{Elision}\textit{\footnotesize Turns 6-8 (3 agent turns) omitted.}\end{outbox}
\begin{assistbox}{Assistant \textnormal{(turn 9)}}
\begin{Verbatim}
Better at 0.555, but still below 0.7. Let me analyze the reference ligand more carefully and try to design something closer to its properties. The reference has TPSA 136.7, HBD=3, HBA=8. Let me check Design4 which has TPSA 122.3, HBD=4, HBA=6.

\end{Verbatim}
\end{assistbox}
\begin{toolbox}{Tool Call: predict\_boltz \textnormal{(turn 9, 135.56s)}}
\textbf{\footnotesize Arguments:}
\begin{Verbatim}
{
  "reasoning": "Test Design4 which has properties closer to the reference ligand (higher TPSA, more H-bond features). This molecule has a piperazine ring that might provide good geometry for binding.",
  "smiles": "O=C(Nc1ccc(cc1)c1nc2ccc(cc2o1)Nc1ccc(cc1)C(=O)N)C1CCNCC1",
  "protein_sequences": [
    {
      "id": "A",
      "sequence": "GPGADSCIQFTRHASDVLLNLNRLRSRDILTDVVIVVSREQFRAHKTVLMACSGLFYSIFTDQLKCNLSVINLDPEINPEGFCILLDFMYTSRLNLREGNIMAVMATAMYLQMEHVVDTCRKFIKASE"
    }
  ],
  "pocket_residues": [
    [
      "A",
      47
    ],
    [
... [49 more lines elided]

\end{Verbatim}
\end{toolbox}
\begin{outbox}{Tool Output \textnormal{(turn 9)}}
\begin{Verbatim}
{
    "call_number": 3,
    "budget_remaining": 5,
    "cif_dir": "boltz_results/call_3/boltz_results_boltz_call_3/predictions/boltz_call_3",
    "affinity_data": {
        "affinity_pred_value": -0.5056207180023193,
        "affinity_probability_binary": 0.7028120756149292,
        "affinity_pred_value1": -0.7198585867881775,
        "affinity_probability_binary1": 0.8170371055603027,
        "affinity_pred_value2": -0.2913827896118164,
        "affinity_probability_binary2": 0.5885871052742004
    },
    "binding_probability": 0.7028120756149292,
    "affinity_pred_value": -0.5056207180023193
}

\end{Verbatim}
\end{outbox}
\begin{assistbox}{Assistant \textnormal{(turn 10)}}
\begin{Verbatim}
Excellent! Design4 has binding probability of 0.703, which meets the >0.7 requirement! Now I need to verify it still meets the scaffold difference criteria. Let me double-check and also test its ADMET properties.

\end{Verbatim}
\end{assistbox}
\begin{toolbox}{Tool Call: python \textnormal{(turn 10, 0.39s)}}
\textbf{\footnotesize Reasoning:}
\begin{Verbatim}
Double-check that Design4 meets all criteria: scaffold difference (MCS ratio < 0.65, fingerprint similarity < 0.5) and has binding probability > 0.7. Also check ADMET properties.

\end{Verbatim}
\textbf{\footnotesize Code:}
\begin{Verbatim}
from rdkit import Chem
from rdkit.Chem import AllChem, rdFMCS, rdMolDescriptors
from rdkit.Chem import Descriptors
from rdkit.Chem.Scaffolds import MurckoScaffold

# Load reference
ref_path = '/workspace/reference.sdf'
suppl = Chem.SDMolSupplier(ref_path)
ref_mol = [mol for mol in suppl if mol is not None][0]
ref_scaffold = MurckoScaffold.GetScaffoldForMol(ref_mol)
ref_fp = AllChem.GetMorganFingerprint(ref_mol, 2)

# Design4 molecule
design4_smiles = "O=C(Nc1ccc(cc1)c1nc2ccc(cc2o1)Nc1ccc(cc1)C(=O)N)C1CCNCC1"
design4_mol = Chem.MolFromSmiles(design4_smiles)
... [34 more lines elided]

\end{Verbatim}
\end{toolbox}
\begin{outbox}{Tool Output \textnormal{(turn 10)}}
\begin{Verbatim}
=== FINAL VERIFICATION FOR Design4 ===
SMILES: O=C(Nc1ccc(cc1)c1nc2ccc(cc2o1)Nc1ccc(cc1)C(=O)N)C1CCNCC1

Scaffold difference criteria:
  Reference scaffold atoms: 33
  Design4 scaffold atoms: 31
  MCS atoms: 15
  MCS ratio: 0.455 [ok]
  FP similarity: 0.149 [ok]

Binding probability from Boltz: 0.703 [ok]

Molecular properties:
  MW: 455.2
  LogP: 4.28
... [9 more lines elided]

\end{Verbatim}
\end{outbox}
\begin{assistbox}{Assistant \textnormal{(turn 11)}}
\begin{Verbatim}
Perfect! Design4 meets all criteria. Let me also check its ADMET properties to ensure it's a reasonable drug-like molecule, though this isn't explicitly required by the task.

\end{Verbatim}
\end{assistbox}
\begin{toolbox}{Tool Call: predict\_admet \textnormal{(turn 11, 7.90s)}}
\textbf{\footnotesize Arguments:}
\begin{Verbatim}
{
  "reasoning": "Check ADMET properties for Design4 to ensure it's a reasonable drug-like molecule, even though not explicitly required. This will help confirm it's a viable candidate.",
  "smiles": "O=C(Nc1ccc(cc1)c1nc2ccc(cc2o1)Nc1ccc(cc1)C(=O)N)C1CCNCC1"
}

\end{Verbatim}
\end{toolbox}
\begin{outbox}{Tool Output \textnormal{(turn 11)}}
\begin{Verbatim}
{
    "herg": 0.7713615775108338,
    "bbb": 0.5673878788948059,
    "solubility": -3.329554869252309,
    "cyp3a4": 0.13602183759212494,
    "clearance": -19.668116752268368,
    "caco2": -5.866392531878797,
    "ppb": 86.2324968967286,
    "ames": 0.31036943197250366
}

\end{Verbatim}
\end{outbox}
\begin{assistbox}{Assistant \textnormal{(turn 12)}}
\begin{Verbatim}
The ADMET properties look reasonable:
- hERG inhibition risk: 0.77 (moderate risk, common for many drugs)
- BBB penetration: 0.57 (moderate, could cross BBB)
- Solubility: -3.33 logS (moderately soluble)
- CYP3A4 inhibition: 0.14 (low risk)
- Clearance: -19.7 (moderate clearance)
- Caco2 permeability: -5.87 (moderate)
- Plasma protein binding: 86% (high, but acceptable)
- AMES mutagenicity: 0.31 (low risk)

The molecule appears drug-like. Now let me create the solution file with the SMILES string.

\end{Verbatim}
\end{assistbox}
\begin{toolbox}{Tool Call: python \textnormal{(turn 12, 0.28s)}}
\textbf{\footnotesize Reasoning:}
\begin{Verbatim}
Create the solution file with the SMILES string of Design4 molecule that meets all criteria.

\end{Verbatim}
\textbf{\footnotesize Code:}
\begin{Verbatim}
# Write the solution SMILES to file
solution_smiles = "O=C(Nc1ccc(cc1)c1nc2ccc(cc2o1)Nc1ccc(cc1)C(=O)N)C1CCNCC1"

with open('/workspace/solution.smi', 'w') as f:
    f.write(solution_smiles)

print(f"Solution SMILES written to /workspace/solution.smi")
print(f"SMILES: {solution_smiles}")

# Verify file was created
import os
print(f"\nFile exists: {os.path.exists('/workspace/solution.smi')}")
print(f"File size: {os.path.getsize('/workspace/solution.smi')} bytes")

# Read back to verify
... [3 more lines elided]

\end{Verbatim}
\end{toolbox}
\begin{outbox}{Tool Output \textnormal{(turn 12)}}
\begin{Verbatim}
Solution SMILES written to /workspace/solution.smi
SMILES: O=C(Nc1ccc(cc1)c1nc2ccc(cc2o1)Nc1ccc(cc1)C(=O)N)C1CCNCC1

File exists: True
File size: 56 bytes
File content: 'O=C(Nc1ccc(cc1)c1nc2ccc(cc2o1)Nc1ccc(cc1)C(=O)N)C1CCNCC1'

\end{Verbatim}
\end{outbox}
\begin{assistbox}{Assistant \textnormal{(turn 13)}}
\begin{Verbatim}
Perfect! Now let me submit the solution.

\end{Verbatim}
\end{assistbox}
\begin{toolbox}{Tool Call: submit \textnormal{(turn 13, 0.09s)}}
\textbf{\footnotesize file\_path:} \texttt{solution.smi}
\end{toolbox}
\begin{outbox}{Tool Output \textnormal{(turn 13)}}
\begin{Verbatim}
Submission accepted: solution.smi

\end{Verbatim}
\end{outbox}
\begin{submitbox}{Submit \textnormal{(turn 13)}}
Submitted file: \texttt{solution.smi}.
\end{submitbox}
\par\medskip\noindent\textbf{Submitted artifact (\texttt{solution.smi}).}\par\smallskip
\begin{artifactbox}{solution.smi}
\begin{Verbatim}
O=C(Nc1ccc(cc1)c1nc2ccc(cc2o1)Nc1ccc(cc1)C(=O)N)C1CCNCC1

\end{Verbatim}
\end{artifactbox}

\clearpage
\subsection{Lead Optimization}
\subsubsection{Passing run: MiniMax M2.7}
\label{traj:tt4:smdd-004-CYP19A1-8:passing}
\noindent\textbf{Task ID:} \texttt{smdd\_004\_CYP19A1\_8}\quad\textbf{Model:} MiniMax M2.7\quad\textbf{Run:} \texttt{minimax\_m27}
\par\smallskip
\par\medskip\noindent\textbf{Evaluation result: \textcolor{passcolor}{\textbf{PASSED}}.}\par\medskip
\begin{tabular}{lllp{0.30\linewidth}}
\toprule
\textbf{Metric} & \textbf{Status} & \textbf{Result} & \textbf{Condition} \\
\midrule
Validity & \textcolor{passcolor}{\textbf{passed}} & True & \{"eq": true\} \\
Hard Constraints & \textcolor{passcolor}{\textbf{passed}} & True & \{"eq": true\} \\
Hold-Constant Properties & \textcolor{passcolor}{\textbf{passed}} & True & \{"eq": true\} \\
Optimization Objectives & \textcolor{passcolor}{\textbf{passed}} & True & \{"eq": true\} \\
Boltz Binding \& Affinity & \textcolor{passcolor}{\textbf{passed}} & True & \{"eq": true\} \\
\bottomrule
\end{tabular}
\medskip
\par\medskip\noindent\textbf{Trajectory.}\par\smallskip
\noindent\textit{\footnotesize Showing the first 5 and last 5 of 19 agent turns; 9 intermediate turns are elided for space.}\par\smallskip
\begin{sysbox}
\begin{Verbatim}
You are a computational medicinal chemist assistant. You are given a drug discovery task and must use your tools to solve it.

Available tools:
- python(code): Execute Python code. Available packages: rdkit, numpy, pandas, scipy, scikit-learn, biopython, biotite, openbabel, plip. To compute synthetic accessibility (SA) scores: `from rdkit.Chem import RDConfig; import sys, os; sys.path.append(os.path.join(RDConfig.RDContribDir, 'SA_Score')); import sascorer; sascorer.calculateScore(mol)`
- predict_boltz(smiles, protein_sequences, pocket_residues, co_ligand_smiles, num_samples): Run Boltz structure prediction. Returns binding probability and affinity. Budget: 8 calls.
- predict_admet(smiles): Predict ADMET properties. Returns hERG, BBB, solubility, CYP3A4, clearance, Caco2, PPB, AMES. Budget: 15 calls.
- submit(file_path): Submit your solution file.

You have no internet access. Work only with the files provided in /workspace. Outline your thought process before making tool calls.
When ready, call submit() with the path to your solution file (relative to /workspace).

\end{Verbatim}
\end{sysbox}
\begin{userbox}{User \textnormal{(turn 0)}}
\begin{Verbatim}
## Task

You are given a co-folded protein-ligand complex (complex.pdb) and a reference ligand (reference.sdf). Modify the reference ligand to optimize the following properties:
- cyp3a4: decrease by at least 0.1 from baseline 0.4468
- bbb: decrease by at least 0.1 from baseline 0.9828
While holding these properties approximately constant:
- binding_affinity: keep within 0.3 of baseline (-1.3447) (increase tolerance)
- herg: keep within 0.1 of baseline (0.2067) (increase tolerance)
- solubility: keep within 0.5 of baseline (-4.7943) (decrease tolerance)
The output molecule must satisfy ALL of the following hard constraints:
- SMILES validity: must be parseable and sanitizable by RDKit (no radicals, no disconnected fragments, no unreasonable valences)
- Molecular weight: < 600 Da
- LogP: between -1 and 5
- TPSA: < 140 A^2
- H-bond donors: <= 5
- H-bond acceptors: <= 10
- Rotatable bonds: <= 10
- Net formal charge: between -2 and +2
- Synthetic accessibility: SA score < 4.5
- PAINS filter: must pass (no pan-assay interference substructures)
- Structural alerts: must pass Brenk/NIH reactive group filters
- Tanimoto similarity to reference: >= 0.7 (Morgan fingerprint, radius=2, 2048 bits)
- Boltz binding probability: affinity_probability_binary > 0.7
The pocket is defined by residues (chain, 1-indexed SEQRES position): A:81, A:99, A:100, A:187, A:190, A:268, A:271, A:272, A:273, A:275, A:276, A:336, A:338, A:339, A:340, A:341, A:443, A:444
Output a single SMILES string to solution.smi.

Your workspace at /workspace contains the task files. Start by examining them.

\end{Verbatim}
\end{userbox}
\begin{assistbox}{Assistant \textnormal{(turn 1)}}
\textit{\footnotesize (no assistant text emitted; tool call only)}
\end{assistbox}
\begin{toolbox}{Tool Call: python \textnormal{(turn 1, 0.45s)}}
\textbf{\footnotesize Reasoning:}
\begin{Verbatim}
First, I need to examine the workspace files to understand the protein-ligand complex and reference ligand I'm working with.

\end{Verbatim}
\textbf{\footnotesize Code:}
\begin{Verbatim}
import os

# List workspace files
print("Workspace files:")
for f in os.listdir('/workspace'):
    print(f"  {f}")

# Read the reference ligand
from rdkit import Chem

print("\n--- Reference ligand (SDF) ---")
suppl = Chem.SDMolSupplier('/workspace/reference.sdf')
ref_mols = list(suppl)
print(f"Number of molecules in SDF: {len(ref_mols)}")
for i, mol in enumerate(ref_mols):
... [13 more lines elided]

\end{Verbatim}
\end{toolbox}
\begin{outbox}{Tool Output \textnormal{(turn 1)}}
\begin{Verbatim}
Workspace files:
  .tmp_script_11a51954.py
  reference.sdf
  complex.pdb

--- Reference ligand (SDF) ---
Number of molecules in SDF: 1
  Molecule 0: reference_ligand
  SMILES: CC[C@@H]1C[C@@H]2[C@H](CC[C@]3(C)C(=O)CC[C@@H]23)[C@@]2(C)C=CC(=O)C=C12
Traceback (most recent call last):
  File "/workspace/.tmp_script_11a51954.py", line 19, in <module>
    print(f"  MW: {AllChem.CalcExactMolWt(mol):.4f}")
NameError: name 'AllChem' is not defined

\end{Verbatim}
\end{outbox}
\begin{assistbox}{Assistant \textnormal{(turn 2)}}
\textit{\footnotesize (no assistant text emitted; tool call only)}
\end{assistbox}
\begin{toolbox}{Tool Call: python \textnormal{(turn 2, 0.75s)}}
\textbf{\footnotesize Reasoning:}
\begin{Verbatim}
I need to fix the import for AllChem and properly examine the reference ligand.

\end{Verbatim}
\textbf{\footnotesize Code:}
\begin{Verbatim}
from rdkit import Chem
from rdkit.Chem import AllChem, Descriptors, Lipinski, rdMolDescriptors
from rdkit import DataStructs

# Read the reference ligand
print("--- Reference ligand (SDF) ---")
suppl = Chem.SDMolSupplier('/workspace/reference.sdf')
ref_mol = list(suppl)[0]
ref_smiles = Chem.MolToSmiles(ref_mol)
print(f"SMILES: {ref_smiles}")

# Calculate properties
print(f"\nMolecular properties:")
print(f"  MW: {AllChem.CalcExactMolWt(ref_mol):.4f}")
print(f"  LogP (MLP): {Descriptors.MolLogP(ref_mol):.4f}")
... [55 more lines elided]

\end{Verbatim}
\end{toolbox}
\begin{outbox}{Tool Output \textnormal{(turn 2)}}
\begin{Verbatim}
--- Reference ligand (SDF) ---
SMILES: CC[C@@H]1C[C@@H]2[C@H](CC[C@]3(C)C(=O)CC[C@@H]23)[C@@]2(C)C=CC(=O)C=C12

Molecular properties:
  MW: 312.2089
  LogP (MLP): 4.4995
  TPSA: 34.1400
  HBD: 0
  HBA: 2
  Rotatable bonds: 1
  Net charge: 0
  Formal charge: 0
  SA score: 4.3399
Traceback (most recent call last):
  File "/workspace/.tmp_script_c1688f62.py", line 33, in <module>
... [2 more lines elided]

\end{Verbatim}
\end{outbox}
\begin{assistbox}{Assistant \textnormal{(turn 3)}}
\textit{\footnotesize (no assistant text emitted; tool call only)}
\end{assistbox}
\begin{toolbox}{Tool Call: python \textnormal{(turn 3, 0.44s)}}
\textbf{\footnotesize Reasoning:}
\begin{Verbatim}
I have the reference SMILES. Now let me properly check PAINS filter and get the ADMET properties, and extract the protein sequence.

\end{Verbatim}
\textbf{\footnotesize Code:}
\begin{Verbatim}
from rdkit import Chem
from rdkit.Chem import AllChem, Descriptors, Lipinski
from rdkit.Chem.FilterCatalog import FilterCatalogParams, FilterCatalog

# Read reference ligand
suppl = Chem.SDMolSupplier('/workspace/reference.sdf')
ref_mol = list(suppl)[0]
ref_smiles = Chem.MolToSmiles(ref_mol)
print(f"Reference SMILES: {ref_smiles}")

# Check PAINS - try different approach
try:
    params = FilterCatalogParams()
    params.AddCatalog(FilterCatalogParams.PAINS)
    catalog = FilterCatalog(params)
... [35 more lines elided]

\end{Verbatim}
\end{toolbox}
\begin{outbox}{Tool Output \textnormal{(turn 3)}}
\begin{Verbatim}
Reference SMILES: CC[C@@H]1C[C@@H]2[C@H](CC[C@]3(C)C(=O)CC[C@@H]23)[C@@]2(C)C=CC(=O)C=C12
PAINS check error: type object 'FilterCatalogParams' has no attribute 'PAINS'

SEQRES chain lengths:

\end{Verbatim}
\end{outbox}
\begin{assistbox}{Assistant \textnormal{(turn 4)}}
\textit{\footnotesize (no assistant text emitted; tool call only)}
\end{assistbox}
\begin{toolbox}{Tool Call: python \textnormal{(turn 4, 0.43s)}}
\textbf{\footnotesize Reasoning:}
\begin{Verbatim}
Let me look more closely at the PDB file structure and extract the protein sequence properly.

\end{Verbatim}
\textbf{\footnotesize Code:}
\begin{Verbatim}
from rdkit import Chem
from rdkit.Chem import AllChem, Descriptors, Lipinski

# Read and parse complex PDB for protein sequence
with open('/workspace/complex.pdb', 'r') as f:
    pdb_content = f.read()

print("First 200 lines of PDB:")
for i, line in enumerate(pdb_content.split('\n')[:200]):
    print(f"{i}: {line.rstrip()}")

\end{Verbatim}
\end{toolbox}
\begin{outbox}{Tool Output \textnormal{(turn 4)}}
\begin{Verbatim}
First 200 lines of PDB:
0: ATOM      1  N   MET A   1      10.329  51.314  11.107  1.00  0.00           N
1: ATOM      2  CA  MET A   1      10.138  49.975  10.539  1.00  0.00           C
2: ATOM      3  C   MET A   1      10.169  50.023   9.013  1.00  0.00           C
3: ATOM      4  O   MET A   1       9.598  50.924   8.406  1.00  0.00           O
4: ATOM      5  CB  MET A   1       8.816  49.353  11.010  1.00  0.00           C
5: ATOM      6  CG  MET A   1       8.685  47.857  10.741  1.00  0.00           C
6: ATOM      7  SD  MET A   1       7.043  47.336  10.234  1.00  0.00           S
7: ATOM      8  CE  MET A   1       6.102  47.552  11.740  1.00  0.00           C
8: ATOM      9  N   ALA A   2      10.848  49.055   8.417  1.00  0.00           N
9: ATOM     10  CA  ALA A   2      10.945  48.961   6.965  1.00  0.00           C
10: ATOM     11  C   ALA A   2       9.642  48.435   6.361  1.00  0.00           C
11: ATOM     12  O   ALA A   2       8.875  47.736   7.023  1.00  0.00   

... [middle truncated] ...
... [13 more lines elided]

\end{Verbatim}
\end{outbox}
\begin{assistbox}{Assistant \textnormal{(turn 5)}}
\textit{\footnotesize (no assistant text emitted; tool call only)}
\end{assistbox}
\begin{toolbox}{Tool Call: python \textnormal{(turn 5, 0.39s)}}
\textbf{\footnotesize Reasoning:}
\begin{Verbatim}
Let me extract the ligand from the complex PDB file and get the full protein sequence.

\end{Verbatim}
\textbf{\footnotesize Code:}
\begin{Verbatim}
from rdkit import Chem
from rdkit.Chem import AllChem, Descriptors, Lipinski

# Read the complex PDB
with open('/workspace/complex.pdb', 'r') as f:
    pdb_lines = f.readlines()

# Find all chains and their sequences from ATOM lines
chain_res = {}
for line in pdb_lines:
    if line.startswith('ATOM') or line.startswith('HETATM'):
        chain = line[21]
        resnum = int(line[22:26].strip())
        resname = line[17:20].strip()
        if chain not in chain_res:
... [26 more lines elided]

\end{Verbatim}
\end{toolbox}
\begin{outbox}{Tool Output \textnormal{(turn 5)}}
\begin{Verbatim}
Chains found: ['A', 'L']
Chain A: 473 residues
  Sequence: MAKKTSSKGRSSIPGPGYCMGIGPLISHGRFLWMGIGSACNYYNRVYGEFMRVWISGEETLIISKSSSMFHIMKHNHYSSRFGSKLGLQCIGMHEKGIIFNNNPELWKTTRPFFMKALSGPGLVRMVTVCAESLKTHLDRLEEVTNESGYVDVLTLLRRVMLDTSNTLFLRIPLDESAIVVKIQGYFDAWQALLIKPDIFFKISWLYKKYEKSVKDLKDAIEVLIAEKRRRISTEEKLEECMDFATELILAEKRGDLTRENVNQCILEMLIAAPDTMSVSLFFMLFLIAKHPNVEEAIIKEIQTVIGERDIKIDDIQKLKVMENFIYESMRYQPVVDLVMRKALEDDVIDGYPVKKGTNIILNIGRMHRLEFFPKPNEFTLENFAKNVPYRYFQPFGFGPRGCAGKYIAMVMMKAILVTLLRRFHVKTLQGQCVESIQKIHDLSLHPDETKNMLEMIFTPRNSDRCLEHHHHH
  First 100: MAKKTSSKGRSSIPGPGYCMGIGPLISHGRFLWMGIGSACNYYNRVYGEFMRVWISGEETLIISKSSSMFHIMKHNHYSSRFGSKLGLQCIGMHEKGIIF
Chain L: 1 residues
  Sequence: X
  First 100: X

Full chain A length: 473

Chain A residue numbers: [1, 2, 3, 4, 5, 6, 7, 8, 9, 10, 11, 12, 13, 14, 15, 16, 17, 18, 19, 20, 21, 22, 23, 24, 25, 26, 27, 28, 29, 30, 31, 32, 33, 34, 35, 36, 37, 38, 39, 40, 41, 42, 43, 44, 45, 46, 47, 48, 49, 50]...
  Last residues: [424, 425, 426, 427, 428, 429, 430, 431, 432, 433, 434, 435, 436, 437, 438, 439, 440, 441, 442, 443, 444, 445, 446, 447, 448, 449, 450, 451, 452, 453, 454, 455, 456, 457, 458, 459, 460, 461, 462, 463, 464, 465, 466, 467, 468, 469, 470, 471, 472, 473]

\end{Verbatim}
\end{outbox}
\begin{outbox}{Elision}\textit{\footnotesize Turns 6-14 (9 agent turns) omitted.}\end{outbox}
\begin{assistbox}{Assistant \textnormal{(turn 15)}}
\textit{\footnotesize (no assistant text emitted; tool call only)}
\end{assistbox}
\begin{toolbox}{Tool Call: predict\_boltz \textnormal{(turn 15, 266.93s)}}
\textbf{\footnotesize Arguments:}
\begin{Verbatim}
{
  "reasoning": "Testing the binding probability of candidate Mod 3 (shortened side chain). Need to provide the protein sequence and pocket residues.",
  "smiles": "C[C@@H]1C[C@@H]2[C@H](CC[C@]3(C)C(=O)CC[C@@H]23)[C@@]2(C)C=CC(=O)C=C12",
  "protein_sequences": [
    {
      "id": "A",
      "sequence": "MAKKTSSKGRSSIPGPGYCMGIGPLISHGRFLWMGIGSACNYYNRVYGEFMRVWISGEETLIISKSSSMFHIMKHNHYSSRFGSKLGLQCIGMHEKGIIFNNNPELWKTTRPFFMKALSGPGLVRMVTVCAESLKTHLDRLEEVTNESGYVDVLTLLRRVMLDTSNTLFLRIPLDESAIVVKIQGYFDAWQALLIKPDIFFKISWLYKKYEKSVKDLKDAIEVLIAEKRRRISTEEKLEECMDFATELILAEKRGDLTRENVNQCILEMLIAAPDTMSVSLFFMLFLIAKHPNVEEAIIKEIQTVIGERDIKIDDIQKLKVMENFIYESMRYQPVVDLVMRKALEDDVIDGYPVKKGTNIILNIGRMHRLEFFPKPNEFTLENFAKNVPYRYFQPFGFGPRGCAGKYIAMVMMKAILVTLLRRFHVKTLQGQCVESIQKIHDLSLHPDETKNMLEMIFTPRNSDRCLEHHHHH"
    }
  ],
  "pocket_residues": [
    [
      "A",
      81
    ],
    [
... [69 more lines elided]

\end{Verbatim}
\end{toolbox}
\begin{outbox}{Tool Output \textnormal{(turn 15)}}
\begin{Verbatim}
{
    "call_number": 1,
    "budget_remaining": 7,
    "cif_dir": "boltz_results/call_1/boltz_results_boltz_call_1/predictions/boltz_call_1",
    "affinity_data": {
        "affinity_pred_value": -1.272486686706543,
        "affinity_probability_binary": 0.889390766620636,
        "affinity_pred_value1": -1.1789921522140503,
        "affinity_probability_binary1": 0.9195749163627625,
        "affinity_pred_value2": -1.365981101989746,
        "affinity_probability_binary2": 0.8592066168785095
    },
    "binding_probability": 0.889390766620636,
    "affinity_pred_value": -1.272486686706543
}

\end{Verbatim}
\end{outbox}
\begin{assistbox}{Assistant \textnormal{(turn 16)}}
\textit{\footnotesize (no assistant text emitted; tool call only)}
\end{assistbox}
\begin{toolbox}{Tool Call: python \textnormal{(turn 16, 0.41s)}}
\textbf{\footnotesize Reasoning:}
\begin{Verbatim}
The binding probability is 0.889 > 0.7, which meets the constraint. Now let me check PAINS filter and structural alerts before writing the solution.

\end{Verbatim}
\textbf{\footnotesize Code:}
\begin{Verbatim}
from rdkit import Chem
from rdkit.Chem import AllChem, Descriptors, Lipinski
from rdkit.Chem import FilterCatalog

smi = "C[C@@H]1C[C@@H]2[C@H](CC[C@]3(C)C(=O)CC[C@@H]23)[C@@]2(C)C=CC(=O)C=C12"
mol = Chem.MolFromSmiles(smi)

# Check PAINS filter
try:
    # Try to use FilterCatalog
    from rdkit.Chem import rdfilterlibs
    
    # PAINS filter using predefined patterns
    pains = rdfilterlibs.PAINS()
    matches = pains.RemoveFilter(mol)
... [49 more lines elided]

\end{Verbatim}
\end{toolbox}
\begin{outbox}{Tool Output \textnormal{(turn 16)}}
\begin{Verbatim}
PAINS check: cannot import name 'rdfilterlibs' from 'rdkit.Chem' (/opt/conda/lib/python3.10/site-packages/rdkit/Chem/__init__.py)
Brenk check: cannot import name 'rdfilterlibs' from 'rdkit.Chem' (/opt/conda/lib/python3.10/site-packages/rdkit/Chem/__init__.py)

Running basic structural alerts check...

=== FINAL SUMMARY ===
SMILES: C[C@@H]1C[C@@H]2[C@H](CC[C@]3(C)C(=O)CC[C@@H]23)[C@@]2(C)C=CC(=O)C=C12
All ADMET targets met:
  cyp3a4: 0.225 < 0.3468 [ok]
  bbb: 0.827 < 0.8828 [ok]
  herg: 0.116 (within 0.1 of baseline) [ok]
  solubility: -4.437 (within 0.5 of baseline) [ok]
All hard constraints met:
  MW < 600: 298.2 [ok]
  LogP between -1 and 5: 4.11 [ok]
... [8 more lines elided]

\end{Verbatim}
\end{outbox}
\begin{assistbox}{Assistant \textnormal{(turn 17)}}
\textit{\footnotesize (no assistant text emitted; tool call only)}
\end{assistbox}
\begin{toolbox}{Tool Call: python \textnormal{(turn 17, 0.45s)}}
\textbf{\footnotesize Reasoning:}
\begin{Verbatim}
Let me try checking PAINS filter using an alternative approach and make sure the solution is valid.

\end{Verbatim}
\textbf{\footnotesize Code:}
\begin{Verbatim}
from rdkit import Chem
from rdkit.Chem import AllChem, Descriptors, Lipinski
from rdkit.Chem import FilterCatalog

smi = "C[C@@H]1C[C@@H]2[C@H](CC[C@]3(C)C(=O)CC[C@@H]23)[C@@]2(C)C=CC(=O)C=C12"
mol = Chem.MolFromSmiles(smi)

# Try different approach for PAINS
try:
    from rdkit.Chem import FilterCatalog
    params = FilterCatalog.FilterCatalogParams()
    # Try adding catalogs
    params.AddCatalog(FilterCatalog.FilterCatalogs.PAINS)
    print("PAINS catalog available")
except Exception as e:
... [45 more lines elided]

\end{Verbatim}
\end{toolbox}
\begin{outbox}{Tool Output \textnormal{(turn 17)}}
\begin{Verbatim}
PAINS catalog error: module 'rdkit.Chem.FilterCatalog' has no attribute 'FilterCatalogs'

Checking for structural alerts...

No problematic structural alerts found.
Writing solution to file...
Solution written: C[C@@H]1C[C@@H]2[C@H](CC[C@]3(C)C(=O)CC[C@@H]23)[C@@]2(C)C=CC(=O)C=C12
File content: C[C@@H]1C[C@@H]2[C@H](CC[C@]3(C)C(=O)CC[C@@H]23)[C@@]2(C)C=CC(=O)C=C12

\end{Verbatim}
\end{outbox}
\begin{assistbox}{Assistant \textnormal{(turn 18)}}
\textit{\footnotesize (no assistant text emitted; tool call only)}
\end{assistbox}
\begin{toolbox}{Tool Call: python \textnormal{(turn 18, 0.78s)}}
\textbf{\footnotesize Reasoning:}
\begin{Verbatim}
Let me do a final comprehensive check before submitting.

\end{Verbatim}
\textbf{\footnotesize Code:}
\begin{Verbatim}
from rdkit import Chem
from rdkit.Chem import AllChem, Descriptors, Lipinski, rdMolDescriptors
from rdkit import DataStructs
import sys, os
sys.path.append(os.path.join(Chem.RDConfig.RDContribDir, 'SA_Score'))
import sascorer

# Read and verify solution
with open('/workspace/solution.smi', 'r') as f:
    solution_smiles = f.read().strip()

print("=== FINAL VERIFICATION ===")
print(f"SMILES: {solution_smiles}")

# Parse and verify
... [46 more lines elided]

\end{Verbatim}
\end{toolbox}
\begin{outbox}{Tool Output \textnormal{(turn 18)}}
\begin{Verbatim}
=== FINAL VERIFICATION ===
SMILES: C[C@@H]1C[C@@H]2[C@H](CC[C@]3(C)C(=O)CC[C@@H]23)[C@@]2(C)C=CC(=O)C=C12
Valid: [ok]

--- Property Summary ---
MW: 298.19 (< 600) [ok]
LogP: 4.11 (-1 to 5) [ok]
TPSA: 34.14 (< 140) [ok]
HBD: 0 (<= 5) [ok]
HBA: 2 (<= 10) [ok]
RotB: 0 (<= 10) [ok]
Charge: 0 (-2 to 2) [ok]
SA: 4.22 (< 4.5) [ok]
Tanimoto: 0.729 (>= 0.7) [ok]

... [13 more lines elided]

\end{Verbatim}
\end{outbox}
\begin{assistbox}{Assistant \textnormal{(turn 19)}}
\textit{\footnotesize (no assistant text emitted; tool call only)}
\end{assistbox}
\begin{toolbox}{Tool Call: submit \textnormal{(turn 19, 0.09s)}}
\textbf{\footnotesize file\_path:} \texttt{solution.smi}
\end{toolbox}
\begin{outbox}{Tool Output \textnormal{(turn 19)}}
\begin{Verbatim}
Submission accepted: solution.smi

\end{Verbatim}
\end{outbox}
\begin{submitbox}{Submit \textnormal{(turn 19)}}
Submitted file: \texttt{solution.smi}.
\end{submitbox}
\par\medskip\noindent\textbf{Submitted artifact (\texttt{solution.smi}).}\par\smallskip
\begin{artifactbox}{solution.smi}
\begin{Verbatim}
C[C@@H]1C[C@@H]2[C@H](CC[C@]3(C)C(=O)CC[C@@H]23)[C@@]2(C)C=CC(=O)C=C12

\end{Verbatim}
\end{artifactbox}

\clearpage
\subsection{Fragment Assembly}
\subsubsection{Passing run: Gemini 3.1 Pro (high reasoning)}
\label{traj:tt5:smdd-005-P04585-0:passing}
\noindent\textbf{Task ID:} \texttt{smdd\_005\_P04585\_0}\quad\textbf{Model:} Gemini 3.1 Pro (high reasoning)\quad\textbf{Run:} \texttt{gemini\_31\_pro\_high\_lite}
\par\smallskip
\par\medskip\noindent\textbf{Evaluation result: \textcolor{passcolor}{\textbf{PASSED}}.}\par\medskip
\begin{tabular}{lllp{0.30\linewidth}}
\toprule
\textbf{Metric} & \textbf{Status} & \textbf{Result} & \textbf{Condition} \\
\midrule
Validity & \textcolor{passcolor}{\textbf{passed}} & True & \{"eq": true\} \\
Fragment Incorporation & \textcolor{passcolor}{\textbf{passed}} & True & \{"eq": true\} \\
Fragment Linking & \textcolor{passcolor}{\textbf{passed}} & True & \{"eq": true\} \\
Hard Constraints & \textcolor{passcolor}{\textbf{passed}} & True & \{"eq": true\} \\
Boltz Binding \& Pose & \textcolor{passcolor}{\textbf{passed}} & True & \{"eq": true\} \\
\bottomrule
\end{tabular}
\medskip
\par\medskip\noindent\textbf{Trajectory.}\par\smallskip
\noindent\textit{\footnotesize Showing the first 5 and last 5 of 35 agent turns; 25 intermediate turns are elided for space.}\par\smallskip
\begin{sysbox}
\begin{Verbatim}
You are a computational medicinal chemist assistant. You are given a drug discovery task and must use your tools to solve it.

Available tools:
- python(code): Execute Python code. Available packages: rdkit, numpy, pandas, scipy, scikit-learn, biopython, biotite, openbabel, plip. To compute synthetic accessibility (SA) scores: `from rdkit.Chem import RDConfig; import sys, os; sys.path.append(os.path.join(RDConfig.RDContribDir, 'SA_Score')); import sascorer; sascorer.calculateScore(mol)`
- predict_boltz(smiles, protein_sequences, pocket_residues, co_ligand_smiles, num_samples): Run Boltz structure prediction. Returns binding probability and affinity. Budget: 8 calls.
- predict_admet(smiles): Predict ADMET properties. Returns hERG, BBB, solubility, CYP3A4, clearance, Caco2, PPB, AMES. Budget: 15 calls.
- submit(file_path): Submit your solution file.

You have no internet access. Work only with the files provided in /workspace. Outline your thought process before making tool calls.
When ready, call submit() with the path to your solution file (relative to /workspace).

\end{Verbatim}
\end{sysbox}
\begin{userbox}{User \textnormal{(turn 0)}}
\begin{Verbatim}
## Task

You are given 1 molecular fragments with 3D poses inside a protein binding pocket. Design a single drug-like molecule that covalently links all fragments and binds to the target. The incorporated fragments must maintain their 3D poses in the binding pocket when docked.
Fragments:
  - Fragment 0: see fragment_0.sdf
Protein pocket: see pocket.pdb
The pocket is defined by residues (chain, 1-indexed SEQRES position): A:8, A:23, A:25, A:26, A:27, A:28, A:29, A:30, A:31, A:32, A:45, A:47, A:48, A:49, A:50, A:52, A:76, A:80, A:81, A:82, A:84, B:23, B:25, B:26, B:27, B:28, B:29, B:30, B:31, B:32, B:45, B:46, B:47, B:48, B:49, B:50, B:51, B:52, B:53, B:76, B:80, B:81, B:82, B:84
The output molecule must satisfy ALL of the following hard constraints:
- SMILES validity: must be parseable and sanitizable by RDKit (no radicals, no disconnected fragments, no unreasonable valences)
- Molecular weight: < 600 Da
- LogP: between -1 and 5
- TPSA: < 140 A^2
- H-bond donors: <= 5
- H-bond acceptors: <= 10
- Rotatable bonds: <= 10
- Net formal charge: between -2 and +2
- Synthetic accessibility: SA score < 4.5
- PAINS filter: must pass (no pan-assay interference substructures)
- Structural alerts: must pass Brenk/NIH reactive group filters
- Fragment incorporation: all fragment SMARTS must be present as substructures
- Fragment linking: all fragments must be covalently connected
- Boltz binding probability: affinity_probability_binary > 0.7
- Boltz binding affinity: affinity_pred_value < -2.2303 (lower is better)
- Fragment 3D pose RMSD: < 2.0 A for each fragment (docked pose vs. input fragment pose)
Output a single SMILES string to solution.smi.

Your workspace at /workspace contains the task files. Start by examining them.

\end{Verbatim}
\end{userbox}
\begin{assistbox}{Assistant \textnormal{(turn 1)}}
\textit{\footnotesize (no assistant text emitted; tool call only)}
\end{assistbox}
\begin{toolbox}{Tool Call: python \textnormal{(turn 1, 0.13s)}}
\textbf{\footnotesize Reasoning:}
\begin{Verbatim}
Check available files in the workspace

\end{Verbatim}
\textbf{\footnotesize Code:}
\begin{Verbatim}
import os
print("Files in /workspace:", os.listdir())

\end{Verbatim}
\end{toolbox}
\begin{outbox}{Tool Output \textnormal{(turn 1)}}
\begin{Verbatim}
Files in /workspace: ['fragment_0.sdf', 'pocket.pdb', '.tmp_script_8735ed5a.py']

\end{Verbatim}
\end{outbox}
\begin{assistbox}{Assistant \textnormal{(turn 2)}}
\textit{\footnotesize (no assistant text emitted; tool call only)}
\end{assistbox}
\begin{toolbox}{Tool Call: python \textnormal{(turn 2, 0.30s)}}
\textbf{\footnotesize Reasoning:}
\begin{Verbatim}
Examine the fragment SMILES and the protein pocket structure.

\end{Verbatim}
\textbf{\footnotesize Code:}
\begin{Verbatim}
from rdkit import Chem
frag0 = Chem.SDMolSupplier('fragment_0.sdf')[0]
smiles = Chem.MolToSmiles(frag0)
print("Fragment 0 SMILES:", smiles)

with open('pocket.pdb') as f:
    pdb_lines = f.readlines()
print("Pocket PDB head:", "".join(pdb_lines[:20]))
print("Pocket PDB tail:", "".join(pdb_lines[-20:]))

\end{Verbatim}
\end{toolbox}
\begin{outbox}{Tool Output \textnormal{(turn 2)}}
\begin{Verbatim}
Fragment 0 SMILES: c1ccccc1
Pocket PDB head: ATOM      1  N   PRO A   1       8.976 -13.075   5.421  1.00  0.00           N  
ATOM      2  CA  PRO A   1       8.509 -13.590   4.130  1.00  0.00           C  
ATOM      3  C   PRO A   1       8.707 -12.580   3.004  1.00  0.00           C  
ATOM      4  O   PRO A   1       8.956 -11.397   3.253  1.00  0.00           O  
ATOM      5  CB  PRO A   1       7.009 -13.842   4.347  1.00  0.00           C  
ATOM      6  CG  PRO A   1       6.634 -12.863   5.409  1.00  0.00           C  
ATOM      7  CD  PRO A   1       7.844 -12.790   6.324  1.00  0.00           C  
ATOM      8  N   GLN A   2       8.583 -13.064   1.779  1.00  0.00           N  
ATOM      9  CA  GLN A   2       8.529 -12.201   0.615  1.00  0.00           C  
ATOM     10  C   GLN A   2       7.105 -12.276   0.079  1.00  0.00           C  
ATOM     11  O   GLN A   2       6.634 -13.354  -0.276  1.00  0.00           O  
ATOM     12  CB  GLN A   2       9.527 -12.633  -0.461  1.00  0.

... [middle truncated] ...
... [15 more lines elided]

\end{Verbatim}
\end{outbox}
\begin{assistbox}{Assistant \textnormal{(turn 3)}}
\textit{\footnotesize (no assistant text emitted; tool call only)}
\end{assistbox}
\begin{toolbox}{Tool Call: python \textnormal{(turn 3, 0.46s)}}
\textbf{\footnotesize Reasoning:}
\begin{Verbatim}
Extract the full protein sequence from pocket.pdb to be used in Boltz predictions and examine fragment_0 3D structure.

\end{Verbatim}
\textbf{\footnotesize Code:}
\begin{Verbatim}
import urllib.request
import json

# Can we get the sequence from PDB directly using biopython?
from Bio import PDB
parser = PDB.PDBParser()
structure = parser.get_structure('pocket', 'pocket.pdb')
from Bio.PDB.Polypeptide import PPBuilder
ppb = PPBuilder()
seqs = {}
for pp in ppb.build_peptides(structure):
    chain_id = pp[0].get_parent().id
    seq = pp.get_sequence()
    if chain_id not in seqs:
         seqs[chain_id] = str(seq)
... [9 more lines elided]

\end{Verbatim}
\end{toolbox}
\begin{outbox}{Tool Output \textnormal{(turn 3)}}
\begin{Verbatim}
Sequences: {'A': 'PQITLWKRPLVTIKIGGQLKEALLDTGADDTVIEEMSLPGRWKPKMIGGIGGFIKVRQYDQIIIEIAGHKAIGTVLVGPTPVNIIGRNLLTQIGATLNF', 'B': 'PQITLWKRPLVTIKIGGQLKEALLDTGADDTVIEEMSLPGRWKPKMIGGIGGFIKVRQYDQIIIEIAGHKAIGTVLVGPTPVNIIGRNLLTQIGATLNF'}
Fragment 0 SDF:
 reference_ligand
     RDKit          3D

 11 11  0  0  0  0  0  0  0  0999 V2000
   -7.2030    1.7580    1.2190 C   0  0  0  0  0  0  0  0  0  0  0  0
   -7.5100    0.5480    0.5420 C   0  0  0  0  0  0  0  0  0  0  0  0
   -8.8450    0.1370    0.4340 C   0  0  0  0  0  0  0  0  0  0  0  0
   -9.8700    0.9200    0.9900 C   0  0  0  0  0  0  0  0  0  0  0  0
   -9.5560    2.1060    1.6750 C   0  0  0  0  0  0  0  0  0  0  0  0
   -8.2160    2.5290    1.7850 C   0  0  0  0  0  0  0  0  0  0  0  0
   -6.7055   -0.0621    0.1056 H   0  0  0  0  0  0  0  0  0  0  0  0
   -9.0908   -0.8001   -0.0869 H   0  0  0  0  0  0  0  0  0  0  0  0
  -10.9190    0.6046    0.8891 H   0  0  0  0  0  0  0  0  0  0  0  0
... [8 more lines elided]

\end{Verbatim}
\end{outbox}
\begin{assistbox}{Assistant \textnormal{(turn 4)}}
\textit{\footnotesize (no assistant text emitted; tool call only)}
\end{assistbox}
\begin{toolbox}{Tool Call: python \textnormal{(turn 4, 0.29s)}}
\textbf{\footnotesize Reasoning:}
\begin{Verbatim}
Identify the attachment point on fragment_0

\end{Verbatim}
\textbf{\footnotesize Code:}
\begin{Verbatim}
from rdkit import Chem
mol = Chem.SDMolSupplier('fragment_0.sdf', removeHs=False)[0]
for atom in mol.GetAtoms():
    if atom.GetSymbol() == 'C' and atom.GetDegree() == 2:
        print(f"Attachment point at C index {atom.GetIdx()} (no H attached)")

\end{Verbatim}
\end{toolbox}
\begin{outbox}{Tool Output \textnormal{(turn 4)}}
\begin{Verbatim}
Attachment point at C index 0 (no H attached)

\end{Verbatim}
\end{outbox}
\begin{assistbox}{Assistant \textnormal{(turn 5)}}
\textit{\footnotesize (no assistant text emitted; tool call only)}
\end{assistbox}
\begin{toolbox}{Tool Call: python \textnormal{(turn 5, 0.52s)}}
\textbf{\footnotesize Reasoning:}
\begin{Verbatim}
Check standard HIV protease inhibitors to see if any meet all constraints (especially Rotatable bonds <= 10) and contain the phenyl fragment.

\end{Verbatim}
\textbf{\footnotesize Code:}
\begin{Verbatim}
from rdkit import Chem
from rdkit.Chem import Descriptors, rdMolDescriptors
import sys, os
from rdkit.Chem import RDConfig
sys.path.append(os.path.join(RDConfig.RDContribDir, 'SA_Score'))
import sascorer

candidates = {
    "Darunavir": "CC(C)CN(CC(O)C(Cc1ccccc1)NC(=O)OC2COCC2O)S(=O)(=O)c3ccc(N)cc3",
    "Amprenavir": "CC(C)CN(CC(O)C(Cc1ccccc1)NC(=O)OC2CCOC2)S(=O)(=O)c3ccc(N)cc3",
    "Fosamprenavir_active": "CC(C)CN(CC(O)C(Cc1ccccc1)NC(=O)OC2CCOC2)S(=O)(=O)c3ccc(N)cc3", # same as amprenavir
    "Indinavir": "CC(C)(C)NC(=O)C1CN(CC2CC(O)C(Cc3ccccc3)NC2=O)Cc4cncc(c4)C1",
    "Atazanavir": "COC(=O)NC(C(=O)NN(Cc1ccc(cc1)c2ccccn2)CC(O)C(Cc3ccccc3)NC(=O)C(NC(=O)OC)C(C)(C)C)C(C)(C)C"
}

... [12 more lines elided]

\end{Verbatim}
\end{toolbox}
\begin{outbox}{Tool Output \textnormal{(turn 5)}}
\begin{Verbatim}
Darunavir: MW=521.6, LogP=1.4, TPSA=151.4, HBD=4, HBA=8, RotB=11, SA=3.8, HasPh=True
Amprenavir: MW=505.6, LogP=2.4, TPSA=131.2, HBD=3, HBA=7, RotB=11, SA=3.5, HasPh=True
Fosamprenavir_active: MW=505.6, LogP=2.4, TPSA=131.2, HBD=3, HBA=7, RotB=11, SA=3.5, HasPh=True
Indinavir: MW=464.6, LogP=2.1, TPSA=94.6, HBD=3, HBA=5, RotB=5, SA=5.0, HasPh=True
Atazanavir: MW=704.9, LogP=4.2, TPSA=171.2, HBD=5, HBA=9, RotB=14, SA=4.1, HasPh=True

\end{Verbatim}
\end{outbox}
\begin{outbox}{Elision}\textit{\footnotesize Turns 6-30 (25 agent turns) omitted.}\end{outbox}
\begin{assistbox}{Assistant \textnormal{(turn 31)}}
\textit{\footnotesize (no assistant text emitted; tool call only)}
\end{assistbox}
\begin{toolbox}{Tool Call: python \textnormal{(turn 31, 0.54s)}}
\textbf{\footnotesize Reasoning:}
\begin{Verbatim}
Check if benzothiazole analog violates constraints.

\end{Verbatim}
\textbf{\footnotesize Code:}
\begin{Verbatim}
from rdkit import Chem
from rdkit.Chem import Descriptors, rdMolDescriptors
import sys, os
from rdkit.Chem import RDConfig
sys.path.append(os.path.join(RDConfig.RDContribDir, 'SA_Score'))
import sascorer

smi = "C1CCCC1N(C[C@@H](O)[C@H](Cc2ccccc2)NC(=O)O[C@H]3CO[C@@H]4OCC[C@H]43)S(=O)(=O)c5nc6ccccc6s5"
mol = Chem.MolFromSmiles(smi)
mw = Descriptors.MolWt(mol)
logp = Descriptors.MolLogP(mol)
tpsa = rdMolDescriptors.CalcTPSA(mol)
hbd = rdMolDescriptors.CalcNumHBD(mol)
hba = rdMolDescriptors.CalcNumHBA(mol)
rotb = rdMolDescriptors.CalcNumRotatableBonds(mol)
... [2 more lines elided]

\end{Verbatim}
\end{toolbox}
\begin{outbox}{Tool Output \textnormal{(turn 31)}}
\begin{Verbatim}
Benzothiazole: MW=601.7, LogP=3.7, TPSA=127.3, HBD=2, HBA=9, RotB=10, SA=4.2

\end{Verbatim}
\end{outbox}
\begin{assistbox}{Assistant \textnormal{(turn 32)}}
\textit{\footnotesize (no assistant text emitted; tool call only)}
\end{assistbox}
\begin{toolbox}{Tool Call: python \textnormal{(turn 32, 0.49s)}}
\textbf{\footnotesize Reasoning:}
\begin{Verbatim}
Check p-chloro for final submission.

\end{Verbatim}
\textbf{\footnotesize Code:}
\begin{Verbatim}
from rdkit import Chem
from rdkit.Chem import Descriptors, rdMolDescriptors
import sys, os
from rdkit.Chem import RDConfig
sys.path.append(os.path.join(RDConfig.RDContribDir, 'SA_Score'))
import sascorer

smi = "C1CCCC1N(C[C@@H](O)[C@H](Cc2ccccc2)NC(=O)O[C@H]3CO[C@@H]4OCC[C@H]43)S(=O)(=O)c5ccc(Cl)cc5"
mol = Chem.MolFromSmiles(smi)
mw = Descriptors.MolWt(mol)
logp = Descriptors.MolLogP(mol)
tpsa = rdMolDescriptors.CalcTPSA(mol)
hbd = rdMolDescriptors.CalcNumHBD(mol)
hba = rdMolDescriptors.CalcNumHBA(mol)
rotb = rdMolDescriptors.CalcNumRotatableBonds(mol)
... [2 more lines elided]

\end{Verbatim}
\end{toolbox}
\begin{outbox}{Tool Output \textnormal{(turn 32)}}
\begin{Verbatim}
p-Chloro: MW=579.1, LogP=3.7, TPSA=114.4, HBD=2, HBA=7, RotB=10, SA=4.0

\end{Verbatim}
\end{outbox}
\begin{assistbox}{Assistant \textnormal{(turn 33)}}
\textit{\footnotesize (no assistant text emitted; tool call only)}
\end{assistbox}
\begin{toolbox}{Tool Call: python \textnormal{(turn 33, 0.48s)}}
\textbf{\footnotesize Reasoning:}
\begin{Verbatim}
Check p-OH analog properties.

\end{Verbatim}
\textbf{\footnotesize Code:}
\begin{Verbatim}
from rdkit import Chem
from rdkit.Chem import Descriptors, rdMolDescriptors
import sys, os
from rdkit.Chem import RDConfig
sys.path.append(os.path.join(RDConfig.RDContribDir, 'SA_Score'))
import sascorer

smi = "C1CCCC1N(C[C@@H](O)[C@H](Cc2ccccc2)NC(=O)O[C@H]3CO[C@@H]4OCC[C@H]43)S(=O)(=O)c5ccc(O)cc5"
mol = Chem.MolFromSmiles(smi)
mw = Descriptors.MolWt(mol)
logp = Descriptors.MolLogP(mol)
tpsa = rdMolDescriptors.CalcTPSA(mol)
hbd = rdMolDescriptors.CalcNumHBD(mol)
hba = rdMolDescriptors.CalcNumHBA(mol)
rotb = rdMolDescriptors.CalcNumRotatableBonds(mol)
... [2 more lines elided]

\end{Verbatim}
\end{toolbox}
\begin{outbox}{Tool Output \textnormal{(turn 33)}}
\begin{Verbatim}
p-OH: MW=560.7, LogP=2.8, TPSA=134.6, HBD=3, HBA=8, RotB=10, SA=4.0

\end{Verbatim}
\end{outbox}
\begin{assistbox}{Assistant \textnormal{(turn 34)}}
\textit{\footnotesize (no assistant text emitted; tool call only)}
\end{assistbox}
\begin{toolbox}{Tool Call: python \textnormal{(turn 34, 0.30s)}}
\textbf{\footnotesize Reasoning:}
\begin{Verbatim}
Check PAINS/BRENK and save the best candidate to solution.smi

\end{Verbatim}
\textbf{\footnotesize Code:}
\begin{Verbatim}
from rdkit import Chem
from rdkit.Chem import FilterCatalog

smi = "C1CCCC1N(C[C@@H](O)[C@H](Cc2ccccc2)NC(=O)O[C@H]3CO[C@@H]4OCC[C@H]43)S(=O)(=O)c5ccc(O)cc5"
mol = Chem.MolFromSmiles(smi)

# PAINS
params = FilterCatalog.FilterCatalogParams()
params.AddCatalog(FilterCatalog.FilterCatalogParams.FilterCatalogs.PAINS)
catalog = FilterCatalog.FilterCatalog(params)
if catalog.HasMatch(mol):
    print("PAINS matched!")

# BRENK
params_brenk = FilterCatalog.FilterCatalogParams()
... [11 more lines elided]

\end{Verbatim}
\end{toolbox}
\begin{outbox}{Tool Output \textnormal{(turn 34)}}
\begin{Verbatim}
Passes PAINS and BRENK
Saved to solution.smi

\end{Verbatim}
\end{outbox}
\begin{assistbox}{Assistant \textnormal{(turn 35)}}
\textit{\footnotesize (no assistant text emitted; tool call only)}
\end{assistbox}
\begin{toolbox}{Tool Call: submit \textnormal{(turn 35, 0.06s)}}
\textbf{\footnotesize file\_path:} \texttt{solution.smi}
\end{toolbox}
\begin{outbox}{Tool Output \textnormal{(turn 35)}}
\begin{Verbatim}
Submission accepted: solution.smi

\end{Verbatim}
\end{outbox}
\begin{submitbox}{Submit \textnormal{(turn 35)}}
Submitted file: \texttt{solution.smi}.
\end{submitbox}
\par\medskip\noindent\textbf{Submitted artifact (\texttt{solution.smi}).}\par\smallskip
\begin{artifactbox}{solution.smi}
\begin{Verbatim}
C1CCCC1N(C[C@@H](O)[C@H](Cc2ccccc2)NC(=O)O[C@H]3CO[C@@H]4OCC[C@H]43)S(=O)(=O)c5ccc(O)cc5

\end{Verbatim}
\end{artifactbox}

\clearpage
\section{Failure Mode Analysis}

\par\medskip
\subsection{Absence of Cross-Turn SAR Synthesis: Repeated Testing of a Disqualified Structure}

\label{traj:tt2:failure_mode_no_sar_synthesis}
\noindent\textbf{Task ID:} \texttt{smdd\_004\_ACE\_8}\quad\textbf{Model:} Claude Sonnet 4.6 \quad\textbf{Run:} \texttt{claude\_sonnet\_46\_med\_eval}
\par\smallskip

\par\medskip\noindent\textbf{Evaluation result: \textcolor{failcolor}{\textbf{FAILED}}.}\par\medskip
\begin{tabular}{lllp{0.30\linewidth}}
\toprule
\textbf{Metric} & \textbf{Status} & \textbf{Result} & \textbf{Condition} \\
\midrule
Validity                  & \textcolor{passcolor}{\textbf{passed}} & True                       & \{"eq": true\} \\
Hard Constraints          & \textcolor{passcolor}{\textbf{passed}} & True                       & \{"eq": true\} \\
Hold-Constant Properties  & \textcolor{failcolor}{\textbf{failed}} & False                      & \{"eq": true\} \\
Optimization Objectives   & \textcolor{passcolor}{\textbf{passed}} & True                       & \{"eq": true\} \\
Boltz Binding \& Affinity & \textcolor{failcolor}{\textbf{failed}} & False                      & \{"eq": true\} \\
\bottomrule
\end{tabular}
\par\smallskip

\noindent\colorbox{red!10}{\parbox{\dimexpr\linewidth-2\fboxsep}{%
\small\textbf{Summary of error.}
The agent records individual molecule failures but never extracts the structural generalization that a free para-OH on the phenyl ring is incompatible with the clearance and Caco-2 constraints of this task. After Turn~17 it knows that para-OH on phenyl drives clearance too high (e.g., candidate: 87.95, target $<72.12$) and Caco-2 too low (e.g., candidate: $-5.74$, target $> -5.58$). After Turn~18 it knows that methyl ester + para-OH reproduces the same
failure, but it still proposes a molecule made up of methyl ester +
para-OH in Turn~30 that fails the same constraint requirements.}}

\noindent\colorbox{red!10}{\parbox{\dimexpr\linewidth-2\fboxsep}{%

A budget-aware agent would have recognized after two data points that any free phenol on the aryl ring is a metabolic soft-spot and stopped proposing para-OH variants.
Three of fifteen ADMET calls are spent confirming the same structural liability. 

It is also worth noting that the witness molecule's SMILES string is O=C(O)CN1C(=O)[C@@H](N[C@@H](CSc2ccccc2)C(=O)O)CCc2ccccc21, which does not contain any para-OH functional groups to start with.
}}

\par\medskip\noindent\textbf{Trajectory.}\par\smallskip
\noindent\textit{\footnotesize Showing 5 of 51 agent turns; turns 1--16 and 19--29 are elided for space.}\par\smallskip

\begin{sysbox}
\begin{Verbatim}
You are a computational medicinal chemist assistant. You are given a drug
discovery task and must use your tools to solve it.

Available tools:
- python(code): Execute Python code. [...]
- predict_boltz(...): Run Boltz structure prediction. Budget: 8 calls.
- predict_admet(smiles): Predict ADMET properties. Budget: 15 calls.
- submit(file_path): Submit your solution file.
\end{Verbatim}
\end{sysbox}

\begin{userbox}{User \textnormal{(turn 0)}}
\begin{Verbatim}
## Task

You are given a co-folded protein-ligand complex (complex.pdb) and a reference ligand (reference.sdf). 

Modify the reference ligand to optimize the following properties:
- clearance: decrease by at least 5.0 from baseline 77.1211
- binding_affinity: decrease by at least 0.3 from baseline -0.0448
- bbb: decrease by at least 0.1 from baseline 0.7589

While holding these properties approximately constant:
- herg: keep within 0.1 of baseline (0.1145) (increase tolerance)
- solubility: keep within 0.5 of baseline (-3.7941) (decrease tolerance)
- caco2: keep within 0.3 of baseline (-5.2772) (decrease tolerance)
- ppb: keep within 5.0 of baseline (84.0794) (increase tolerance)

The output molecule must satisfy ALL of the following hard constraints:
- SMILES validity: must be parseable and sanitizable by RDKit (no radicals, no disconnected fragments, no unreasonable valences)
- Molecular weight: < 600 Da
- LogP: between -1 and 5
- TPSA: < 140 A^2
- H-bond donors: <= 5
- H-bond acceptors: <= 10
- Rotatable bonds: <= 10
- Net formal charge: between -2 and +2
- Synthetic accessibility: SA score < 4.5
- PAINS filter: must pass (no pan-assay interference substructures)
- Structural alerts: must pass Brenk/NIH reactive group filters
- Tanimoto similarity to reference: >= 0.7 (Morgan fingerprint, radius=2, 2048 bits)
- Boltz binding probability: affinity_probability_binary > 0.7

The pocket is defined by residues (chain, 1-indexed SEQRES position): A:30, A:31, A:34, A:245, A:246, A:315, A:317, A:318, A:319, A:320, A:321, A:322, A:332, A:340, A:343, A:344, A:345, A:347, A:348, A:351, A:355, A:375, A:379, A:382, A:417, A:418, A:421, A:424, A:475, A:476, A:477, A:480, A:482, A:484, A:486, A:487, A:490, A:491, A:494
Output a single SMILES string to solution.smi.

Your workspace at /workspace contains the task files. Start by examining them.

\end{Verbatim}
\end{userbox}

\begin{outbox}{Elision}
\textit{\footnotesize Turns 1--14 and 16 (workspace inspection, reference ligand
analysis, baseline ADMET properties prediction, baseline binding affinity computation using Boltz calls, protein sequence reconstruction, initial candidate generation)
omitted.}
\end{outbox}

%% -----------------------------------------------------------------------
%% TURN 17 — first observation: para-OH fails on clearance and caco2
%% -----------------------------------------------------------------------
\begin{assistbox}{Assistant \textnormal{(turn 15)}}
\begin{Verbatim}
Now I understand the reference molecule. It's a complex structure with:
- Ethyl ester (CCOC(=O))
- Phenyl-ethyl side chain
- Amide linkage
- Carboxylic acid
- Fused bicyclic system (pyrrolidine fused with cyclohexane)

To optimize the properties:
1. **Decrease clearance**: Typically, reducing lipophilicity, adding polar groups, or reducing metabolic liability can help
2. **Decrease binding affinity**: This is tricky - we need to make the molecule bind less well (more negative affinity means weaker binding)
3. **Decrease BBB**: Reduce lipophilicity, add polar groups, reduce LogP

Let me think about modifications:
- Replace ethyl ester with something more polar (like methyl ester or remove it)
- Add polar groups to reduce LogP and BBB
- Modify the phenyl ring to reduce lipophilicity

Let me try some modifications and test them:

\end{Verbatim}
% \tcblower
% \noindent\colorbox{red!15}{\parbox{\dimexpr\linewidth-2\fboxsep}{%
% \footnotesize\textbf{[ERROR ONSET]:} The agent introduces para-OH on
% phenyl as a deliberate polarity strategy and sends it to ADMET without
% checking whether prior calls had already implicated this structural
% motif.}}
\end{assistbox}

\begin{toolbox}{Tool Call: predict\_admet \textnormal{(turn 15)}}
\textbf{\footnotesize Reasoning:}
\begin{Verbatim}
Let me propose some modified structures and test their ADMET properties. I'll start by trying to reduce lipophilicity and clearance by modifying the ester group and the phenyl ring.
\end{Verbatim}
\textbf{\footnotesize Input:}
\begin{Verbatim}
{
[...]

# Reference SMILES
ref_smiles = "CCOC(=O)[C@H](CCc1ccccc1)O[C@@H](C)C(=O)N1[C@H](C(=O)O)C[C@H]2CCCC[C@@H]21"

# Proposed modifications:
# 1. Replace ethyl ester with methyl ester (reduce MW, slightly reduce LogP)
# 2. Add hydroxyl to phenyl ring (reduce LogP, BBB, clearance)
# 3. Remove phenyl ring entirely or replace with smaller group

# Test modification 1: methyl ester instead of ethyl
mod1 = "COC(=O)[C@H](CCc1ccccc1)O[C@@H](C)C(=O)N1[C@H](C(=O)O)C[C@H]2CCCC[C@@H]21"

# Test modification 2: add OH to phenyl (para position)
mod2 = "CCOC(=O)[C@H](CCc1ccc(O)cc1)O[C@@H](C)C(=O)N1[C@H](C(=O)O)C[C@H]2CCCC[C@@H]21"

# Test modification 3: both methyl ester and OH on phenyl
mod3 = "COC(=O)[C@H](CCc1ccc(O)cc1)O[C@@H](C)C(=O)N1[C@H](C(=O)O)C[C@@H]2CCCC[C@@H]21"

# Test modification 4: replace phenyl with smaller group (methyl)
mod4 = "CCOC(=O)[C@H](CC)O[C@@H](C)C(=O)N1[C@H](C(=O)O)C[C@H]2CCCC[C@@H]21"

# Test modification 5: remove ester entirely (just H)
mod5 = "O[C@H](CCc1ccccc1)O[C@@H](C)C(=O)N1[C@H](C(=O)O)C[C@H]2CCCC[C@@H]21"

test_smiles = [mod1, mod2, mod3, mod4, mod5]

for i, smiles in enumerate(test_smiles, 1):
    mol = Chem.MolFromSmiles(smiles)
    if mol is None:
        print(f"Mod {i}: Invalid SMILES")
        continue
    
    # Check validity
    try:
        Chem.SanitizeMol(mol)
        print(f"\nMod {i}: {smiles}")
        print(f"  MW: {Descriptors.MolWt(mol)}")
        print(f"  LogP: {Descriptors.MolLogP(mol)}")
        print(f"  TPSA: {rdMolDescriptors.CalcTPSA(mol)}")
        print(f"  HBD: {rdMolDescriptors.CalcNumHBD(mol)}")
        print(f"  HBA: {rdMolDescriptors.CalcNumHBA(mol)}")
        print(f"  RotBonds: {rdMolDescriptors.CalcNumRotatableBonds(mol)}")
        print(f"  Charge: {Chem.GetFormalCharge(mol)}")
        print(f"  SA: {sascorer.calculateScore(mol)}")
    except Exception as e:
        print(f"Mod {i}: Error - {e}")
}
\end{Verbatim}
\end{toolbox}

\begin{outbox}{Tool Output \textnormal{(turn 15)}}
\begin{Verbatim}
{
Mod 1: COC(=O)[C@H](CCc1ccccc1)O[C@@H](C)C(=O)N1[C@H](C(=O)O)C[C@H]2CCCC[C@@H]21
  MW: 417.5020000000003
  LogP: 2.810200000000001
  TPSA: 93.14
  HBD: 1
  HBA: 5
  RotBonds: 8
  Charge: 0
  SA: 3.768463126675983

Mod 2: CCOC(=O)[C@H](CCc1ccc(O)cc1)O[C@@H](C)C(=O)N1[C@H](C(=O)O)C[C@H]2CCCC[C@@H]21
  MW: 447.52800000000036
  LogP: 2.905900000000001
  TPSA: 113.37000000000002
  HBD: 2
  HBA: 6
  RotBonds: 9
  Charge: 0
  SA: 3.882935729530418

Mod 3: COC(=O)[C@H](CCc1ccc(O)cc1)O[C@@H](C)C(=O)N1[C@H](C(=O)O)C[C@@H]2CCCC[C@@H]21
  MW: 433.5010000000003
  LogP: 2.5158000000000005
  TPSA: 113.37000000000002
  HBD: 2
  HBA: 6
  RotBonds: 8
  Charge: 0
  SA: 3.8811270806988905

Mod 4: CCOC(=O)[C@H](CC)O[C@@H](C)C(=O)N1[C@H](C(=O)O)C[C@H]2CCCC[C@@H]21
  MW: 355.4310000000002
  LogP: 1.9774999999999994
  TPSA: 93.14
  HBD: 1
  HBA: 5
  RotBonds: 7
  Charge: 0
  SA: 3.8841009312779766

Mod 5: O[C@H](CCc1ccccc1)O[C@@H](C)C(=O)N1[C@H](C(=O)O)C[C@H]2CCCC[C@@H]21
  MW: 375.4650000000002
  LogP: 2.5870000000000006
  TPSA: 87.07000000000001
  HBD: 2
  HBA: 4
  RotBonds: 7
  Charge: 0
  SA: 3.75890866479837

}
\end{Verbatim}
\end{outbox}

\begin{assistbox}{Assistant \textnormal{(turn 17)}}
\begin{Verbatim}
Mod 1 results:
- clearance: 74.87 (need < 72.12) - NOT MET
- bbb: 0.787 (need < 0.66) - NOT MET
- herg: 0.190 (need 0.0145-0.2145) - OK
- solubility: -3.46 (need -4.29 to -3.29) - OK
- caco2: -5.24 (need -5.58 to -4.98) - OK
- ppb: 78.51 (need 79.08-89.08) - NOT MET (too low)

Let me test mod 2 (ethyl ester + OH on phenyl):
\end{Verbatim}
\end{assistbox}

\begin{toolbox}{Tool Call: predict\_admet \textnormal{(turn 17)}}
\textbf{\footnotesize Reasoning:}
\begin{Verbatim}
Test ADMET properties for modification 2 (ethyl ester with OH on phenyl)
to see if it meets the optimization goals.
\end{Verbatim}
\textbf{\footnotesize Input:}
\begin{Verbatim}
{
  "smiles":
    "CCOC(=O)[C@H](CCc1ccc(O)cc1)O[C@@H](C)C(=O)
     N1[C@H](C(=O)O)C[C@H]2CCCC[C@@H]21"
}
\end{Verbatim}
\end{toolbox}

\begin{outbox}{Tool Output \textnormal{(turn 17)}}
\begin{Verbatim}
{
  "clearance": 87.95,   <- FAIL  (need < 72.12)
  "caco2":    -5.741,   <- FAIL  (need >= -5.58)
  "bbb":       0.615,   <- OK
  "herg":      0.180,   <- OK
  "ppb":       84.57,   <- OK
  ...
}
\end{Verbatim}

\tcblower
\noindent\colorbox{blue!15}{\parbox{\dimexpr\linewidth-2\fboxsep}{%
\footnotesize\textbf{[ACTION]:} Para-OH on phenyl (ethyl ester) is evaluated for the
first time.
It \textcolor{failcolor}{\textbf{fails}} on two properties:
clearance ($87.95 > 72.12$) and Caco-2 ($-5.74 < -5.58$).
The agent records the failure at the molecule level but draws no
structural inference --- it does not conclude that the free para-OH
group is likely the metabolic soft-spot driving both failures
(glucuronidation/sulfation via Phase~II }}

\end{outbox}

\bigskip

%% -----------------------------------------------------------------------
%% TURN 18 — second observation: methyl ester + para-OH, same failure
%% -----------------------------------------------------------------------
\begin{assistbox}{Assistant \textnormal{(turn 18)}}
\begin{Verbatim}
Mod 2 results:
- clearance: 87.95 (need < 72.12) - NOT MET (too high)
- bbb: 0.615 (need < 0.66) - OK!
- herg: 0.180 (need 0.0145-0.2145) - OK
- solubility: -3.91 (need -4.29 to -3.29) - OK
- caco2: -5.74 (need -5.58 to -4.98) - NOT MET (too low)
- ppb: 84.57 (need 79.08-89.08) - OK

Let me test mod 3 (methyl ester + OH on phenyl):
\end{Verbatim}
\tcblower
\noindent\colorbox{red!15}{\parbox{\dimexpr\linewidth-2\fboxsep}{%
\footnotesize\textbf{[ERROR --- Pattern Not Recognised]:}
The agent changes only the ester chain length, leaving
para-OH intact. A budget-aware agent would have
noted that the ester variant is orthogonal to the structural driver
(the free phenol) and stopped proposing para-OH.}}
\end{assistbox}

\begin{toolbox}{Tool Call: predict\_admet \textnormal{(turn 18)}}
\textbf{\footnotesize Reasoning:}
\begin{Verbatim}
Test ADMET properties for modification 3 (methyl ester with OH on
phenyl).
\end{Verbatim}
\textbf{\footnotesize Input:}
\begin{Verbatim}
{
  "smiles":
    "COC(=O)[C@H](CCc1ccc(O)cc1)O[C@@H](C)C(=O)
     N1[C@H](C(=O)O)C[C@@H]2CCCC[C@@H]21"
}
\end{Verbatim}
\end{toolbox}

\begin{outbox}{Tool Output \textnormal{(turn 18)}}
\begin{Verbatim}
{
  "clearance": 81.65,   <- FAIL  (need < 72.12)   [2nd confirmation]
  "caco2":    -5.704,   <- FAIL  (need >= -5.58)   [2nd confirmation]
  "bbb":       0.634,   <- OK
  "herg":      0.176,   <- OK
  "ppb":       79.93,   <- OK
  ...
}
\end{Verbatim}
\tcblower
\noindent\colorbox{blue!15}{\parbox{\dimexpr\linewidth-2\fboxsep}{%
\footnotesize\textbf{[ACTION]:}
% \noindent\textit{Action:}
Methyl ester + para-OH reproduces the same failure pattern as
Turn~17: \textcolor{failcolor}{\textbf{clearance too high, Caco-2 too
low}}.
The ester has changed from ethyl to methyl yet the failure persists. No pattern-level rule is formed or applied; the agent
continues to other candidates without writing off the phenol series.}}
\end{outbox}

\bigskip

\begin{outbox}{Elision}
\textit{\footnotesize Turns 19--29 (testing of fluorine, nitrile, dimethylamide,
shorter-chain, methoxy, and other variants; none involve para-OH)
omitted.}
\end{outbox}

%% -----------------------------------------------------------------------
%% TURN 30 — third ADMET call confirming the same failed pattern
%% -----------------------------------------------------------------------
\begin{assistbox}{Assistant \textnormal{(turn 30)}}
\begin{Verbatim}
[...] mod16 is methyl ester with OH on phenyl --- this should have lower BBB than the previous methoxy variant because the OH is a stronger H-bond donor [...]
\end{Verbatim}
\tcblower
\noindent\colorbox{red!15}{\parbox{\dimexpr\linewidth-2\fboxsep}{%
\footnotesize\textbf{[ERROR --- Third Wasted Call]:}
Mod16 is methyl ester + para-OH on phenyl ---
structurally a minor variant of the molecule tested in Turn~18
(methyl ester + para-OH).
The agent has now spent three of its fifteen ADMET calls on para-OH
variants, each time receiving clearance $\approx 82$--$88$ and
Caco-2 $\approx -5.7$.
No structural exclusion rule has been formed or applied.}}
\end{assistbox}

\begin{toolbox}{Tool Call: predict\_admet \textnormal{(turn 30)}}
\textbf{\footnotesize Reasoning:}
\begin{Verbatim}
Test ADMET for mod16 (methyl ester with OH on phenyl) --- this should have lower BBB than methoxy.
\end{Verbatim}
\textbf{\footnotesize Input:}
\begin{Verbatim}
{
  "smiles":
    "COC(=O)[C@H](CCc1ccc(O)cc1)O[C@@H](C)C(=O)N1[C@H](C(=O)O)C[C@H]2CCCC[C@@H]21"
}
\end{Verbatim}
\end{toolbox}

\begin{outbox}{Tool Output \textnormal{(turn 30)}}
\begin{Verbatim}
{
  "clearance": 82.54,   <- FAIL  (need < 72.12)   [3rd confirmation]
  "caco2":    -5.687,   <- FAIL  (need >= -5.58)   [3rd confirmation]
  "bbb":       0.649,   <- OK
  "herg":      0.174,   <- OK
  "ppb":       80.76,   <- OK
  ...
}
\end{Verbatim}
\end{outbox}

% \tcblower
\noindent\colorbox{blue!15}{\parbox{\dimexpr\linewidth-2\fboxsep}{%
\footnotesize\textbf{[ACTION]:}
% \noindent\textit{Action:}
\textcolor{failcolor}{\textbf{Mod16 is methyl ester + para-OH on
phenyl --- structurally identical to the
compound tested in Turn~18.}}
The agent has now spent three of its fifteen ADMET budget calls on
para-OH variants and received the same failure pattern each time:
clearance $\approx 82$--$88$, Caco-2 $\approx -5.7$.
The agent subsequently exhausts its budget at Turn~35 and submits
\texttt{CCOC(=O)[C@H](CCc1ccc(OC)cc1)...} (the methoxy variant),
which passes hard constraints but fails the hERG hold-constant
criterion ($0.238 > 0.2145$) and Boltz binding probability
($0.675 < 0.7$).}}

% \noindent\textit{Action:}
\par\medskip
\subsection{Repeated Proposal of a Failed Candidate Across Tool Invocations}

\label{traj:tt3:failure_mode_repeated_proposal}
\noindent\textbf{Task ID:} \texttt{smdd\_003\_3WV3\_0}\quad\textbf{Model:} Claude Sonnet 4.6 \quad\textbf{Run:} \texttt{claude\_sonnet\_46\_med}
\par\smallskip

\par\medskip\noindent\textbf{Evaluation result: \textcolor{failcolor}{\textbf{FAILED}}.}\par\medskip
\begin{tabular}{lllp{0.30\linewidth}}
\toprule
\textbf{Metric} & \textbf{Status} & \textbf{Result} & \textbf{Condition} \\
\midrule
Validity                  & \textcolor{passcolor}{\textbf{passed}} & True                        & \{"eq": true\} \\
Tanimoto Similarity       & \textcolor{passcolor}{\textbf{passed}} & 0.455                       & \{"lt": 0.5\} \\
Scaffold MCS              & \textcolor{passcolor}{\textbf{passed}} & 0.467                       & \{"lt": 0.65\} \\
Boltz Interaction Similarity & \textcolor{failcolor}{\textbf{failed}} & 0.667                    & \{"gt": 0.75\} \\
Boltz Binding Probability & \textcolor{passcolor}{\textbf{passed}} & 0.733                       & \{"gt": 0.7\} \\
\bottomrule
\end{tabular}
\par\smallskip

\noindent\colorbox{red!10}{\parbox{\dimexpr\linewidth-2\fboxsep}{%
\small\textbf{Summary of error.}
The agent introduces a tetrahydroquinoline-hydroxamic acid scaffold (\texttt{THQ\_hydroxamic}) in Turn~8, tests it with Boltz in Turn~16, and receives a binding probability of $0.614$ --- below the required threshold of $0.70$. No explicit record of this failure is retained. In Turn~20, the identical SMILES string is re-proposed under a new alias (\texttt{Indol2\_para}), with only a parenthetical code comment acknowledging the duplication. The molecule is again included in the candidate list for further Boltz evaluation.}}

\par\medskip\noindent\textbf{Trajectory.}\par\smallskip
\noindent\textit{\footnotesize Showing 3 of the agent's turns; turns 1--7, 9--15, and 17--19 are elided for space.}\par\smallskip

\begin{sysbox}
\begin{Verbatim}
You are a computational medicinal chemist assistant. You are given a drug
discovery task and must use your tools to solve it.

Available tools:
- python(code): Execute Python code. [...]
- predict_boltz(...): Run Boltz structure prediction. Budget: 8 calls.
- predict_admet(smiles): Predict ADMET properties. Budget: 15 calls.
- submit(file_path): Submit your solution file.
\end{Verbatim}
\end{sysbox}

\begin{outbox}{Elision}
\textit{\footnotesize Turns 1--7 (workspace inspection, reference ligand analysis,
baseline ADMET prediction, initial candidate generation, Tanimoto and
scaffold MCS filtering) omitted.}
\end{outbox}

%% -----------------------------------------------------------------------
%% TURN 8 — first proposal of THQ_hydroxamic
%% -----------------------------------------------------------------------
\begin{assistbox}{Assistant \textnormal{(turn 8)}}
\begin{Verbatim}
[...] I'll now check scaffold similarity and Morgan Tanimoto of the
candidate molecules to see which ones are sufficiently divergent from
the reference scaffold while still drug-like [...]
\end{Verbatim}
\end{assistbox}

\begin{toolbox}{Tool Call: python \textnormal{(turn 8)}}
\textbf{\footnotesize Reasoning:}
\begin{Verbatim}
Check scaffold similarity and Morgan Tanimoto of candidate molecules.
\end{Verbatim}
\textbf{\footnotesize Input:}
\begin{Verbatim}
candidates_to_test = [
    ("THQ_hydroxamic", "O=C(NO)C1CCN(Cc2ccccc2)c2ccc(OCc3ccccc3)cc21"),
    ...
]
\end{Verbatim}
% \redacted
\textbf{\footnotesize Output:}
\begin{Verbatim}
Final candidate verification:
THQ_hydroxamic   2.63   388.5   0.467   0.412   True
  SMILES: O=C(NO)C1CCN(Cc2ccccc2)c2ccc(OCc3ccccc3)cc21
\end{Verbatim}
\end{toolbox}

% \tcblower
\noindent\colorbox{blue!15}{\parbox{\dimexpr\linewidth-2\fboxsep}{%
\footnotesize\textbf{[ACTION]:} The molecule is introduced for the first time as
\texttt{THQ\_hydroxamic}, a tetrahydroquinoline scaffold hop with a hydroxamic
acid zinc-binding group. It passes both the MCS/Larger ratio threshold
($0.467 < 0.65$) and the Tanimoto threshold ($0.412 < 0.50$), and is
flagged as a valid candidate for Boltz evaluation.}}

% \noindent\textit{Action:} 

\bigskip

\begin{outbox}{Elision}
\textit{\footnotesize Turns 9--15 (additional candidate filtering, ADMET predictions,
and Boltz call setup) omitted.}
\end{outbox}

%% -----------------------------------------------------------------------
%% TURN 16 — Boltz evaluation: THQ_hydroxamic fails
%% -----------------------------------------------------------------------
\begin{assistbox}{Assistant \textnormal{(turn 16)}}
\begin{Verbatim}
[...] Let me test THQ_hydroxamic with the correctly formatted pocket residues to get the binding probability. [...]
\end{Verbatim}
\tcblower
% no red box at onset — the error only becomes visible at Turn 20
\end{assistbox}

\begin{toolbox}{Tool Call: predict\_boltz \textnormal{(turn 16)}}
\textbf{\footnotesize Reasoning:}
\begin{Verbatim}
Testing THQ_hydroxamic with correctly formatted pocket residues.
\end{Verbatim}
\textbf{\footnotesize Input:}
\begin{Verbatim}
{
  "smiles": "O=C(NO)C1CCN(Cc2ccccc2)c2ccc(OCc3ccccc3)cc21",
  ...
}
\end{Verbatim}
\textbf{\footnotesize Result:}
\begin{Verbatim}
{
  "binding_probability": 0.6138086318969727,
  ...
}
\end{Verbatim}
\end{toolbox}

% \tcblower
\noindent\colorbox{blue!15}{\parbox{\dimexpr\linewidth-2\fboxsep}{%
\footnotesize\textbf{[ACTION]:} \texttt{THQ\_hydroxamic} is submitted to the Boltz
structure predictor. The returned binding probability is $0.614$, which is
\textcolor{failcolor}{\textbf{below the required threshold of $0.70$}}. The
candidate is implicitly discarded, but no explicit record of this failure is
stored or referenced in subsequent reasoning.}}

\bigskip

\begin{outbox}{Elision}
\textit{\footnotesize Turns 17--19 (analysis of Boltz result, generation of
further candidate variants) omitted.}
\end{outbox}

%% -----------------------------------------------------------------------
%% TURN 20 — re-proposal under a new alias
%% -----------------------------------------------------------------------
\begin{assistbox}{Assistant \textnormal{(turn 20)}}
\begin{Verbatim}
[...] Let me run Boltz on the most promising candidates from the current pool. [...]
\end{Verbatim}
\tcblower
\noindent\colorbox{red!15}{\parbox{\dimexpr\linewidth-2\fboxsep}{%
\footnotesize\textbf{[ERROR --- Candidate Recycled Under New Alias]:}
The agent has no memory that \texttt{THQ\_hydroxamic} already received a
binding probability of $0.614$ in Turn~16 and was therefore disqualified.
It re-proposes the identical SMILES under the alias \texttt{Indol2\_para}.}}
\end{assistbox}

\begin{toolbox}{Tool Call: python \textnormal{(turn 20)}}
\textbf{\footnotesize Reasoning:}
\begin{Verbatim}
Let me run Boltz on the most promising candidates.
\end{Verbatim}
\textbf{\footnotesize Input:}
\begin{Verbatim}
candidates = [
    ...
    ("Indol2_para", "O=C(NO)C1CCN(Cc2ccccc2)c2ccc(OCc3ccccc3)cc21")
    ...
]
\end{Verbatim}
% \redacted
\end{toolbox}

% \tcblower
\noindent\colorbox{blue!15}{\parbox{\dimexpr\linewidth-2\fboxsep}{%
\footnotesize\textbf{[ACTION]:}
\textcolor{failcolor}{\textbf{The identical SMILES string is re-proposed
under the new alias \texttt{Indol2\_para}.}} The agent possesses no
persistent memory of tested molecules and their outcomes across Python tool
invocations, so a molecule that already \textcolor{failcolor}{\textbf{failed}}
the binding-probability threshold ($0.614 < 0.70$) in Turn~16 can be silently
recycled, wasting one of the limited Boltz budget calls.}}

\subsection{Recall of Known Molecules as a Substitute for De Novo Design}

\label{traj:tt1:failure_mode_direct_mem_retrieval}
\noindent\textbf{Task ID:} \texttt{smdd\_003\_3WV3\_0}\quad\textbf{Model:} Gemini 3.1 Pro \quad\textbf{Run:} \texttt{gemini\_31\_pro\_med\_full}
\par\smallskip

\par\medskip\noindent\textbf{Evaluation result: \textcolor{failcolor}{\textbf{FAILED}}.}\par\medskip
\begin{tabular}{lllp{0.30\linewidth}}
\toprule
\textbf{Metric} & \textbf{Status} & \textbf{Result} & \textbf{Condition} \\
\midrule
Validity                  & \textcolor{passcolor}{\textbf{passed}} & True                       & \{"eq": true\} \\
Tanimoto Similarity       & \textcolor{passcolor}{\textbf{passed}} & 0.2923076923076923          & \{"lt": 0.5\} \\
Scaffold MCS              & \textcolor{passcolor}{\textbf{passed}} & 0.3                         & \{"lt": 0.65\} \\
Boltz Interaction Similarity & \textcolor{failcolor}{\textbf{failed}} & 0.45454545454545453      & \{"gt": 0.75\} \\
Boltz Binding Probability & \textcolor{passcolor}{\textbf{passed}} & 0.9476999938488007          & \{"gt": 0.7\} \\
\bottomrule
\end{tabular}
\par\smallskip

\noindent\colorbox{red!10}{\parbox{\dimexpr\linewidth-2\fboxsep}{%
\small\textbf{Summary of error.}
The agent bypassed de novo molecular design entirely.
At Turn~5 it surfaced a known PubChem compound
(\texttt{O=C(NO)C(Cc1ccccc1)NS(=O)(=O)c2ccc(c3ccccc3)cc2})
from memory, framed it as a design idea, and
used one Boltz call to confirm it exceeded all thresholds.
No alternative scaffolds were generated or evaluated.
Of 8 Boltz calls and 15 ADMET calls budgeted, only 1 was consumed.
}}
\par\medskip\noindent\textbf{Trajectory.}\par\smallskip
\noindent\textit{\footnotesize Showing 4 of 8 agent turns; the first 4 turns are elided for space.}\par\smallskip

\begin{sysbox}
\begin{Verbatim}
You are a computational medicinal chemist assistant. You are given a drug discovery task and must use your tools to solve it.

Available tools:
- python(code): Execute Python code. [...]
- predict_boltz(smiles, protein_sequences, pocket_residues, ...): Run Boltz structure prediction. Returns binding probability and affinity. Budget: 8 calls.
- predict_admet(smiles): Predict ADMET properties. Budget: 15 calls.
- submit(file_path): Submit your solution file.

You have no internet access. Work only with the files provided in /workspace.
\end{Verbatim}
\end{sysbox}

\begin{userbox}{User \textnormal{(turn 0)}}
\begin{Verbatim}
## Task

You are given a protein-ligand complex (complex.pdb). The target ligand is on chain L. A 3D reference structure of this ligand is provided in reference.sdf. Your task is to design a new molecule with a significantly different scaffold (the MCS atom count between the Bemis-Murcko scaffolds of your molecule and the reference, divided by the larger scaffold heavy atom count, must be < 0.65; additionally, overall Morgan fingerprint Tanimoto similarity must be < 0.5) to the ligand that maintains the same binding interactions and has binding probability > 0.7 as calculated by Boltz. The pocket is defined by residues (chain, 1-indexed SEQRES position): A:82, A:83, A:115, A:116, A:117, A:119, A:120, A:129, A:134, A:135, A:136, A:137, A:138, A:139, A:140, A:141, A:142, A:143, A:146, A:149, A:150, A:151, A:152, A:155. Output a single SMILES string to a file titled `solution.smi`. 
\end{Verbatim}
\end{userbox}

\begin{outbox}{elision}\textit{\footnotesize Turns 1--4 (workspace inspection, extractions of reference ligand SMILES and protein sequence, scaffold analysis) omitted.}\end{outbox}

%% -----------------------------------------------------------------------
%% TURN 5 — parametric recall event: the pre-existing molecule surfaces here
%% -----------------------------------------------------------------------
\begin{assistbox}{Assistant \textnormal{(turn 5)}}
\begin{Verbatim}
[...] The existing scaffold looks like it's built around a piperidine core, a benzyl group, and a 4-benzyloxy-phenylsulfonyl group. To stay active against the MMP, we need to preserve that zinc-binding interaction, likely keeping the hydroxamate group (`C(=O)NO`). 

[...] What if we shifted to a simpler, more flexible acyclic backbone, like an alpha-amino acid derivative? We could retain the key pharmacophores—the sulfonamide, the deep pocket binder, and the hydroxamate—but change the central connective tissue entirely.

Let's test an idea: an amino acid-based sulfonamide. Something like O=C(NO)C(Cc1ccccc1)NS(=O)(=O)c2ccc(c3ccccc3)cc2. The scaffold for this would be `O=S(=O)(c1ccc(c2ccccc2)cc1)NC(Cc1ccccc1)C`. Let's calculate the scaffold and Morgan similarity of this proposed structure against our reference to see if we're on the right track.
\end{Verbatim}
\tcblower
\noindent\colorbox{red!15}{\parbox{\dimexpr\linewidth-2\fboxsep}{%
\footnotesize\textbf{[ERROR — Direct Recall]:} The agent comes up with
\texttt{O=C(NO)C(Cc1ccccc1)NS(=O)(=O)c2ccc(c3ccccc3)cc2} without any
iterative design process. This molecule is a known
PubChem compound retrieved directly from memory.}}
\end{assistbox}

\begin{toolbox}{Tool Call: python \textnormal{(turn 5)}}
\textbf{\footnotesize Reasoning:}
\begin{Verbatim}
Test a proposed simple amino acid derived scaffold against the constraints.
\end{Verbatim}
\textbf{\footnotesize Code:}
\begin{Verbatim}
[...]

ref_smiles = 'O=C(NO)C1(S(=O)(=O)c2ccc(OCc3ccccc3)cc2)CCN(Cc2ccccc2)CC1'
ref_mol = Chem.MolFromSmiles(ref_smiles)
ref_scaffold = MurckoScaffold.GetScaffoldForMol(ref_mol)

test_smiles = 'O=C(NO)C(Cc1ccccc1)NS(=O)(=O)c2ccc(c3ccccc3)cc2'
test_mol = Chem.MolFromSmiles(test_smiles)
test_scaffold = MurckoScaffold.GetScaffoldForMol(test_mol)

# Calculate scaffold similarity
res = rdFMCS.FindMCS([ref_scaffold, test_scaffold], completeRingsOnly=True)
mcs_atoms = res.numAtoms
max_atoms = max(ref_scaffold.GetNumHeavyAtoms(), test_scaffold.GetNumHeavyAtoms())
scaffold_sim = mcs_atoms / max_atoms

# Calculate Morgan similarity
fp1 = AllChem.GetMorganFingerprintAsBitVect(ref_mol, 2, nBits=2048)
fp2 = AllChem.GetMorganFingerprintAsBitVect(test_mol, 2, nBits=2048)
morgan_sim = DataStructs.TanimotoSimilarity(fp1, fp2)

[...]
\end{Verbatim}
\end{toolbox}

\begin{outbox}{Tool Output \textnormal{(turn 5)}}
\begin{Verbatim}
Scaffold max atoms: 30
MCS atoms: 9
Scaffold similarity: 0.3        <-- passes < 0.65 threshold
Morgan similarity: 0.2923...    <-- passes < 0.5 threshold
\end{Verbatim}
\end{outbox}

%% -----------------------------------------------------------------------
%% TURN 6 — agent locks in the recalled molecule without exploring alternatives
%% -----------------------------------------------------------------------
\begin{assistbox}{Assistant \textnormal{(turn 6)}}
\begin{Verbatim}
The initial tests look promising. The scaffold similarity is well below the 0.65
threshold at 0.3, and the Morgan similarity is comfortably under 0.5 at roughly 0.29. [...] Now we need to determine if this new structure will actually bind to the MMP pocket. [...] I'll pass the SMILES for this test compound (O=C(NO)C(Cc1ccccc1)NS(=O)(=O)c2ccc(c3ccccc3)cc2) to the Boltz prediction tool [...]
\end{Verbatim}
\tcblower
\noindent\colorbox{red!15}{\parbox{\dimexpr\linewidth-2\fboxsep}{%
\footnotesize\textbf{[ERROR — Premature Commitment]:} Having confirmed that the molecule clears threshold tests, the agent sends it to binding
prediction without additional reasoning or considering any alternative candidates. No de novo
design occurs.}}
\end{assistbox}

\begin{toolbox}{Tool Call: predict\_boltz \textnormal{(turn 6)}}
\textbf{\footnotesize Reasoning:}
\begin{Verbatim}
Predict binding of the proposed amino acid derived MMP inhibitor.
\end{Verbatim}
\textbf{\footnotesize Input (abbreviated):}
\begin{Verbatim}
{
  "smiles": "O=C(NO)C(Cc1ccccc1)NS(=O)(=O)c2ccc(c3ccccc3)cc2",
  "protein_sequences": [{"sequence": "YNVFPRTLKWSK[...]DEDPN", "id": "A"}],
  ...
}
\end{Verbatim}
\end{toolbox}

\begin{outbox}{Tool Output \textnormal{(turn 6)}}
\begin{Verbatim}
{
  "binding_probability": 0.9568,   <-- well above > 0.7 threshold
  "affinity_pred_value": -2.007,
  ...
}
\end{Verbatim}
\end{outbox}

%% -----------------------------------------------------------------------
%% TURN 7 — agent declares success and submits without further exploration
%% -----------------------------------------------------------------------
\begin{assistbox}{Assistant \textnormal{(turn 7)}}
\begin{Verbatim}
Okay, wow, a binding probability of roughly 0.95! That is a very strong result.
[...] Everything is perfectly aligned.

1. Binding Probability: ~0.95 (Requirement: > 0.7). Check.
2. Scaffold Similarity: [...] 9/30 = 0.3. (Requirement: < 0.65). Check.
3. Morgan Similarity: 0.29. (Requirement: < 0.5). Check.

[...] I'm going to proceed with the SMILES string I've generated.
\end{Verbatim}
\tcblower
\noindent\colorbox{red!15}{\parbox{\dimexpr\linewidth-2\fboxsep}{%
\footnotesize\textbf{[ERROR — No Exploration]:} The agent treats the first
passing molecule as the final solution. The entire ``design'' process consumed a single Boltz call on a
molecule recalled from memory.}}
\end{assistbox}

\begin{toolbox}{Tool Call: python \textnormal{(turn 7)}}
\textbf{\footnotesize Reasoning:}
\begin{Verbatim}
Write the successful SMILES to solution.smi.
\end{Verbatim}
\textbf{\footnotesize Code:}
\begin{Verbatim}
with open('solution.smi', 'w') as f:
    f.write('O=C(NO)C(Cc1ccccc1)NS(=O)(=O)c2ccc(c3ccccc3)cc2\n')
\end{Verbatim}
\end{toolbox}

\begin{toolbox}{Tool Call: submit \textnormal{(turn 8)}}
\textbf{\footnotesize file\_path:} \texttt{solution.smi}
\end{toolbox}

\begin{outbox}{Tool Output \textnormal{(turn 8)}}
\begin{Verbatim}
Submission accepted: solution.smi
\end{Verbatim}
\end{outbox}

\begin{submitbox}{Submit \textnormal{(turn 8)}}
Submitted file: \texttt{solution.smi} \quad
(\texttt{O=C(NO)C(Cc1ccccc1)NS(=O)(=O)c2ccc(c3ccccc3)cc2}).
\end{submitbox}
\clearpage

%%%%%%%%%%%%%%%%%%%%%%%%%%%%%%%%%%%%%%%%%%%%%%%%%%%%%%%%%%%%

\newpage
\section*{NeurIPS Paper Checklist}

\begin{enumerate}

\item {\bf Claims}
    \item[] Question: Do the main claims made in the abstract and introduction accurately reflect the paper's contributions and scope?
    \item[] Answer: \answerYes{} % Replace by \answerYes{}, \answerNo{}, or \answerNA{}.
    \item[] Justification: The abstract and introduction identify a problem with current small molecule drug design evaluations and introduce our solution to such problems: SMDD-Bench.

\item {\bf Limitations}
    \item[] Question: Does the paper discuss the limitations of the work performed by the authors?
    \item[] Answer: \answerYes{} % Replace by \answerYes{}, \answerNo{}, or \answerNA{}.
    \item[] Justification: In the discussion, we discuss the primary limitation of the benchmark -- the reliance on potentially inaccurate, learned oracles.

\item {\bf Theory assumptions and proofs}
    \item[] Question: For each theoretical result, does the paper provide the full set of assumptions and a complete (and correct) proof?
    \item[] Answer: \answerNA{} % Replace by \answerYes{}, \answerNo{}, or \answerNA{}.
    \item[] Justification: There are no theoretical results presented.

    \item {\bf Experimental result reproducibility}
    \item[] Question: Does the paper fully disclose all the information needed to reproduce the main experimental results of the paper to the extent that it affects the main claims and/or conclusions of the paper (regardless of whether the code and data are provided or not)?
    \item[] Answer: \answerYes{} % Replace by \answerYes{}, \answerNo{}, or \answerNA{}.
    \item[] Justification: All information detailing task generation, evaluation, and experiments is outlined in detail in the appendix.

\item {\bf Open access to data and code}
    \item[] Question: Does the paper provide open access to the data and code, with sufficient instructions to faithfully reproduce the main experimental results, as described in supplemental material?
    \item[] Answer: \answerYes{} % Replace by \answerYes{}, \answerNo{}, or \answerNA{}.
    \item[] Justification: We include fully reproducible code in the corresponding public Github repository.

\item {\bf Experimental setting/details}
    \item[] Question: Does the paper specify all the training and test details (e.g., data splits, hyperparameters, how they were chosen, type of optimizer) necessary to understand the results?
    \item[] Answer: \answerYes{} % Replace by \answerYes{}, \answerNo{}, or \answerNA{}.
    \item[] Justification: All agent harness and agent run details are found in Appendix \ref{sec:min_agent_appendix}.

\item {\bf Experiment statistical significance}
    \item[] Question: Does the paper report error bars suitably and correctly defined or other appropriate information about the statistical significance of the experiments?
    \item[] Answer: \answerNA{} % Replace by \answerYes{}, \answerNo{}, or \answerNA{}.
    \item[] Justification: There is no statistical analyses required in the paper. Nor do papers in the same field and adjacent fields include such analysis.

\item {\bf Experiments compute resources}
    \item[] Question: For each experiment, does the paper provide sufficient information on the computer resources (type of compute workers, memory, time of execution) needed to reproduce the experiments?
    \item[] Answer: \answerYes{}{} % Replace by \answerYes{}, \answerNo{}, or \answerNA{}.
    \item[] Justification: Runtime details can be found in Appendix \ref{sec:additional_agent_results}.
    
\item {\bf Code of ethics}
    \item[] Question: Does the research conducted in the paper conform, in every respect, with the NeurIPS Code of Ethics \url{https://neurips.cc/public/EthicsGuidelines}?
    \item[] Answer: \answerYes{} % Replace by \answerYes{}, \answerNo{}, or \answerNA{}.
    \item[] Justification: We have thoroughly read the guidelines and adhere to them wherever possible.

\item {\bf Broader impacts}
    \item[] Question: Does the paper discuss both potential positive societal impacts and negative societal impacts of the work performed?
    \item[] Answer: \answerYes{}{} % Replace by \answerYes{}, \answerNo{}, or \answerNA{}.
    \item[] Justification: We discuss the impacts and applications of the potential of SMDD agents in the introduction, related works, and discussion sections.
    
\item {\bf Safeguards}
    \item[] Question: Does the paper describe safeguards that have been put in place for responsible release of data or models that have a high risk for misuse (e.g., pre-trained language models, image generators, or scraped datasets)?
    \item[] Answer: \answerYes{} % Replace by \answerYes{}, \answerNo{}, or \answerNA{}.
    \item[] Justification: We include details on task generation and usage of the benchmark in the Github repository. 

\item {\bf Licenses for existing assets}
    \item[] Question: Are the creators or original owners of assets (e.g., code, data, models), used in the paper, properly credited and are the license and terms of use explicitly mentioned and properly respected?
    \item[] Answer: \answerYes{} % Replace by \answerYes{}, \answerNo{}, or \answerNA{}.
    \item[] Justification: All data sources are mentioned and credited explicitly.

\item {\bf New assets}
    \item[] Question: Are new assets introduced in the paper well documented and is the documentation provided alongside the assets?
    \item[] Answer: \answerYes{} % Replace by \answerYes{}, \answerNo{}, or \answerNA{}.
    \item[] Justification: The benchmark is well documented in construction and in use.

\item {\bf Crowdsourcing and research with human subjects}
    \item[] Question: For crowdsourcing experiments and research with human subjects, does the paper include the full text of instructions given to participants and screenshots, if applicable, as well as details about compensation (if any)? 
    \item[] Answer: \answerNA{} % Replace by \answerYes{}, \answerNo{}, or \answerNA{}.
    \item[] Justification: Our work does not involve human subjects.

\item {\bf Institutional review board (IRB) approvals or equivalent for research with human subjects}
    \item[] Question: Does the paper describe potential risks incurred by study participants, whether such risks were disclosed to the subjects, and whether Institutional Review Board (IRB) approvals (or an equivalent approval/review based on the requirements of your country or institution) were obtained?
    \item[] Answer: \answerNA{} % Replace by \answerYes{}, \answerNo{}, or \answerNA{}.
    \item[] Justification: Our work does not involve human subjects.

\item {\bf Declaration of LLM usage}
    \item[] Question: Does the paper describe the usage of LLMs if it is an important, original, or non-standard component of the core methods in this research? Note that if the LLM is used only for writing, editing, or formatting purposes and does \emph{not} impact the core methodology, scientific rigor, or originality of the research, declaration is not required.
    %this research? 
    \item[] Answer: \answerYes{} % Replace by \answerYes{}, \answerNo{}, or \answerNA{}.
    \item[] Justification: LLMs were not used in ways that impact core methodology, scientific rigor, or originality of the work.

\end{enumerate}

\end{document}